\documentclass[12pt]{iopart}

\usepackage{paralist}  % list embedded into paragraph
\usepackage{subcaption}    % multiple sub-titles
\usepackage{graphicx}
\graphicspath{{Figs/}}
\expandafter\let\csname equation*\endcsname=\relax 
\expandafter\let\csname endequation*\endcsname=\relax
\usepackage{amsmath}

\usepackage{amsfonts}
\usepackage{fullpage}
\usepackage{multirow}

\usepackage{pgfplots}
\pgfplotsset{compat=1.16,width=8cm,height=8cm}

\usepackage[export]{adjustbox}

\usepackage{microtype} % to fix overfull \hbox
\usepackage{url}
\usepackage{algorithm}

\usepackage{algpseudocode}
\usepackage{algcompatible}

\newcommand{\bfA}{{\bf A}}

\newcommand{\bfC}{{\bf C}}

\newcommand{\bfE}{{\bf E}}

\newcommand{\bfG}{{\bf G}}

\newcommand{\bfI}{{\bf I}}

\newcommand{\bfK}{{\bf K}}

\newcommand{\bfZ}{{\bf Z}}

\newcommand{\bfb}{{\bf b}}

\newcommand{\bfx}{{\bf x}}
\newcommand{\bfy}{{\bf y}}
\newcommand{\bfu}{{\bf u}}

\newcommand{\bfd}{{\bf d}}

\newcommand{\bfr}{{\bf r}}

\newcommand{\bfw}{{\bf w}}
\newcommand{\bfz}{{\bf z}}

\newcommand{\hf}{{\frac 12}}
\newcommand{\grad}{{\boldsymbol \nabla}}

\newcommand{\bfepsilon}{{\boldsymbol \epsilon}}
\newcommand{\bftheta}{{\boldsymbol \theta}}
\newcommand{\bfxi}{{\boldsymbol \xi}}

\newtheorem{theorem}{Theorem}[section]

\begin{document}

\title{DRIP: Deep Regularizers for Inverse Problems}

\author{Moshe Eliasof\textsuperscript{1}, Eldad Haber\textsuperscript{2}, Eran Treister\textsuperscript{1}}

\address{\textsuperscript{1} Computer Science Department, Ben-Gurion University of the Negev, Beer Sheva, Israel.}

\address{\textsuperscript{2} Department of Earth Ocean and Atmospheric Sciences, University of British Columbia, Vancouver, Canada.}

\eads{eliasof@post.bgu.ac.il, eldadhaber@gmail.com, erant@cs.bgu.ac.il}

\begin{abstract}
{In this paper we consider} inverse problems {that} are mathematically ill-posed. {That is}, given some (noisy) data, there is more than one solution that {approximately} fits the data. In recent years, deep neural techniques that find the most appropriate solution, in the sense that it contains a-priori information, were developed. However, they suffer from several shortcomings. First, most techniques cannot guarantee that the solution fits the data at inference. Second, while the derivation of the techniques is inspired by the existence of a valid scalar regularization function, such techniques do not in practice rely on such a function, and therefore veer away from classical variational techniques. In this work we introduce a new family of neural regularizers for the solution of inverse problems. These regularizers are based on a variational formulation and are guaranteed to fit the data.
We demonstrate their use on a number of highly ill-posed problems, from image deblurring to limited angle tomography.
\end{abstract}

\section{Introduction}

Employing machine learning and in particular deep learning frameworks
for the solution of inverse problems is now well established (see e.g.
\cite{ongie2020deep,bai2020deep,aggarwal2018modl,jin2017deep,gottschling2020troublesome,adler2017solving,lucas2018using,mardani2018neural, barbano2022unsupervised,haltmeier_dataconsistent} and references within). In this work, we consider the canonical discrete linear inverse problem, where
the forward problem is given by
\begin{equation}
 \label{forlin} 
 \bfA \bfu + \bfepsilon = \bfb,
\end{equation}
where $\bfA$ is an $m \times n$  forward mapping matrix that is a discretization of
a compact linear operator, $\bfepsilon$ is a noise vector, which we
assume to be Gaussian i.i.d, and $\bfb$ denotes the observed data. In most
applications $m<n$ (i.e., fewer equations than unknowns), and for the applications we consider here, $\bfA$ is either rank deficient or contains many small singular values.
The effective null space of $\bfA$ and the noise implies that one cannot uniquely recover $\bfu$ from the observed data $\bfb$. We broadly classify the techniques  to obtain a unique solution for $\bfu$ given
$\bfb$ into two main categories.  
The first is {\em goal based} and the second
is {\em process based}.  In the goal-based algorithms, one converts the
solution of the inverse problem into an optimization problem, and then,
uses optimization algorithms to solve the converted problem. Algorithms
that belong to this category are, for example, MAP estimates in Bayesian reconstruction \cite{tensiam}, Total Variation recovery \cite{rudin1992nonlinear}, sparsity-based recovery \cite{donoho04sparsest,treister2016multilevel}, and algorithms that are based on minimizing a regularized least squares \cite{zibulevsky2010l1,tensiam}, some of which employ the approach of proximal methods \cite{parikh2014proximal}.
A second type of algorithm is based on a process. Typically, the
process is iterative and is motivated by the goal. For example, 
algorithms that are based on inverse scale space \cite{burger2006}, on the early stopping of Conjugate Gradient Least Squares \cite{hansen} or Landweber iteration \cite{ehn1}, and iterated Tikhonov regularization \cite{hanke-nonstationary}.
Such algorithms do not necessarily minimize an objective, but they may end up being not far from the minimizer of the goal and therefore yield small reconstruction errors. Also, they
can be more computationally efficient compared to goal-based methods.

With the rise and success of machine and deep learning in recent years  \cite{krizhevsky2012imagenet,he2016deep, ronneberger2015u}, it became natural to employ deep learning frameworks to solve inverse problems \cite{adler2017solving, haltmeier_dataconsistent, barbano2022unsupervised}.
In general, learning algorithms are utilized to obtain
a unique solution to an inverse problem by combining them with an existing algorithm, either goal based or process based. Early
algorithms (\cite{haten} and later \cite{ChungSq2021}) used bi-level optimization in order
to learn to optimize the parameters of a regularization function.
However, due to the repeated solution of a nonlinear
optimization problem at their core, such algorithms suffer from high computational complexity. More recently, process based
algorithms have been proposed to replace the bi-level optimization approach. In particular, 
algorithms that are based on ``neural proximal'' methods \cite{mardani2018neural} have been used,
where the proximal iteration is replaced with a deep neural network.
Such methods are relatively simple to implement and are therefore popular for the solution of inverse problems. Consequently, in recent years,
algorithms that are based on bi-level optimization techniques have been 
often overlooked in favor of process-based algorithms.

Nonetheless, while
process-based methods are indeed attractive, they suffer
from several significant shortcomings \cite{gottschling2020troublesome}.
The first and most important is that they do not necessarily accurately fit the
observed data. This is because the process, while motivated by regularized data fitting, is rarely carried to completion. In fact, 
most methods propose to use only several proximal iterations or their equivalent. As a result,
data fitting is hardly guaranteed. Furthermore, as we show in our
numerical experiments in Section \ref{sec5}, if the number of proximal iterations is changed
after training, such methods may diverge. This makes proximal methods
work well only under very controlled scenarios and assumptions.
A second shortcoming of these techniques is that they usually do not compute
a scalar regularization or a potential, but rather compute only
approximations to its gradient. As a result, given two possible solutions
$\bfu_1$ and $\bfu_2$, it is difficult to compare them or to assign some uncertainty to these solutions. 

These observations motivate us to develop a new class of learning deep neural regularization 
functionals for the solution of inverse problems, called DRIP. Our method is based on bi-level optimization, with a convex lower level optimization problem. In this
way, we can use a goal-based algorithm that allows us to fit the
data on the one hand, and learn a valid scalar regularization on the other. 
Because our DRIP is based on a variational approach, our method can be used to fit data with high accuracy, when needed. Also, it is more stable than proximal-based methods in the presence of varying noise levels.

The rest of the paper is organized as follows.
In Section~\ref{sec2} we review deep learning
techniques that are based on proximal methods for inverse problems and discuss their shortcomings. In Section~\ref{sec3}, we propose a new
regularization functional that is based on the least action principle and show how it can be used efficiently for the solution of inverse problems. 
In Section~\ref{sec4}
we discuss the training details of the network. In Section~\ref{sec5}, we perform a number of numerical experiments and we summarize the paper in Section~\ref{sec6}.

\section{From Goal  to Process Oriented Algorithms}
\label{sec2}

In this section, we give a short overview of different algorithms, and we
show how these algorithms are motivated and used for the solution of inverse problems.
We start with \emph{goal} oriented recovery, that is, the solution to the inverse
problem is obtained by solving the following minimization problem
\begin{eqnarray}
\label{optreg}
\bfu_*(\bftheta) = {\rm arg}\min_{\bfu} \ \hf \|\bfA \bfu - \bfb\|^2 + R(\bfu, \bftheta).
\end{eqnarray}
The objective function is composed of two parts. The first is a data-fitting one, and the second is the regularization, $R$, that depends on
both $\bfu$ and some trainable parameters, $\bftheta$.
To obtain the optimal parameters, $\bftheta$, we assume to have  $m$ realizations
$\bfu^{(j)}, j=1,\ldots,m$ and we solve the bi-level optimization
problem
\begin{subequations}
\label{bilevel}
\begin{eqnarray}
\label{bilevel1}
\min_{\bftheta} && \hf \sum_{j=1}^m \|\bfu^{(j)} - \bfu_*^{(j)}(\bftheta) \|^2, \\
\label{bilevel2}
{\rm s.t} && \bfu_*^{(j)}(\bftheta) = {\rm arg}\min_{\bfu} \ \hf \|\bfA \bfu - \bfb^{(j)}\|^2   + R(\bfu, \bftheta),
\end{eqnarray}
\end{subequations}
where $\bfb^{(j)} = \bfA \bfu^{(j)} + \bfepsilon^{(j)}$. Solving this bi-level optimization problem is challenging. The main difficulty stems from the fact that the lower optimization problem in Equation \eqref{bilevel2}
is not easy to solve. To this end, in a pioneering work of 
Chen and Pock \cite{chen2016trainable}, the lower optimization problem is replaced
with a process-based algorithm that, if carried to completion, yields
the solution of the optimization problem. For example, one can replace
Equation \eqref{bilevel2} with a gradient-based algorithm of the form (omitting the superscript $^{(j)}$ for simplicity)  
\begin{eqnarray}
\label{sd}
\bfu_{\ell+1} = \bfu_{\ell} - \alpha(\bfA^{\top} (\bfA \bfu_{\ell} - \bfb) - \nabla_{\bfu} R(\bfu, \bftheta)),
\end{eqnarray}
where $\alpha > 0$ is a step-size. Such a simple algorithm may require many iterations, and therefore, proximal methods \cite{parikh2014proximal} were suggested, where the following algorithm is applied
\begin{subequations}
\label{prox}
\begin{eqnarray}
\label{prox0}
\widehat \bfu_{\ell+1} &=& \bfu_{\ell} - \alpha \bfA^{\top} (\bfA \bfu_{\ell} - \bfb), \\
\label{prox1}
\bfu_{\ell+1} &=&  {\rm arg}\min_{\bfu}\ \hf \|\bfu-\widehat\bfu_{\ell+1} \|^2 + R(\bfu, \bftheta)).
\end{eqnarray}
\end{subequations}
The iteration in Equation~\eqref{prox} is divided into two parts. The first, \eqref{prox0}, is the data fitting, and the second, \eqref{prox1}, is the proximal iteration that aims to find a solution that decreases the value of the regularization functional. The method is best suited for cases where $R(\bfu, \bftheta)$ is separable in $\bfu$ (obtained element-wise over the entries of $\bfu$), and hence \eqref{prox1} is separable as well and can be solved element-wise. A known example for this case is the $\ell_1$ norm sparsity-promoting regularization \cite{zibulevsky2010l1,donoho04sparsest,treister2016multilevel}. 

For a general (coupled) regularizer, the proximal methods face a problem: solving for the proximal step can be very difficult. Noting that the proximal step is an optimization problem---a mapping---with input $\widehat \bfu_j$ and output $\bfu_j$, the work of \cite{mardani2018neural} suggested learning a ``Neural Proximal'' operator for this case and to replace the optimization in the proximal step by some learned mapping, that
is
\begin{equation}
\label{net}
\bfu_{\ell+1} = f(\widehat \bfu_{\ell+1}, \bftheta),
\end{equation}
where $f$ is a deep neural network parameterized by learnable weights  $\bftheta$. 

{Typically, there are two common architectural choices for $f$. First,  residual neural networks, also known as ResNet \cite{he2016deep}, where the pooling layers are omitted, and the typical classification layer is replaced with a $1\times1$ convolution that projects the high dimensional representation of the solution to the desired output dimension. Second, and more straightforward, a UNet \cite{ronneberger2015u} architecture is used, where the output dimension is regarded as the desired solution shape, as opposed to the typical interpretation of segmentation classes (e.g., RGB channels of a recovered image vs. background/object classes).}

This approach makes the implementation of the algorithm easily tractable. Combining Equations \eqref{prox} and \eqref{net},  we summarize the neural proximal methods as
\begin{equation}
\label{net1}
\bfu_{\ell+1} = f(\bfu_{\ell} - \alpha \bfA^{\top} (\bfA \bfu_{\ell} - \bfb), \bftheta).
\end{equation}
The convergence of such algorithms for a fixed $\bftheta$  can be proven when considering fixed point iteration. 
However, since the parameters $\bftheta$ are obtained by training, the fixed point iteration conditions may not be fulfilled, 
and, more importantly, may perform poorly when facing out-of-distribution inputs, or setups that are different than the training setup (e.g., different noise levels or distributions). Furthermore, it has been observed in practice (as we also show in our numerical results), that when the algorithm is trained using a fixed number of iterations, it tends to yield worse results when increasing the number of iterations  
during the inference phase.
It is important to note that while the proximal iteration in Equation \eqref{prox1} is defined as an optimization problem, the network $f$ that replaces it is, in general, not a solution to any optimization problem. We therefore do not have any access to a valid regularization functional $R$ and, the convergence properties of such a scheme are poorly understood \cite{gottschling2020troublesome}.

The morphing of algorithms from goal-oriented in Equation \eqref{optreg} to process-based in Equation \eqref{net1}, allowed the incorporation of deep learning techniques into inversion algorithms. Nonetheless, giving up on the goal, comprised of data fitting and regularization, yields algorithms that may not even fit the data. We now discuss how to alleviate this problem by proposing regularization
techniques that are easy to compute and incorporate within the process. Such a regularization results in a new type of neural architecture that we call DRIP, and it can be efficiently implemented and used to solve inverse problems with data fitness guarantees.

\section{Variational Methods for Regularization and the Least Action Network}
\label{sec3}

The DRIP approach we present here is built upon variational techniques. That is, we present a new \emph{learnable} regularization functional that is minimized, and show how it can be used in the context of the goal-based solution of the inverse problem.

\subsection{The Least Action Regularization}

Similar to many scenarios in machine learning, we define our regularization function on a larger latent space ${\cal Z} \subseteq \mathbb{R}^s$. To this end, rather than solving for $\bfu$, we define 
a learnable embedding matrix $\bfE \in \mathbb{R}^{n \times s}$ that maps the vectors $\bfz \in \mathbb{R}^s $ to vectors $\bfu \in \mathbb{R}^n$, that is
\begin{equation}
\label{emb}
\bfu = \bfE \bfz.
\end{equation}
The matrix $\bfE$ can also be thought of as a dictionary (see \cite{eladReview} and references within). Next, we define the matrix $\bfZ = [\bfz_0, \bfz_1, \ldots, \bfz_N]$ where ${\bfz_i\in \cal Z}$, and replace the original problem \eqref{optreg}
with the following optimization problem
\begin{equation}
\label{optRegla}
\min_{\bfz, \bfZ} = \hf \|\bfA \bfE \bfz - \bfb\|^2 +
\alpha R(\bfz, \bfZ; \bftheta).
\end{equation}

We define the learnable regularization function by
\begin{equation}
\label{lsa}
R(\bfz, \bfZ;  \bftheta) = \hf \| \bfz - \bfz_N\|^2 +  E_K(\bfZ) + E_P(\bfZ)
\end{equation}
with
\begin{subequations}
\label{energy}
\begin{eqnarray}
E_K(\bfZ) &=& {\frac 1{2}} \sum_{\ell=0}^{N-1} \|\bfz_{\ell+1} - \bfz_{\ell}\|^2 \\
E_P(\bfZ;  \bftheta) &=&  \sum_{\ell=0}^{N}
 \phi(\bfz_{{l}}; \bftheta_{{l}}). 
\end{eqnarray}
\end{subequations}
Here, $\phi(\cdot, \cdot)$ is a nonlinear convex function with learnable parameters, that typically involves convolutions and activations (we discuss its exact form later).
The regularization can be interpreted as having a vector in a continuous time domain $\bfz(t), \ 0\le t \le T$ where $\bfz(t=0) = \bfz_0$ is a rough approximation to the solution of the inverse problem, and $\bfz(t=T)$ is closer to the true solution. If the vectors $\bfz_j$ are, for example, the coordinates (in $\mathbb{R}^s$) of a particle, then, by solving the optimization problem we recover the path from $\bfz_0$ to $\bfz_N$ that closely fits the data.
The path minimizes the energy composed from kinetic energy $E_K$ (the squared velocity)
and the potential energy
$E_P$ that is learned. This form is common for control problems where the least-action principle
is applied \cite{kalman1960contributions}. 

Before we discuss the solution to the optimization problem, we introduce the following theorem:
\begin{theorem}    
Let the regularization function $R$ be defined by equations
\eqref{lsa} and \eqref{energy}. Let the nonlinearities $\phi(\bfz, \bftheta)$ be strongly convex with respect to $\bfz$. Then, the optimization 
problem \eqref{optRegla} has a unique solution $\bfz, \bfz_1, \ldots, \bfz_N$ given $\bftheta$.
\end{theorem}
The proof of the theorem is trivial, as in addition to the sum over the $\phi$ terms in $E_P$, all the other terms in \eqref{optRegla} and \eqref{lsa} are least-squares terms, and hence convex. The theorem is important for two main reasons. First, it implies that the solution to the problem does not depend on the initial choice of $\bfz$ when using an iterative method. This is in contrast to proximal approximations where the initialization of the network determines its last state. Second, it means that the solution of the network is stable. Indeed, the stability of boundary value problems that evolve from the minimization of convex functions is well studied \cite{strang}
with provable stability and existence, which makes the solution to the training problem easier. This formulation also overcomes the difficulties of deep learning in the context of inverse problems that are presented in \cite{gottschling2020troublesome}. That is, since the Hessian of the regularization is strictly positive, the network is stable with respect to noise.

\subsection{Choice of Nonlinearities and Potentials}

To have a convex objective \eqref{optreg}, $\phi$ needs to be convex according to our theorem. This can be achieved if we choose $\phi$ to include an integration of common monotonically non-decreasing activation functions such as ReLU$()$ or $\tanh()$. This way, its derivative can resemble a neural network layer of a common type. Specifically, here we use a single-layer potential of the form
\begin{eqnarray}
\label{potentials}
\phi(\bfz) &=& \exp\left(\bfw^{\top}\right)\sigma (\bfK \bfz)  \\
\nonumber
 \nabla \phi(\bfz) &=& \bfK^{\top} {\rm diag}\left( \exp(\bfw) \right) \sigma'(\bfK \bfz),
\end{eqnarray}
where  $\sigma()$ is a point-wise nonlinear function. Here, we used the nonlinearity
\begin{eqnarray}
\label{nonlin}
\sigma(t) = \left\{ \begin{matrix} {\frac a2} t^2 & {\rm if} \ t>0 \\  {\frac b2} t^2 & {\rm if} \ t\le 0 \end{matrix} \right.,
\end{eqnarray} whose derivative, which is required for the gradient of the potential, yields the common leaky ReLU activation function \cite{xu2020reluplex}. This way, $\nabla \phi$ has the form of a rather standard neural network layer. Our choice in \eqref{nonlin} also guarantees that we have a bounded and convex energy function as the Hessian of $\phi$ is positive semi-definite:
\begin{equation}
\nabla^2\phi = \bfK^\top{\rm diag}\left( \exp(\bfw)\odot  \sigma'(\bfK \bfz)\right)\bfK \succeq 0.
\end{equation} 
This implies that the solution of the regularized inverse problem~\eqref{optRegla} has a unique solution. As we see next, choosing
$\phi$ this way leads, in our case,
to a particular form of a residual network with a double skip connection. The use of double skip connections was previously abundantly used \cite{ruthotto2019deep, eliasof2021mimetic, eliasof2021pde} to design hyperbolic neural architectures that are motivated by the underlying neural energy of the network.

\subsection{Solution of the Inverse Problem}

Given the convex function $\phi$, we now consider the following iteration for the solution of the optimization problem in Equation \eqref{optRegla} by the following alternating minimization procedure\footnote{Note that the superscript $t$ refers to an inner iteration and not a training sample.}:
\begin{subequations}
\label{cd}
\begin{eqnarray}
\label{cd1}
\bfZ^{(t+1)} &=& {\rm arg}\min_{\bfZ} \ \hf \| \bfz_N - \bfz_*^{(t)}\|^2 + E_k(\bfZ)   + E_p(\bfZ; \bftheta) \\
\label{cd2}
\bfz_*^{(t+1)} &=& {\rm arg} \min_{\bfz} \hf \|\bfA \bfE \bfz - \bfb\|^2 + 
{\frac \alpha 2} \| \bfz-\bfz_N^{(t+1)}\|^2, 
\end{eqnarray}
\end{subequations}
starting from some $\bfz_*^{(0)}$ (we choose $\bfz_*^{(0)}$ as the solution of \eqref{cd2} for $\bfz^{(0)}_N=0$).
\noindent This iteration can be interpreted as a block coordinate descent method for the solution of the problem, where in the first stage we solve for a trajectory and in the second we solve for the solution that is close to this trajectory but also fits the data. This way, at the end of the process the data is guaranteed to be fitted. We note that in practice, neither problem needs to be solved exactly for the method to converge \cite{eich} but rather an approximate solution of these problems suffice. Thus we now focus on the approximate solutions of each of the sub-problems.

\subsubsection{Solving the Data Fitting Problem.}

The solution of the data fitting problem from Equation \eqref{cd2} is  {obtained by solving a regularized data fitting problem that reads}
\begin{eqnarray}
\label{datafit}
\bfz_*^{(t+1)} = ( \bfE^{\top}\bfA^{\top}\bfA \bfE + \alpha \bfI)^{-1}( \bfE^{\top}\bfA^{\top}\bfb + \alpha \bfz_N^{(t+1)}).
\end{eqnarray}
That is, the algorithm requires the solution of a regularized linear system at each step. In our implementation we use 
the conjugate gradient least squares (CGLS) \cite{hansen}.
Typically, a small number
of CGLS iterations is sufficient to fit the data, especially at later stages of the algorithms where $\bfz_N$ is already a good initialization for the solution. 
 
\subsubsection{Approximately Solving the Least Action Problem.}

To minimize the regularization term in \eqref{cd1} we note that the conditions for a minimum read
\begin{subequations}
\label{el}
\begin{eqnarray}
\label{dlsa}
&& 2\bfz_{1} - \bfz_2 +  \nabla\phi(\bfz_{1}, \bftheta_{1}) = \bfz_0\\
&& 2 \bfz_{\ell} -\bfz_{\ell-1} - \bfz_{\ell+1} +  \nabla\phi(\bfz_{\ell}, \bftheta_{\ell}) = 0\ \ \ \ell=2, .., N-1 \\
&&  2\bfz_N-\bfz_{N-1} +  \nabla\phi(\bfz_{N}, \bftheta_{N})  = \bfz_*^{(t)}.
\end{eqnarray}
\end{subequations}
This system is a boundary value problem for $[\bfz_1, \ldots, \bfz_N]$ that seems to be difficult to solve. Note that the problem can also be compactly written as
\begin{eqnarray}
\label{bvp}
&& \begin{bmatrix}
2\bfI &   -\bfI  &    &   &   &    \\
-\bfI &   2\bfI  & -\bfI   &   &   & \\
 &   \ddots  & \ddots  & \ddots  &   & \\
 &     &   & -\bfI  & 2\bfI  &  -\bfI \\
 &     &   & &  -\bfI & 2\bfI\\
\end{bmatrix}
\begin{bmatrix}
\bfz_1 \\ \bfz_2 \\  \vdots \\ \bfz_{N-1} \\ \bfz_N
\end{bmatrix}  
= \begin{bmatrix}
\bfz_0 - \grad \phi(\bfz_1,\bftheta_1) \\ -\grad \phi(\bfz_2,\bftheta_2) \\  \vdots \\ -\grad\phi(\bfz_{N-1}, \bftheta_{N-1}) \\  \bfz_*^{(t)} - \grad \phi(\bfz_N,\bftheta_N)  
\end{bmatrix} 
\end{eqnarray}
The matrix on the left-hand side is a block triangular matrix with an analytic
Cholesky factorization: 
\begin{eqnarray}
\label{chol}
\bfC = 
\begin{bmatrix}
 a_1 \bfI &   -{\frac 1 {a_1}}\bfI  &    &   &   &    \\
0 &   a_2\bfI  & -{\frac 1 {a_2}}\bfI   &   &   & \\
 &   0  & \ddots  & \ddots  &   & \\
 &     &   &   & {a_{N-1}}\bfI  &  -{\frac 1 {a_{N-1}}}\bfI \\
 &     &   & &   & a_N\bfI\\
\end{bmatrix},
\end{eqnarray}
where $a_j = \sqrt{{\frac{j+1}{j}}}$.
This suggests that the following fixed point iteration for the solution of the problem is first made by a forward 
substitution for the matrix, or a forward network propagation:
\begin{subequations}
\label{forprop}
\begin{eqnarray}
\label{forprop0}
\bfy_1 &=& a_1^{-1} (\bfu_0 - \grad \phi(\bfz_1, \theta_1)) \\ 
\bfy_j &=&  a_j^{-1} \left( a_{\ell-1}^{-1} \bfy_{\ell-1} - \grad \phi(\bfz_\ell, \bftheta_\ell) \right) \\
\bfy_N &=&  a_N^{-1} \left( a_{N-1}^{-1} \bfy_{N-1} - \grad \phi(\bfz_N, \bftheta_{N}) \right).
\end{eqnarray}
\end{subequations}
This is followed by a backward propagation (or, substitution)
\begin{subequations}
\label{backprop}
\begin{eqnarray}
\label{backprop1}
\bfz_N &=& a_N^{-1} (\bfy_N + a_N^{-1} \bfz_*) \\ 
\bfz_{\ell-1} &=&  a_{\ell-1}^{-1} \left( a_{\ell-1}^{-1} \bfz_{\ell} - \bfy_{\ell-1} \right),\quad \ell=N,...,1.
\end{eqnarray}
\end{subequations}
Note that the procedure above is only an approximate solution of the problem \eqref{cd1} because of the nonlinear $\nabla\phi(\bfz_j)$ terms on the {right}-hand side of \eqref{bvp}. Nonetheless, since only an approximate solution is required for the convergence of the outer this solution is sufficient. 

\subsection{The Complete Algorithm}

Combining the data fitting problem from Equation \eqref{datafit} with the least action problem from Equation \eqref{dlsa} we obtain Algorithm \ref{alg:ipsolve} for the solution of the inverse problem, which we refer to as the least action neural network, or {\em LA-Net}. For the examples presented in this paper we apply only a single iteration of the algorithm (unless stated otherwise), as we have found experimentally that more iterations do not significantly improve the results. However, several iterations may be required for other scenarios.

\begin{algorithm}[tb]
   \caption{LA-Net - Solution of the least action regularized inverse problem}
   \label{alg:ipsolve}
\begin{algorithmic}
   \STATE Initialize $\bfz_1, \ldots, \bfz_N=0$, $\bfz_0 = \bfz_* = ( \bfE^{\top}\bfA^{\top}\bfA \bfE + \alpha \bfI)^{-1}( \bfE^{\top}\bfA^{\top}\bfb)$.
\FOR{$t=1,\ldots,maxIter$}
   \STATE \textit{\# Solve Least Action:}
   \STATE Set $\bfy_1 = a_1^{-1} (\bfz_0 - \grad \phi(\bfz_1, \theta_1)) $
   \FOR{$\ell=1,\ldots,N$} 
   \STATE Forward pass: $\bfy_{\ell} =  a_{\ell}^{-1} \left( a_{\ell-1}^{-1} \bfy_{\ell-1} - \grad \phi(\bfz_{\ell}, \bftheta_{\ell}) \right) $
   \ENDFOR
   \STATE Set $\bfy_N =  a_N^{-1} \left( a_{N-1}^{-1} \bfy_{N-1} - \grad \phi(\bfz_N, \bftheta_N) + \bfz_* \right) $
 \FOR{$\ell=N:-1:1$} 
    \STATE Backward pass $\bfz_{\ell-1} =  a_{\ell-1}^{-1} \left( a_{\ell-1}^{-1} \bfz_{\ell} - \bfy_{\ell-1} \right)$
   \ENDFOR
\STATE \textit{\#Solve Data Fit:}
\STATE $\bfz_* = ( \bfE^{\top}\bfA^{\top}\bfA \bfE + \alpha \bfI)^{-1}( \bfE^{\top}\bfA^{\top}\bfb + \alpha \bfz_N)$
\ENDFOR
\STATE \textbf{return} $\bfz_*$.
\end{algorithmic}
\end{algorithm}

There are a number of important characteristics of the algorithm proposed here. First, the algorithm generates solutions $\bfz_*$ that fit the data at each step. Therefore, even if terminated early, it yields feasible solutions. This is 
in contrast to algorithms that are based on the proximal iteration \cite{ongie2020deep, mardani2018neural}, that if terminated early do not always fit the data.
We demonstrate this point in our experiments in Section~\ref{sec5}.
A second feature of our algorithm is the ability to adapt to problem-dependent settings. For example, if the noise
statistics involve outliers, it is straightforward to replace the
data-fitting term from Equation \eqref{datafit} with an M-estimator \cite{tyler1987distribution} that is designed for such statistics. This flexibility allows our method to cope with 
different loss functions that are used for the solution of inverse problems and various noise regimes. Finally, our algorithm is based on a true, valid potential function.
This implies that standard techniques, for example, Bayesian analysis, or other techniques that require the value of another potential or energy function can be applied to our DRIP.

\subsection{A Non-iterative Approximation}

Algorithm \ref{alg:ipsolve} is an iterative algorithm {which} requires solving a boundary value problem (BVP). 
The BVP estimates $\bfz_*$ and $\bfZ = [\bfz_1, \ldots, \bfz_N]$ given
$\bfz_0$ and the data $\bfb$, and yields an equation for
the final (terminal) condition.
However, note that at the solution, $\bfz_1$ depends only on $\bfz_0$ and $\bfz_*$. To this end, we learn a mapping from $(\bfz_0,\bfz_*)$ to $\bfz_1$
\begin{equation}
\label{netz1}
\bfz_1 = \bfG(\bfz_0, \bfz_*; \bfxi).
\end{equation}
We choose the mapping $\bfG$ to be a two-layer neural network with parameters $\bfxi$, and we initially feed it with approximations to $\bfz_*$ and $\bfz_0$. For $\bfz_*$, one simple approximation is to solve the data fitting problem with $\bfz_N=0$, as we also use in Algorithm \ref{alg:ipsolve} for initialization. For simplicity, we use the same for $\bfz_0$, and learn to predict $\bfz_1$. Other choices are worthy of consideration.

Given $\bfz_1$ we can replace
the solution of the nonlinear BVP \eqref{bvp} with a forward propagating network, that is, with an {\em initial value problem} (IVP). The idea of replacing BVP with an IVP is not new and is the backbone of many so-called "shooting methods" \cite{ascherBook}.
Here {we} follow an approach similar to \cite{haber2022estimating} to solve the shooting problem. We learn the (nonlinear) mapping that maps the terminal condition to
the initial condition. Putting it all together, we obtain a simple direct algorithm, presented in Algorithm \ref{alg:ipsolveDirect}, which we refer to as hyperbolic ResNet or Hyper-ResNet.

\begin{algorithm}
\begin{algorithmic}
\caption{Hyper-ResNet - Direct solution of the regularized inverse problem}
\label{alg:ipsolveDirect}
\State Initialize $\bfz_0  = \bfz_* = (\bfE^{\top} \bfA^T \bfA \bfE + \alpha \bfI)^{-1} \bfE^{\top} \bfA^{\top} \bfb$

\FOR{$t=1,\ldots,maxIter$}
   \STATE \textit{\# Approximate Least Action by Shooting:}
\State Estimate $\bfz_1$ using $\bfz_1 = G(\bfz_0, \bfz_*; \bfxi)$
    
\State Propagate for $\ell=1,...,N-1$: $\bfz_{\ell+1} = 2\bfz_{\ell} - \bfz_{\ell-1} - \grad \phi(\bfz_{\ell}, \bftheta_{\ell})$   
   \STATE \textit{\# Solve Data Fit:}

\State $\bfz_* = (\bfE^{\top} \bfA^{\top} \bfA \bfE + \alpha \bfI)^{-1} (\bfE^{\top} \bfA^{\top} \bfb + \alpha \bfz_N)$
\ENDFOR
\STATE \textbf{return} $\bfz_*$.

\end{algorithmic}
\end{algorithm}

However, when replacing the BVP with an IVP we end up with two different, potentially contradicting, 
equations for the final state $\bfz_*$. These equations read
\begin{subequations}
\label{bvpend}
\begin{eqnarray}
&& 2 \bfz_N - \bfz_* - \bfz_{N-1} - \grad \phi(\bfz_N, \bftheta_N) = 0 \\
&& (\bfE^{\top} \bfA^{\top} \bfA \bfE + \alpha \bfI)\bfz_* =  \bfE^{\top} \bfA^{\top} \bfb + \alpha \bfz_N(\bfz_1) 
\end{eqnarray}
\end{subequations}
The first equation is the last step of the network, while the second equation is the data fitting, which by definition will hold after the last step of Algorithm \ref{alg:ipsolveDirect}.
Here, we write $\bfz_N(\bfz_1)$ to emphasize its dependence on the (learned) initial conditions.
Clearly, these equations may not be consistent; however, for an appropriate initialization of $\bfz_1$ they agree. That is, the eventual residual of the shooting process defined by 
\begin{equation}
\label{eq:residual}
\bfr_s = 2 \bfz_N - \bfz - \bfz_{N-1} - \grad \phi(\bfz_N, \bftheta_N),
\end{equation}
is 0. Therefore, in the training process of the network in \eqref{netz1} that map $\bfz_*$ and $\bfz_0$ to $\bfz_1$ we aim to yield an output $\bfz_1$ such that the residual in \eqref{eq:residual} is sufficiently small, and both equations agree at the terminal condition. 
This is done by outputting $\bfr_s$ and minimizing it throughout the learning process of the parameters of Algorithm \ref{alg:ipsolveDirect}.

\subsection{Interpretation}
We now further interpret the proposed method in Algorithm \ref{alg:ipsolveDirect}, using projection techniques.
To this end, consider the final state of the network $\bfz_N(\bftheta;\bfb)$ that is computed. Such an estimate does not
necessarily fit the data. Therefore, the fourth step in Algorithm \ref{alg:ipsolveDirect} can be interpreted as a projection that is used to fit the data. This step is equivalent to the solution of the optimization problem
\begin{eqnarray}
\label{Reg}
\min_{{{{\bfz}}}} && R(\bfz, \bftheta) = \hf \|\bfz -  \bfz_N(\bftheta; \bfb)\|^2 \\
{\rm s.t.} && \|\bfA \bfE\bfz - \bfb \|^2 \le \tau,
\end{eqnarray}
where the choice of the parameter $\tau$ here corresponds to the choice of $\alpha$. This view gives a different interpretation of the algorithm.
The first part, computing $\bfz_N$, requires evaluating or exposing the components that are in the null space of $\bfA \bfE$.
The null space components introduced by $\bfz_N$ are not changed 
when projecting the solution into the data fit. Thus, the goal of the network is to estimate the null space of the solution. 
Since the null-space components of the solution are not directly available, the data $\bfb$ is needed in order to "hint" the network at what null-space components are to be reproduced.

\begin{figure}[h]
    \centering
    \includegraphics[]{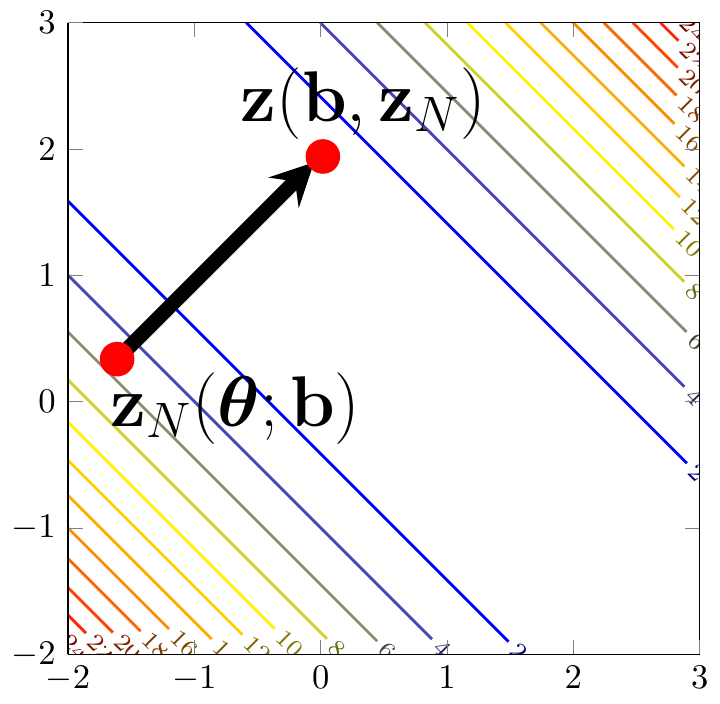}
\caption{The two steps of Algorithm \ref{alg:ipsolveDirect} for the solution of the linear inverse problem
$z_1 + z_2 = 1$. In the first step, a network is used to obtain $\bfz_N(\bfb)$ that does not necessarily fit the data. In the second step, we project the solution to fit the data using $\bfz_N$ as a reference.}
    \label{fig1}
\end{figure}

To demonstrate that we consider a very simple example. Let 
$$
 \bfA = \begin{bmatrix} 1 & 1\end{bmatrix} \quad {\rm and} \quad   \bfE = \begin{bmatrix}  1  & 1 \\ 1  & -1 
\end{bmatrix}
$$
For this simple choice, we note that the second column in $\bfE$ is in the null space of $\bfA$.
The forward problem then becomes
$$ \bfA \bfE \bfz = \begin{bmatrix} 2 & 0 \end{bmatrix} \begin{bmatrix}  z_1   \\ z_2 
\end{bmatrix} = \bfd $$
Therefore, the second entry of $\bfz$ cannot be 
determined by the data because the data has no direct information about the second component of $\bfz$.
Standard minimum norm solutions choose $z_2 = 0$. However, if we are given training data, we may find a connection between $z_2$ and $z_1$ that is estimated from $\bfd$. This connection fills in the missing equations for $z_2$, which in our case is introduced to \eqref{Reg} by $\bfz_N$ when solving the regularized problem that fits the data. The two-step process, the first that estimates the null space and the second that projects the solution to fit the data is demonstrated in Figure~\ref{fig1}.

Equipped with this interpretation we note that it is easy to obtain different variants of the algorithm. For example, it is easy to modify the projection step such that it involves a different norm, for example, penalize the $1$-norm between $\bfz$ and $\bfz_N$ that will yield a sparse solution for the difference. It is also possible to use different {ways} to measure the projection to the data. For example, in the presence of noise with non-uniform standard deviation (a common case in electromagnetic imaging) it is easy to add a covariance term to the projection or use M-estimators \cite{tyler1987distribution,friedman2001elements}.

\section{Training DRIP}
\label{sec4}
We now discuss the data generation and training procedures that are used in this work toward the numerical experiments shown in the next section.

\paragraph{Problem Settings.}
We assume a linear forward model as in \eqref{forlin}, for which we assume to have a set of plausible realizations, 
$\{\bfu^{(j)}\}_{j=1}^{m}$. In order to train the network we generate data from these realizations by applying the forward operator $\bfA$ and adding noise. This yields
$m$ pairs of data $\{ (\bfu^{(j)}, \bfb^{(j)}) \}_{j=1}^{m}$
that we can learn from.

\paragraph{Optimization and Loss Functions}

We consider several loss functions in the optimization process of our proposed network.
First, we consider the minimization of the error term:\footnote{Note that the superscript $^{(j)}$ in this case represents different realizations of $\bfu$}.
\begin{equation}
    \mathcal{L}_{error} = \frac{1}{m}\sum_{j=1}^{m}\|\bfu_*^{(j)} - \bfu^{(j)} \|_2^2,
\end{equation}
where $\bfu_*^{(j)} = \bfE\bfz_*^{(j)}$ is the obtained approximate solution for the $j$-th problem. The next loss describes the data fitness, i.e., the residual term:
\begin{equation}
    \mathcal{L}_{residual} = \frac{1}{m}\sum_{j=1}^{m}\| A\bfu^{(j)}_* - A\bfu^{(j)} \|_2^2 + \bfr_s.
\end{equation}
where $\bfr_s$ is the shooting error defined in Equation~\eqref{eq:residual}.

The third and final considered loss demands the similarity between the initial guess that stems from a simple data fitting procedure, and the solution predicted by our network. In our experiments, we found that adding this loss promotes the network toward yielding a solution that corrects the initial guess rather than significantly altering it. Formally, this loss is defined as:
\begin{equation}
    \label{eq:xrefLoss}
    \mathcal{L}_{sim} = \frac{1}{m}\sum_{j=1}^{m}\| \bfu^{(j)}_* - \tilde{\bfu}^{(j)} \|_2^2.
\end{equation}
where $\tilde{\bfu}^{(j)}$ is obtained by a simple data fitting projection (i.e., our initial guess in Algorithms \eqref{alg:ipsolve} and \eqref{alg:ipsolveDirect}). In total, our loss function is given by the possibly weighted sum of the described losses, as follows:
\begin{equation}
    \label{eq:totalLoss}
    \mathcal{L} = \mathcal{L}_{error} + \alpha \mathcal{L}_{residual} + \beta \mathcal{L}_{sim},
\end{equation}
where $\alpha$ and $\beta$ are hyper-parameters. In our experiments, we found that setting $\alpha=1, \ \beta=0.1$ worked well. 

The overall optimization procedure is summarized in Algorithm~\ref{alg:train}. We used the Adam \cite{kingma2014adam} optimizer to minimize our loss functions (discussed next), with an initial learning rate of $10^{-3}$ and a scheduling policy that multiplies the learning rate by a factor of 0.8 after every 20 epochs, for a total of 250 epochs. We use a weight decay factor of $10^{-4}$. All our experiments are run on an Nvidia RTX-3090 with a total memory of 24GB.  

\begin{algorithm}
\caption{A Training Epoch of DRIP}\label{alg:train}
\begin{algorithmic}
\FOR{$j=1,\ldots, M$}
    \STATE Load  $\bfu^{(j)}$ and compute $ \bfb^{(j)} = \bfA \bfu^{(j)} + \epsilon^{(j)}$, where $\epsilon^{(j)}$ is a randomly generated noise.
\STATE Use Algorithm \ref{alg:ipsolve} or \ref{alg:ipsolveDirect}
to evaluate $\bfz$ and $\bfr_s$.
\STATE  Compute the loss $\mathcal{L}$ from Equation \eqref{eq:totalLoss}.  

\STATE Compute the gradient $\grad_{\bftheta} \mathcal{L}$ and update the weights $\bftheta$ using the Adam optimizer.

\ENDFOR

\end{algorithmic}
\end{algorithm}

\section{Numerical Experiments}
\label{sec5}

In this section, we experiment with our algorithm, solving two different linear inverse problems. In what follows, we describe each forward problem, and then present and discuss the obtained results using baseline models and our DRIP approach. {We consider three baseline models. The first, is Neural-PGD \cite{mardani2018neural}, which utilizes a recurrent neural network (RNN) that learns an approximation of a neural gradient, applied several times (8 in our experiments). The backbone of the RNN is composed of a 5-layer ResNet \cite{he2016deep} network. Furthermore, we consider a formulation as discussed in Section \ref{sec2} and specifically in Equation \eqref{net1}, where we experiment with $f$ being a ResNet (also with 5 layers), or a UNet \cite{ronneberger2015u}.}
{Throughout this section, all plots are logarithmic, both in x and y axes. 
}
{
Plots that report the 'Residual' and 'Error' values consider the relative data and model fitness, respectively.  That is:
\begin{equation}
    \rm{Residual} = \frac{\| \bfA \bfE \bfz - \bfb \|_2}{\|\bfb\|_2} , \quad     \rm{Error} = \frac{\| \bfx_{\rm{pred}} - \bfx_{\rm{true}} \|_2}{\|\bfx_{\rm{true}}\|_2}.
\end{equation}
}

\subsection{Forward Problems}
\label{subsec:forwardProblems}

\subsubsection{Image Deblurring.}

We follow the formulation and code presented in \cite{nagyHansenBook}
 for the image deblurring problem. The mathematical model in this case is a simple convolution
\begin{eqnarray}
\label{blur}
{
b(x_1,x_2) = \int_{\Omega} K(x_1-\xi_1, x_2-\xi_2) u(\xi_1, \xi_2) d\xi_1 d\xi_2}
\end{eqnarray}
where $K(x_1,x_2)$ is a convolution kernel of the form.
\begin{equation}
            {K(x_1,x_2) = \exp\left(-\frac{x_1^2 + x_2^2}{2\sigma^2}\right),}
\end{equation} with $\sigma=2$.

Discretizing the integral \eqref{blur} on a regular mesh,
yields a linear system
$$ \bfA \bfu = \bfb $$
where $\bfA$ is a discretization of $K$ typically obtained by 2D fast Fourier transform.
The singular values of $\bfA$ used in our example are plotted in Figure~\ref{fig:svdBlur}.
It is well known that for problems where the singular values decay exponentially simple recovery methods fail and the role of
regularization becomes crucial \cite{Tenorio2011}.

\subsubsection{Limited Angle Computed Tomography.}

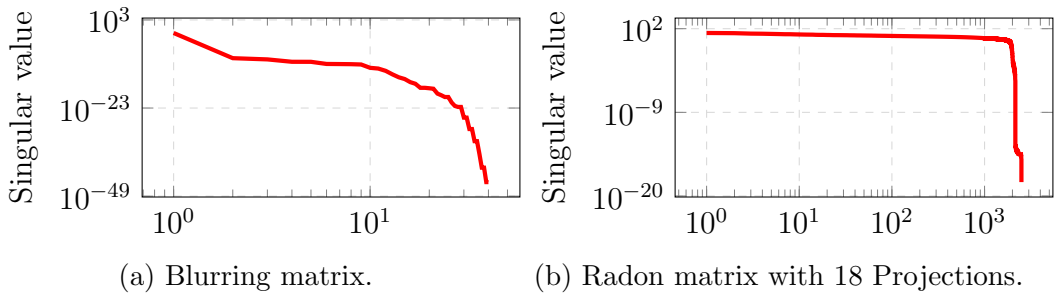
\begin{figure}
    \centering
        \begin{minipage}{.40\textwidth}
\centering
\begin{tikzpicture}[define rgb/.code={\definecolor{mycolor}{RGB}{#1}},
                    rgb color/.style={define rgb={#1},mycolor}]
  \begin{axis}[
      width=1.0\linewidth, 
      height=0.6\linewidth,
      grid=major,
      grid style={dashed,gray!30},
      %xlabel=,
      ylabel=Singular value,
      ylabel near ticks,
      legend style={at={(10,1.3)},anchor=north,scale=1.0, draw=none, cells={anchor=west}, font=\tiny, fill=none},
      legend columns=-1,
      yticklabel style={
        /pgf/number format/fixed,
        /pgf/number format/precision=3
      },
      scaled y ticks=false,
      every axis plot post/.style={ultra thick},
      xmode=log,
      ymode=log
    ]
    \addplot[color=red]
    table[x=entry,y=singval,col sep=comma] {data/blur_singvals.csv};
    \addlegendimage{}
    \addlegendimage{}
    \addlegendimage{}
    \addlegendentry{}
    \end{axis}
\end{tikzpicture}
\subcaption{Blurring matrix.}\label{fig:svdBlur}
\end{minipage} \hspace{0.5em}
    \begin{minipage}{.40\textwidth}
\centering
\begin{tikzpicture}[define rgb/.code={\definecolor{mycolor}{RGB}{#1}},
                    rgb color/.style={define rgb={#1},mycolor}]
  \begin{axis}[
      width=1.0\linewidth, 
      height=0.6\linewidth,
      grid=major,
      grid style={dashed,gray!30},
      %xlabel=,
      ylabel=Singular value,
      ylabel near ticks,
      legend style={at={(10,1.3)},anchor=north,scale=1.0, draw=none, cells={anchor=west}, font=\tiny, fill=none},
      legend columns=-1,
      yticklabel style={
        /pgf/number format/fixed,
        /pgf/number format/precision=3
      },
      scaled y ticks=false,
      every axis plot post/.style={ultra thick},
      xmode=log,
      ymode=log
    ]
    \addplot[color=red]
    table[x=entry,y=singval,col sep=comma] {data/radon_singvals.csv};
    \addlegendimage{}
    \addlegendimage{}
    \addlegendimage{}
    \addlegendentry{}
    \end{axis}
\end{tikzpicture}
\subcaption{Radon matrix with 18 Projections.}\label{fig:svdRadon}
\end{minipage}
    \caption{The singular values of the blurring convolution matrix, and the Radon transform matrix 
    with 18 projections. Both problems exhibit exponential singular values decay, leading to a highly ill-posed deblurring problem.}
    
\end{figure}

A very common inverse problem is computed tomography. We consider its extreme and more ill-posed case: the limited-angle version.
Here, the data is obtained by projecting the model onto a line
(in 2D) using line integrals, i.e., the Radon transform (see \cite{natterer2001mathematics} for details).
Here, we experiment with a limited angle version that samples the space using 18 projections at different angles equally spaced between $[0, \pi]$. The singular values of
the projection operator are presented in Figure~\ref{fig:svdRadon}. It is evident that some of the singular values for the 18 projections are practically $0$ thus there is a significant null space to the matrix that needs to be recovered by our training models.

\subsection{Results}

We now present the obtained results with our DRIP approach. {For both the deblurring and tomography tasks,} we use the image collection from
STL-10 \cite{coates2011analysis}, containing 5000 and 8000 train and test images, respectively, of size $96\times 96$ pixels. {Additionally, we present tomographic reconstruction results with the OrganCMNIST from \cite{medmnistv2}. This dataset includes 13,000 training and 8,268 testing abdominal CT 2D images.} Below we illustrate three properties of our DRIP approach compared to the neural proximal method.

\subsubsection{Neural Proximal Regularization May not Fit the Data.}

First, we show that while it is possible to successfully train a 'general' neural network on given data, its behavior during
inference time can be unpredictable. To this end, similar to \cite{mardani2018neural}, as baseline architectures, we use the UNet \cite{ronneberger2015u}, ResNet \cite{he2016deep}, {and Neural-PGD \cite{mardani2018neural} networks}. To demonstrate our DRIP approach, we use the LA-Net and Hyper-ResNet architectures defined in Algorithms \ref{alg:ipsolve} and \ref{alg:ipsolveDirect}, respectively. Here, the chosen neural network populates $f$ 
in Equation~\eqref{net1}. We train the network for $250$ epochs with randomly sampled noise levels of 5\%-10\%.
After training, both the training and test splits yield data fit of roughly $10^{-2}$, which is in line with, or slightly higher than the noise level, regardless of the specific architecture. We show several tomographic reconstructions from the test set of STL-10 in Figure \ref{fig:tomography_results} {after training on the STL-10 training split, as well as results trained and tested using the OrganCMNIST in Figure \ref{fig:tomography_results}}, using the aforementioned architectures. From those results, we can say that when the test (inference) data is similar with respect to the noise regime of the training data, all architectures yield acceptable results.

However, as we decrease the noise levels during inference, for both of the considered tasks we see in Figures \ref{fig:residual_and_error_tomography}, \ref{fig:residual_and_error_tomography_CTorgan}, and \ref{fig:residual_and_error_deblur} that the data fitness (i.e., residual term) and the error term increase, when using a {UNet, ResNet, or a Neural-PGD}  network. To correct this, one may consider applying the trained networks several times, aiming to reduce the residual and error terms. Unfortunately, as can be seen from Figures \ref{fig:ood_iterations} and \ref{fig:residual_and_error_tomography_ood},
such an approach not only does not quantitatively improve the aforementioned terms, but rather also qualitatively worsens them as more iterations are added. 
We see that by increasing the number of iterations the data fit deteriorates. This phenomenon does not happen by coincidence. The neural proximal
network is not guaranteed to be a valid proximity operator of any optimization problem. Thus, more iterations do not necessarily lead to convergence.

As we show next, such problems are not observed with our proposed DRIP approach.

\subsubsection{DRIP Always Fit the Data.}

\begin{table}
    \centering
    \renewcommand{\arraystretch}{2}
    \begin{tabular}{|cccccc|}
    \hline 
     \tiny{Data} & \includegraphics[width=1.6cm, valign=c, trim={1cm 1cm 1cm 1cm},clip]{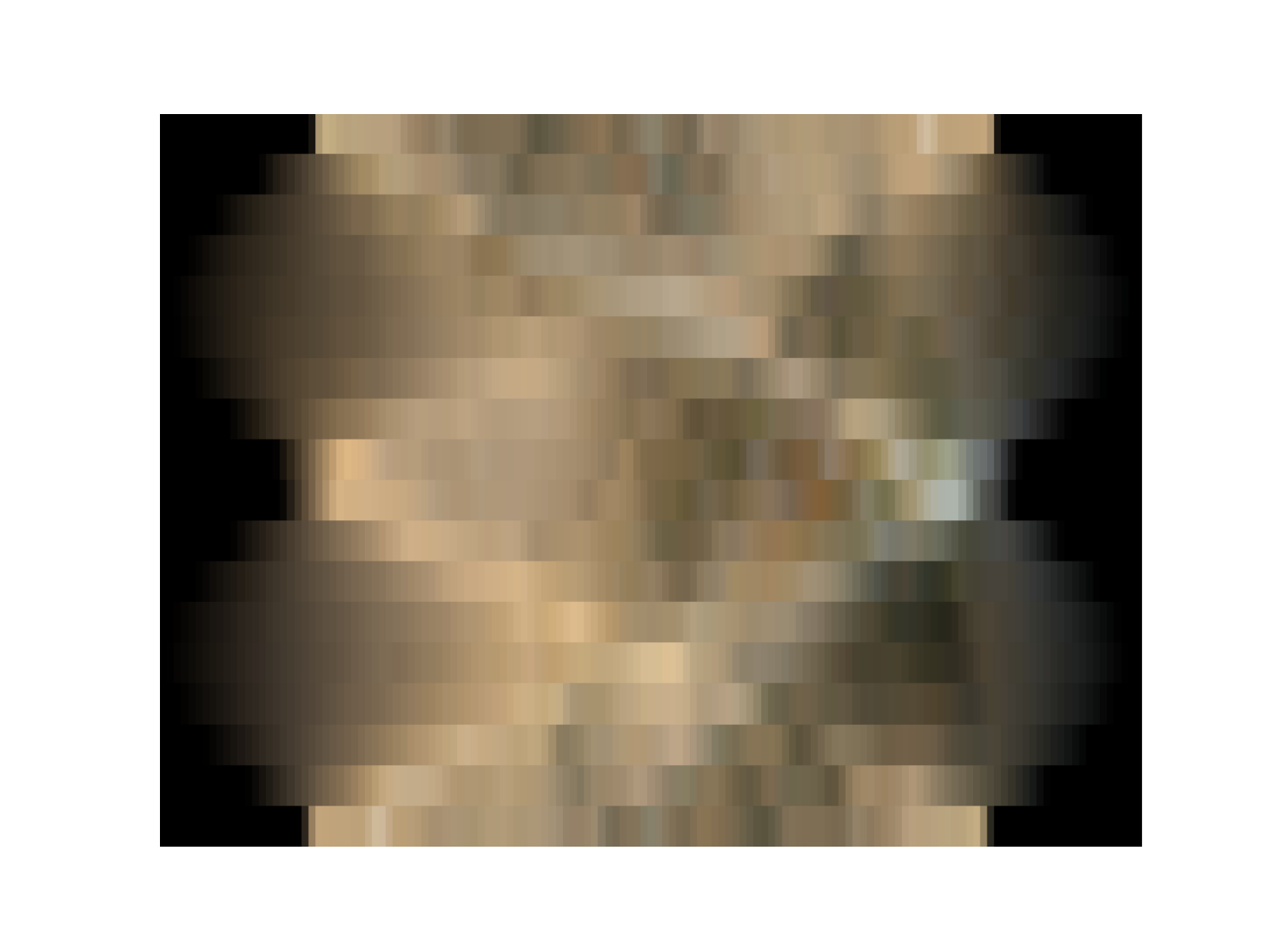} & \includegraphics[width=1.6cm, valign=c, trim={1cm 1cm 1cm 1cm},clip]{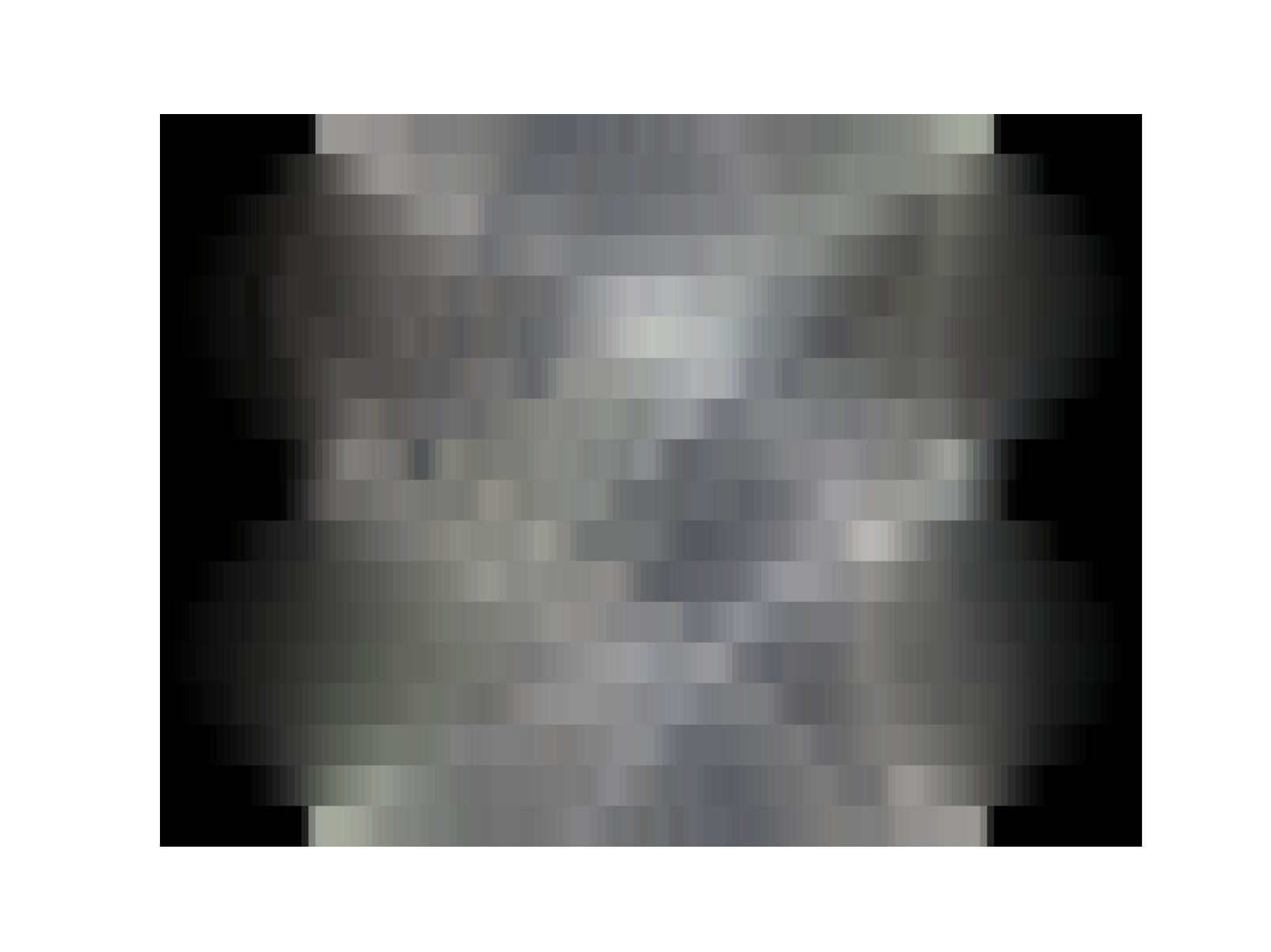} & \includegraphics[width=1.6cm, valign=c, trim={1cm 1cm 1cm 1cm},clip]{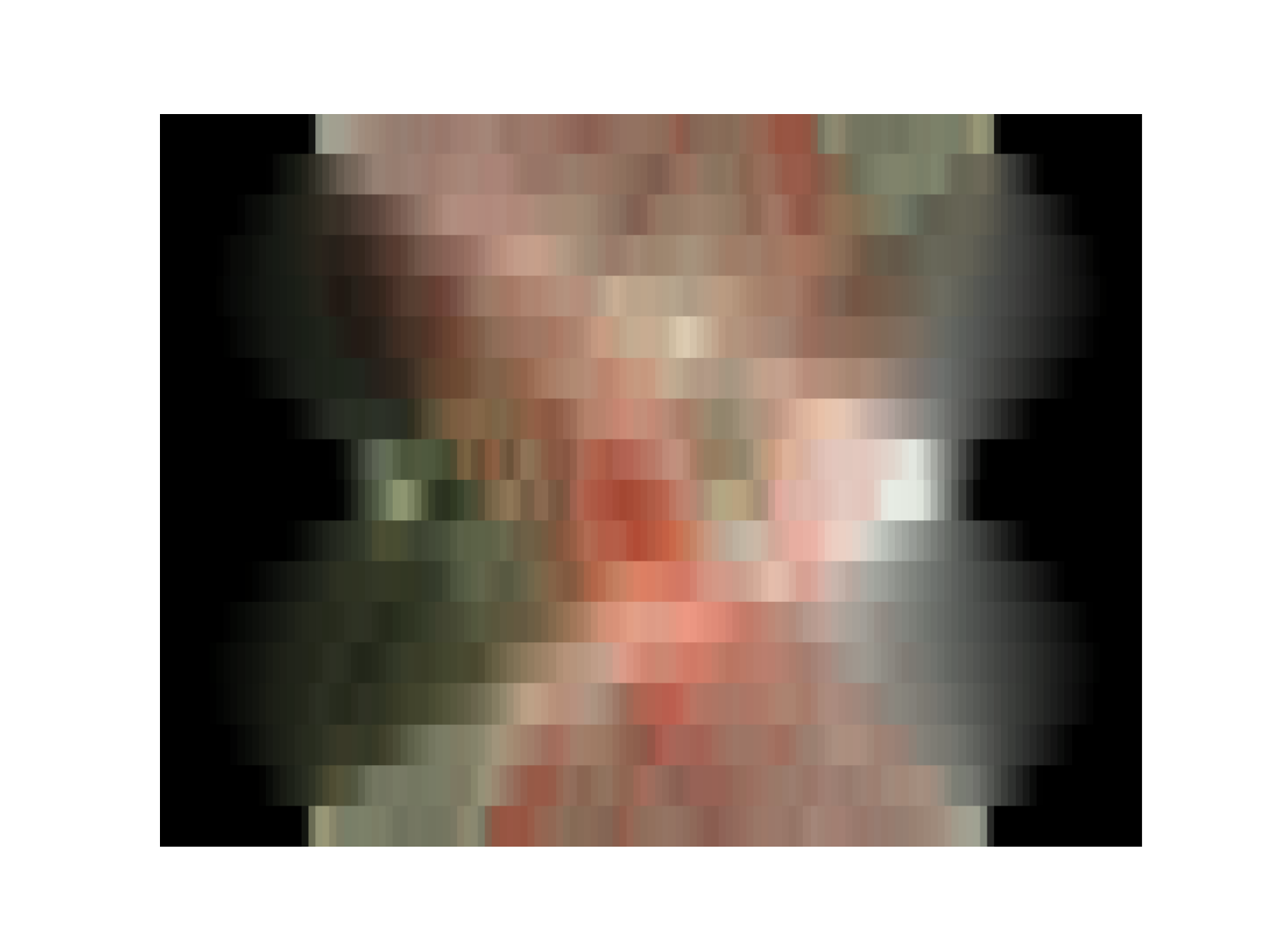} & \includegraphics[width=1.6cm, valign=c, trim={1cm 1cm 1cm 1cm},clip]{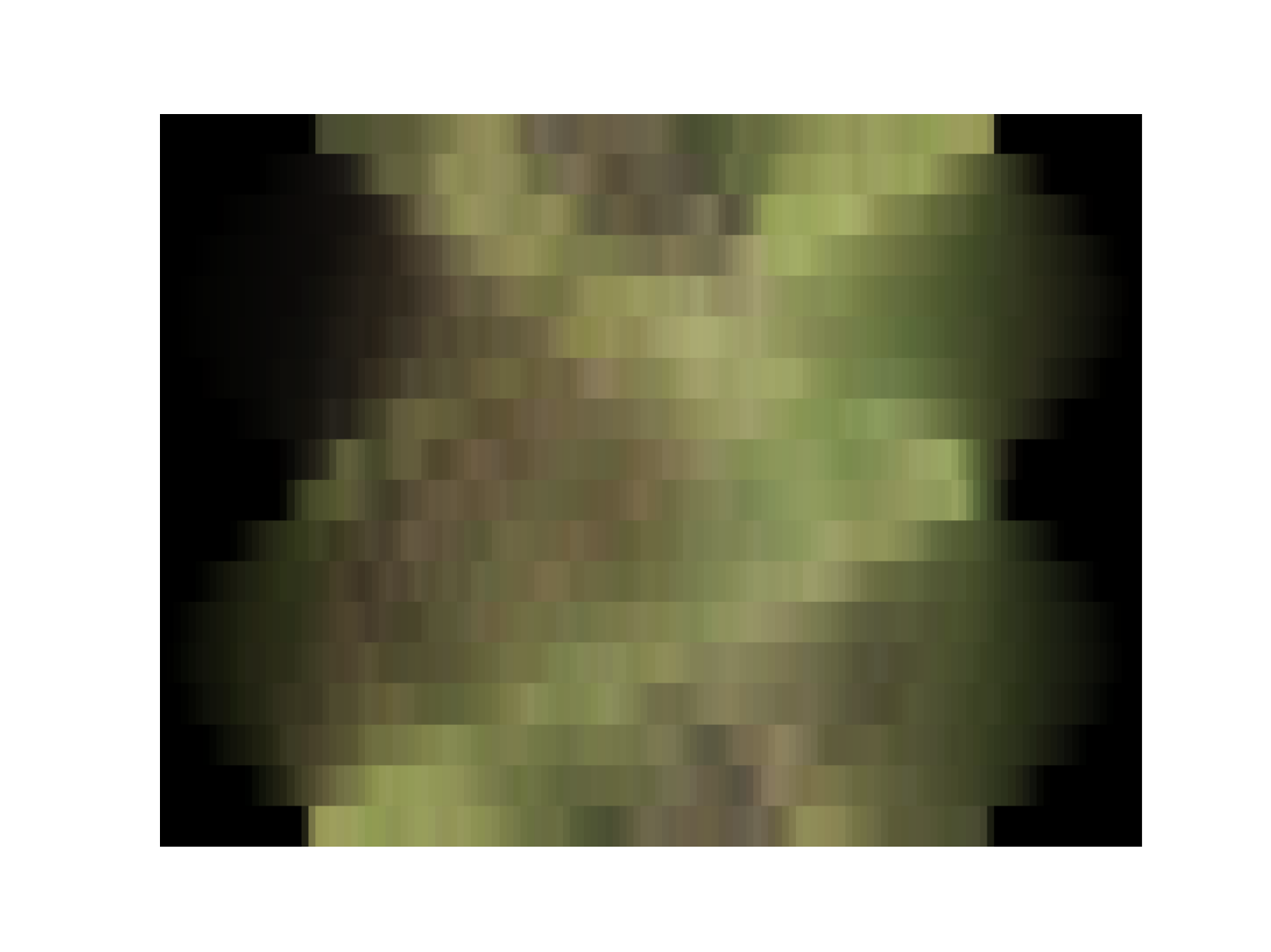} & \includegraphics[width=1.6cm, valign=c, trim={1cm 1cm 1cm 1cm},clip]{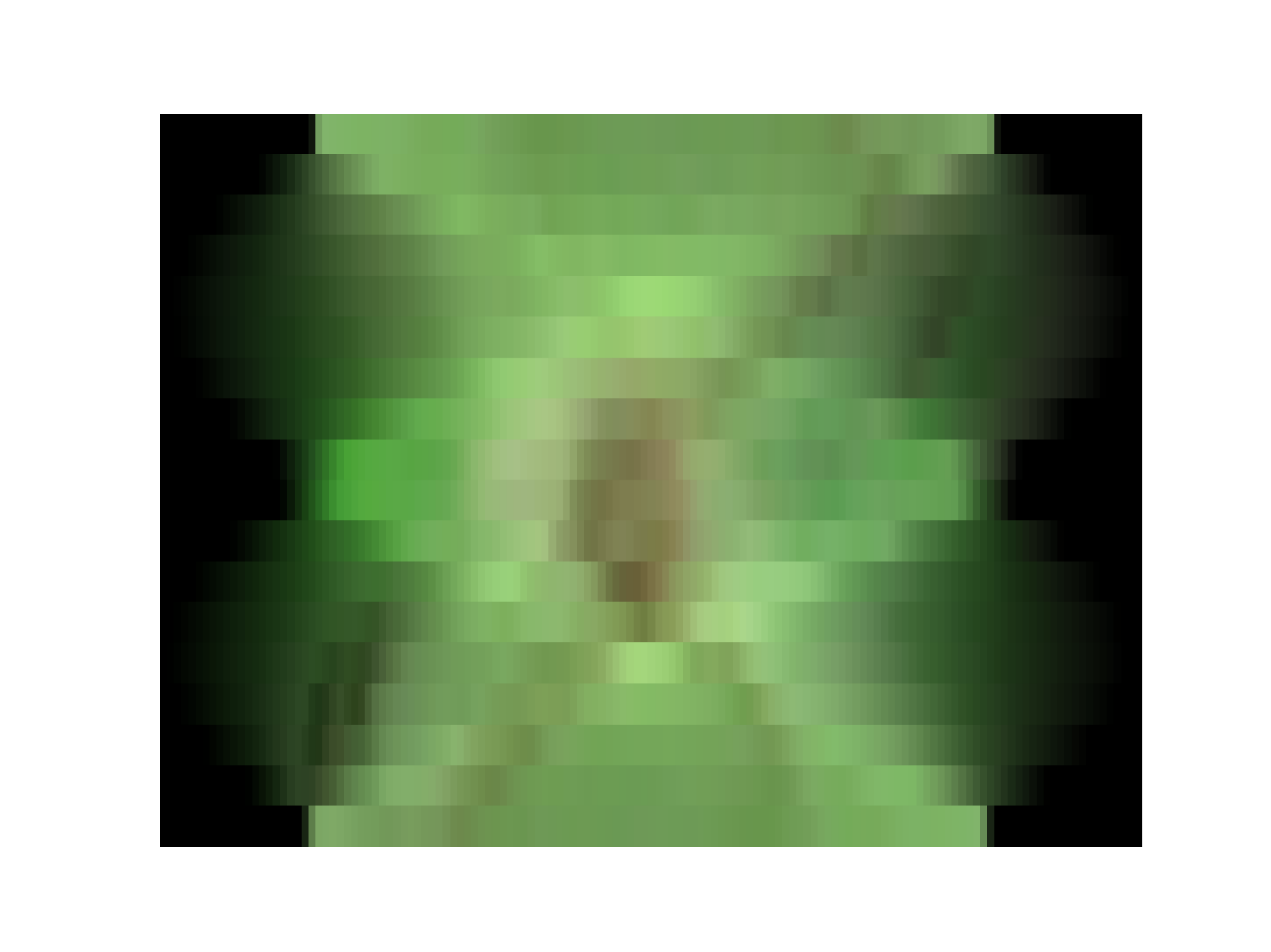} \\
     \hline
    \tiny{LA-Net} & 
    \includegraphics[width=1.6cm, valign=c, trim={1cm 1cm 1cm 1cm},clip]{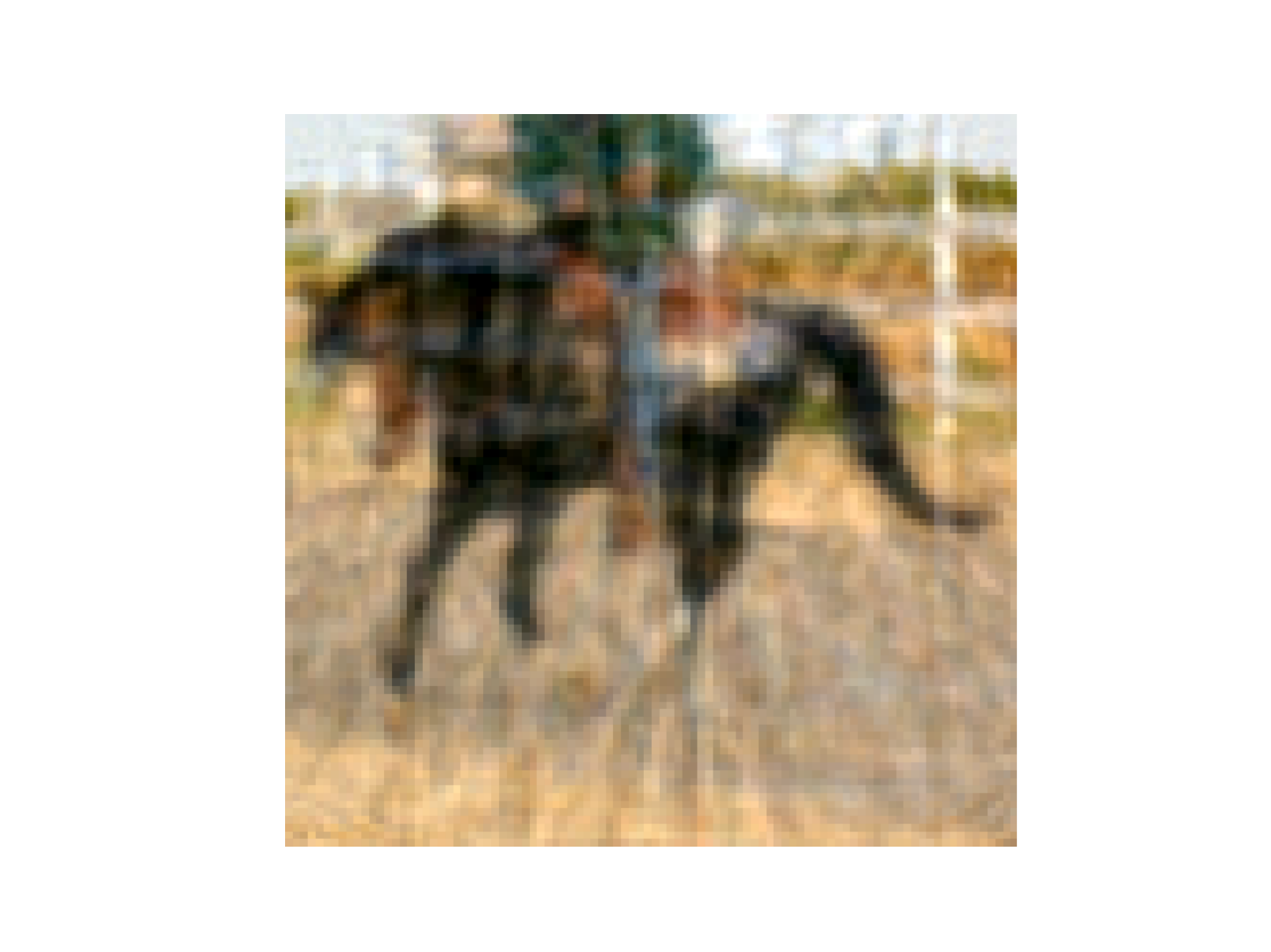}
     &
    \includegraphics[width=1.6cm, valign=c, trim={1cm 1cm 1cm 1cm},clip]{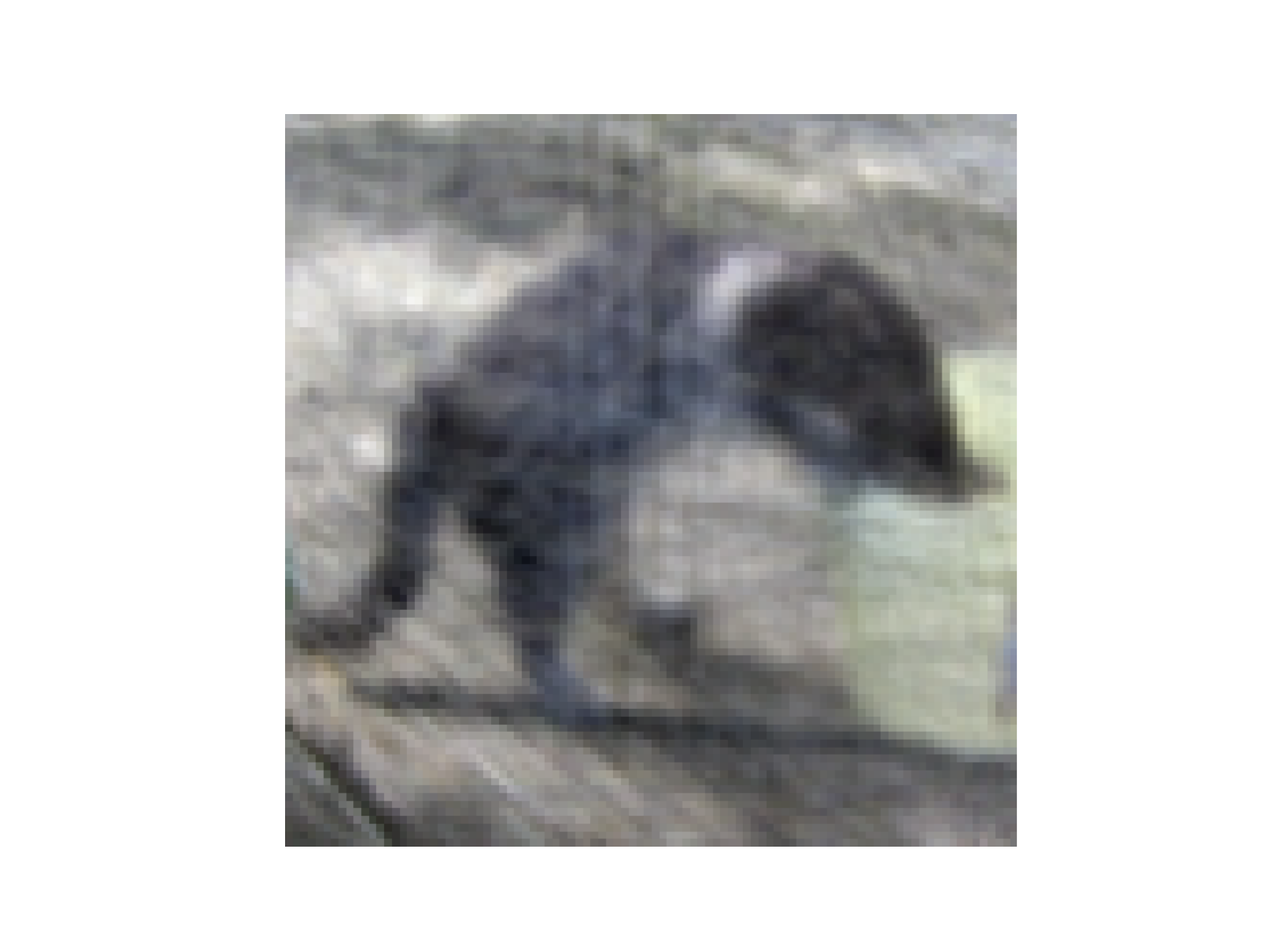} &
    \includegraphics[width=1.6cm, valign=c, trim={1cm 1cm 1cm 1cm},clip]{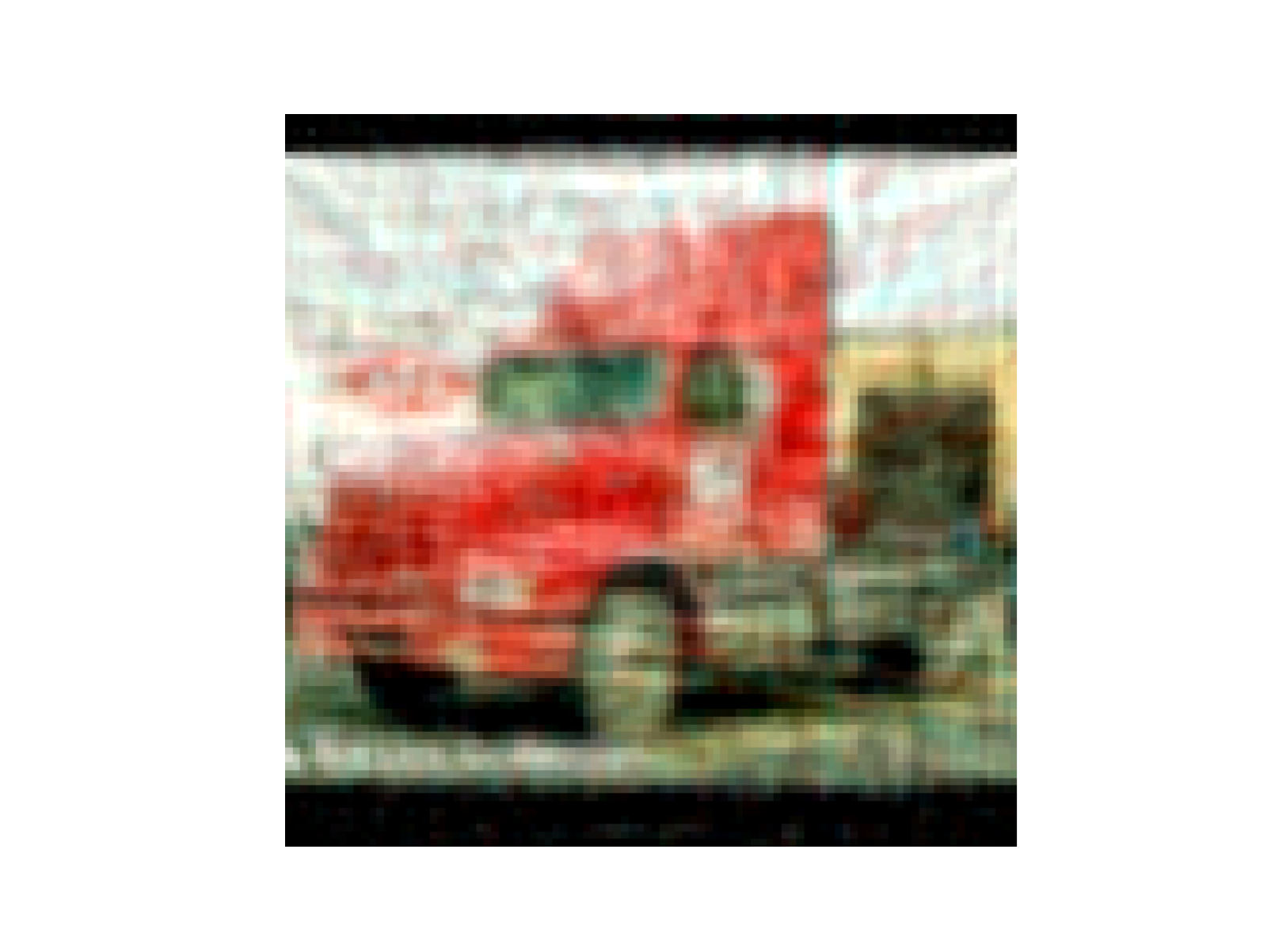} &
    \includegraphics[width=1.6cm, valign=c, trim={1cm 1cm 1cm 1cm},clip]{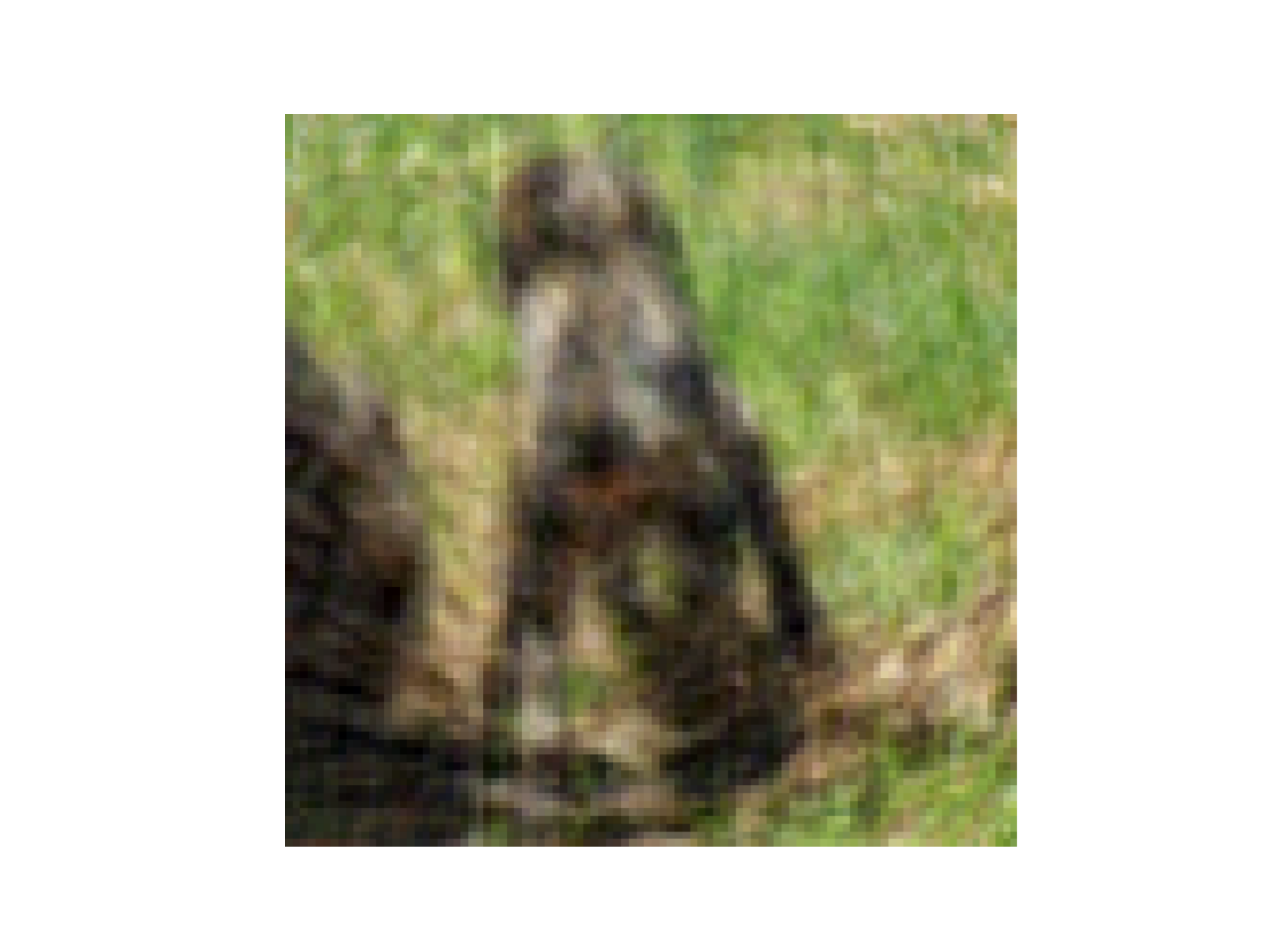} &
    \includegraphics[width=1.6cm, valign=c, trim={1cm 1cm 1cm 1cm},clip]{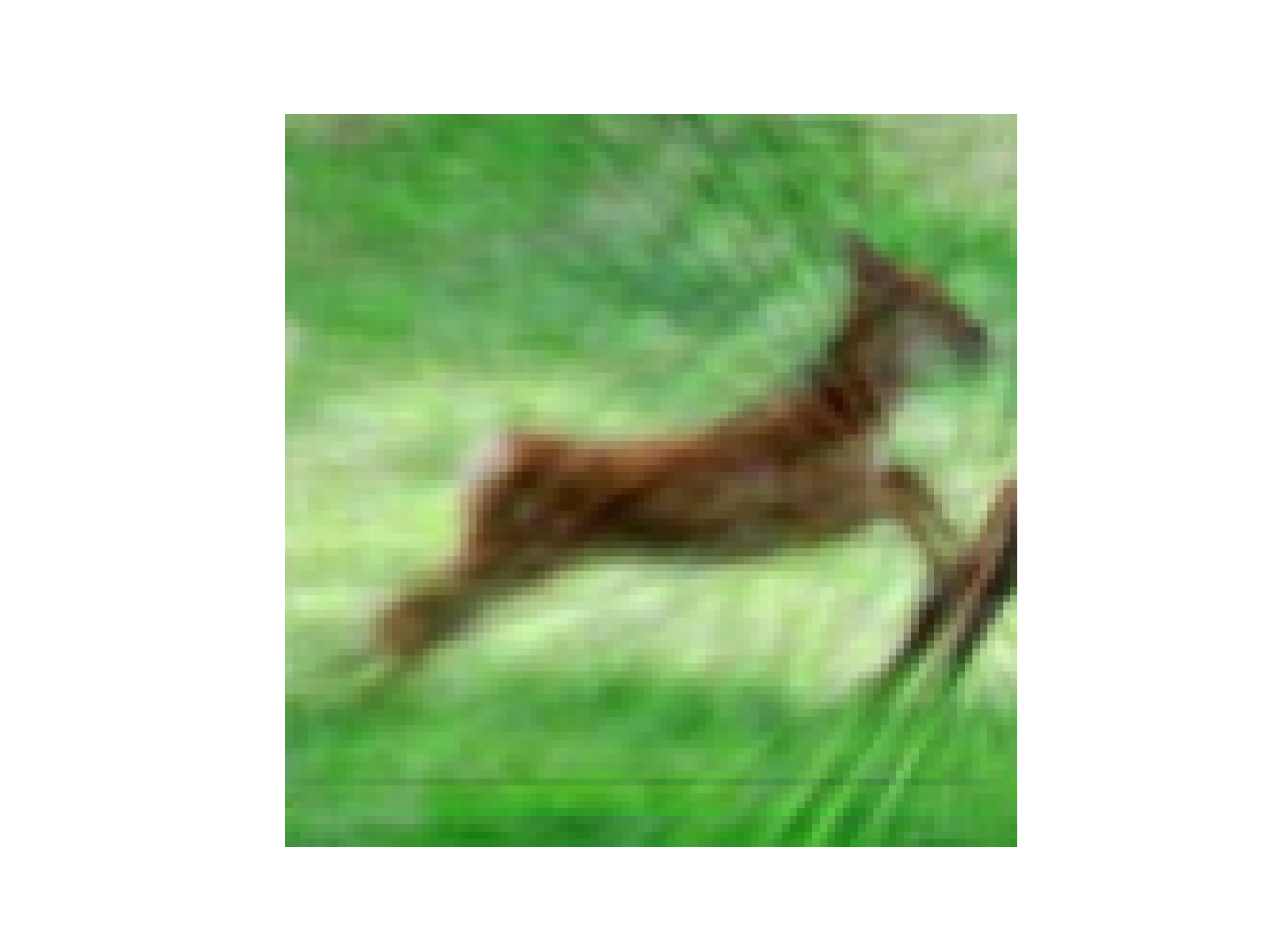}
    \\
        \hline
     
      \tiny{Hyper-ResNet} &  \includegraphics[width=1.6cm, valign=c, trim={1cm 1cm 1cm 1cm},clip]{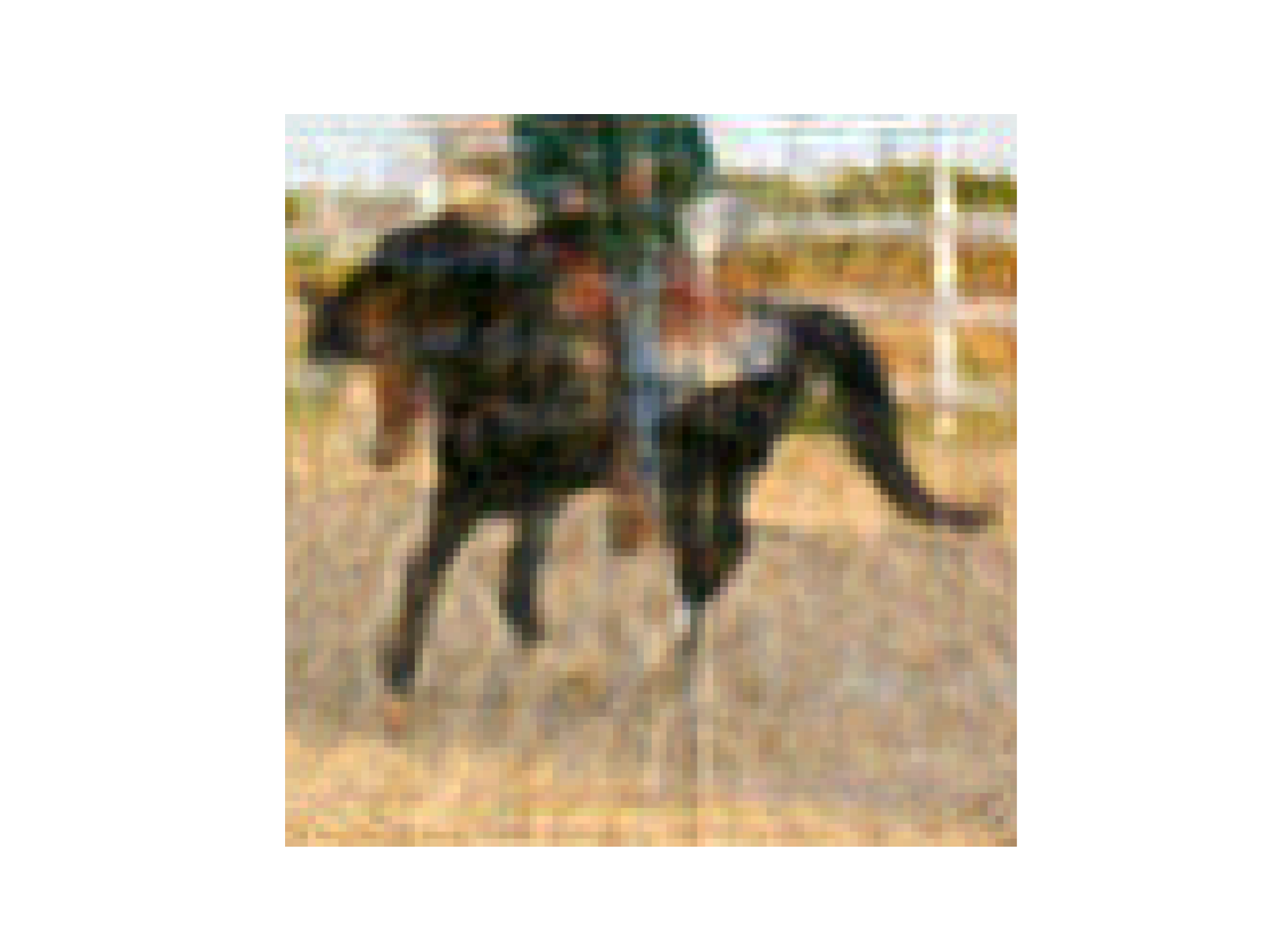}  &\includegraphics[width=1.6cm, valign=c, trim={1cm 1cm 1cm 1cm},clip]{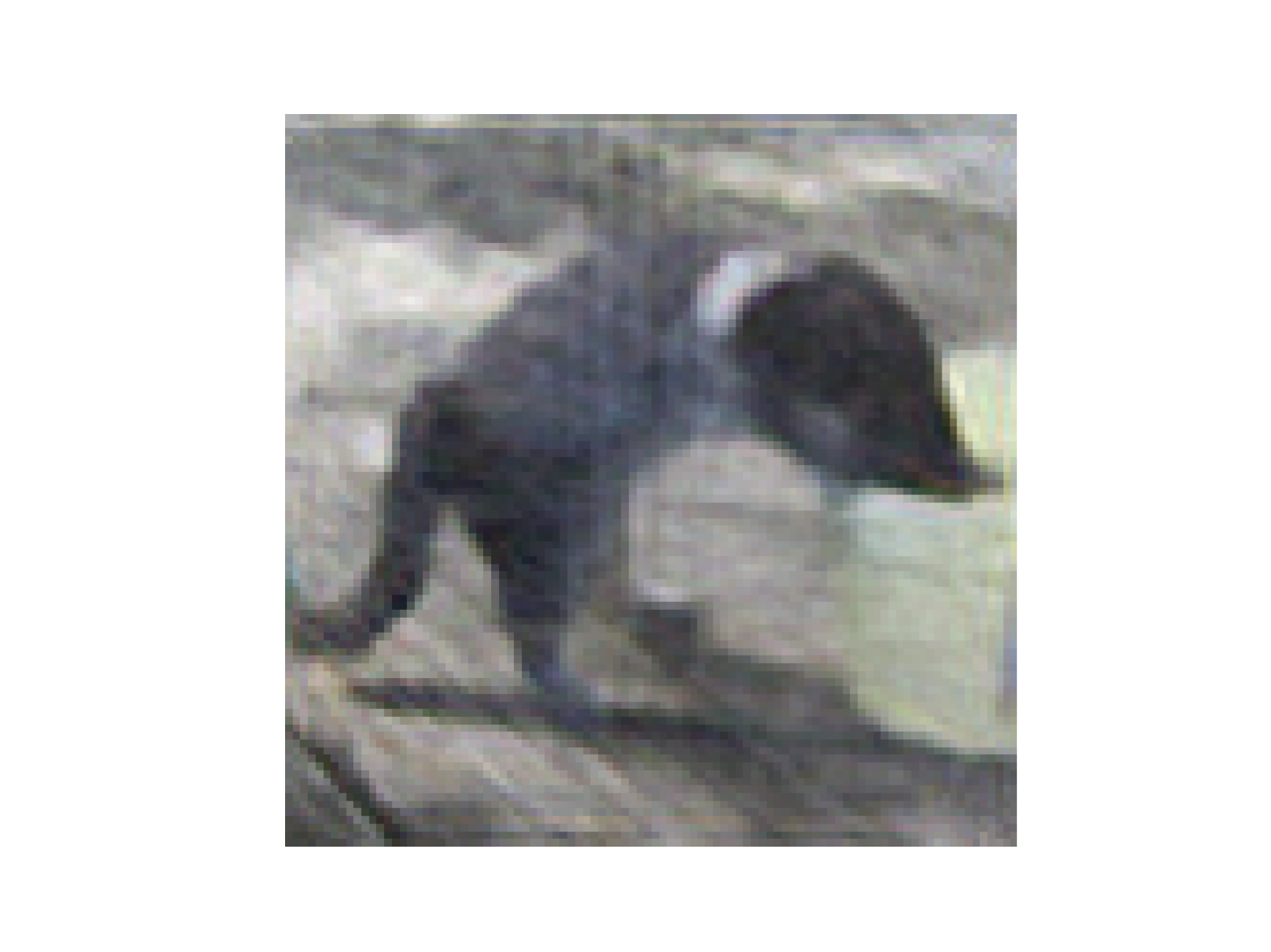} & \includegraphics[width=1.6cm, valign=c, trim={1cm 1cm 1cm 1cm},clip]{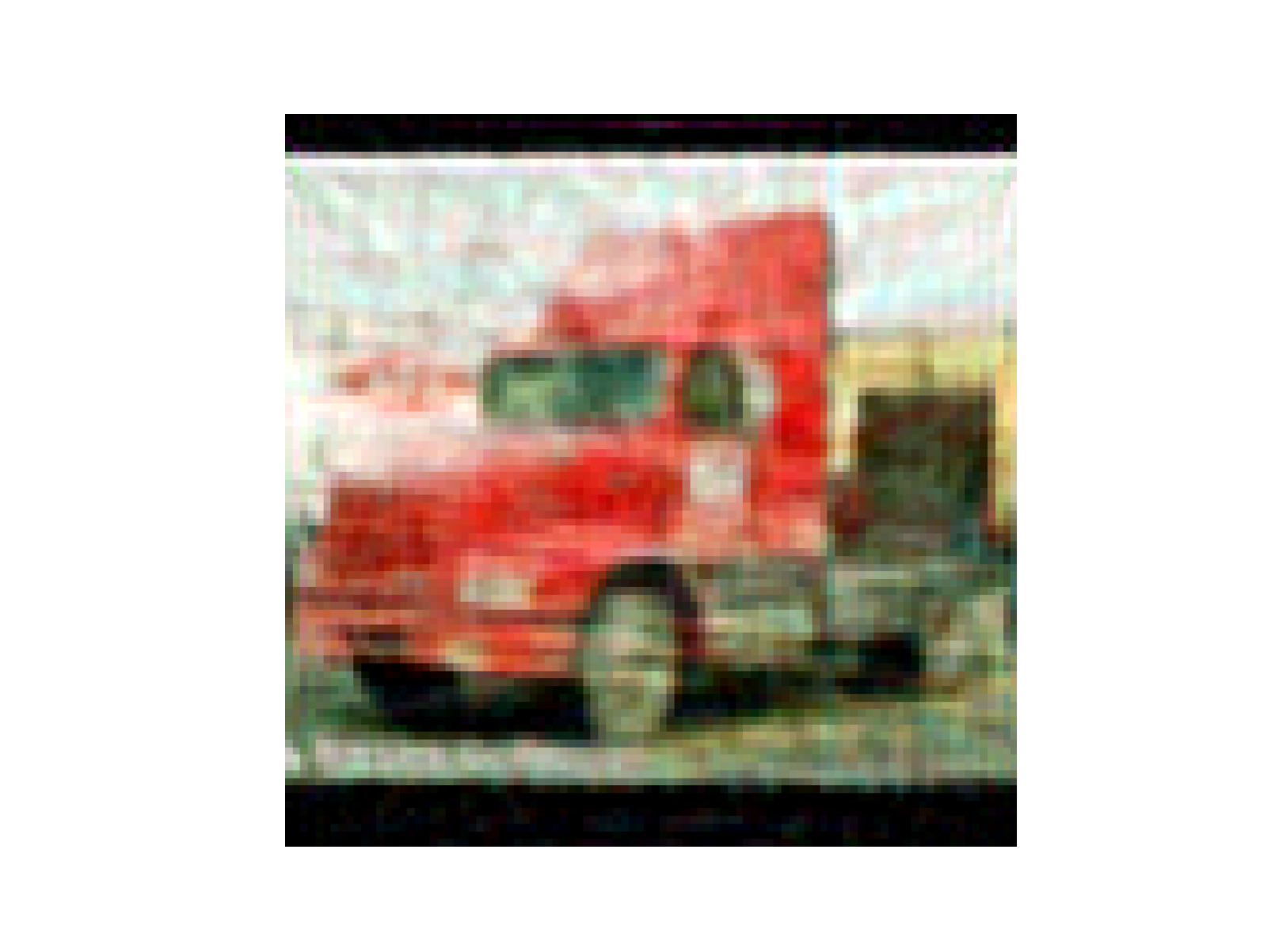} & \includegraphics[width=1.6cm, valign=c, trim={1cm 1cm 1cm 1cm},clip]{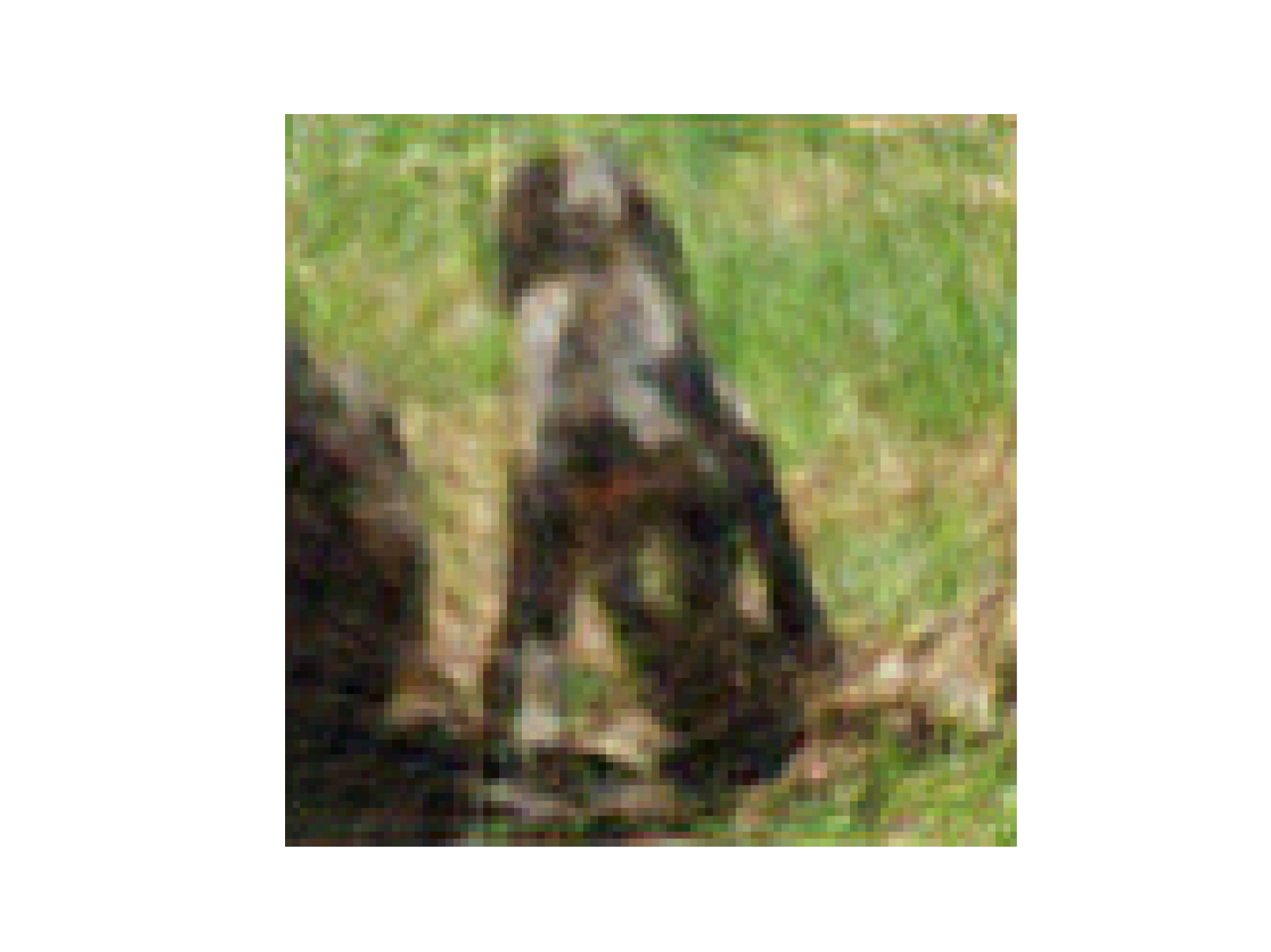} & \includegraphics[width=1.6cm, valign=c, trim={1cm 1cm 1cm 1cm},clip]{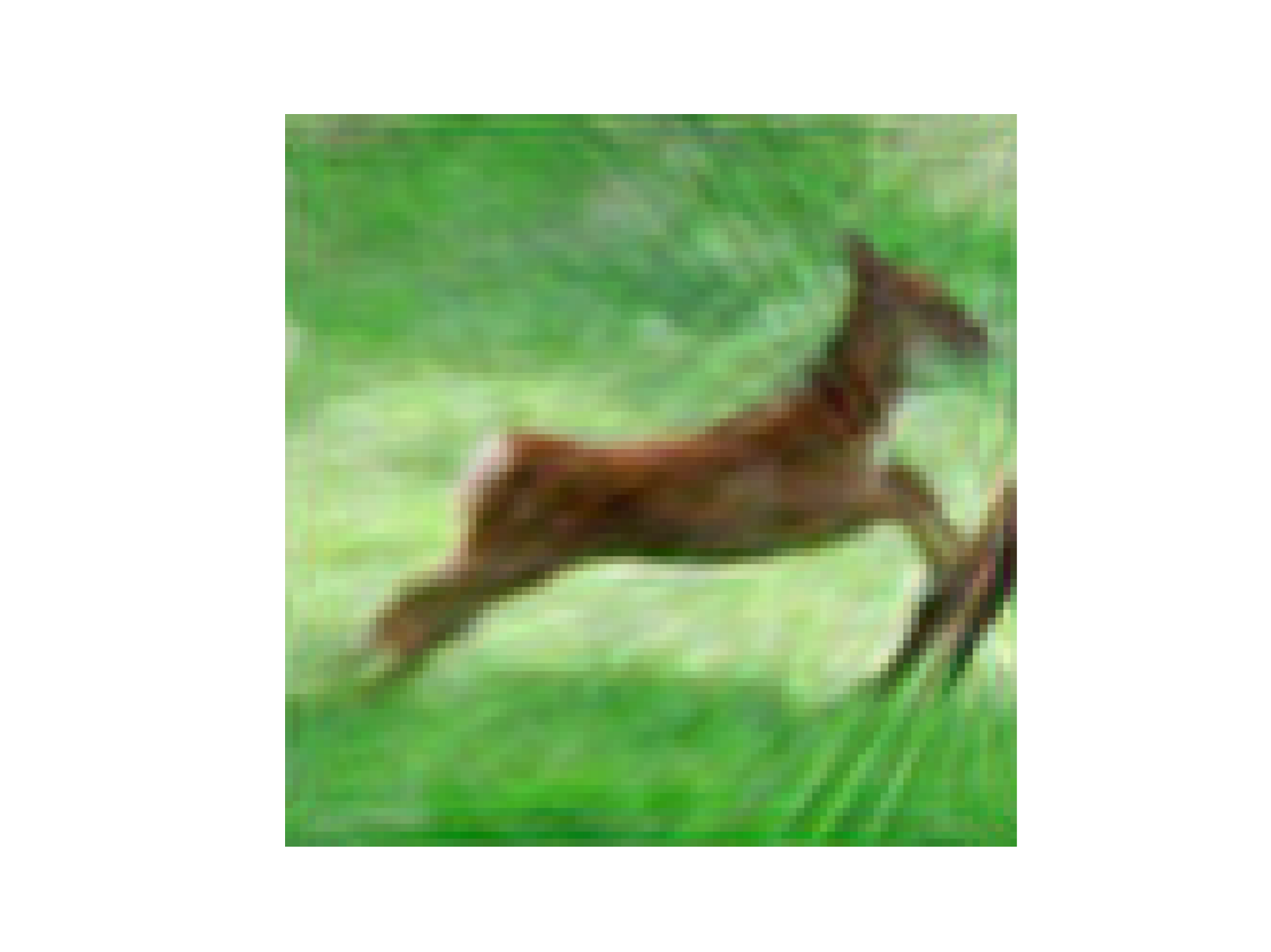}      \\
        \hline

            \tiny{{Neural-PGD}} & 
    \includegraphics[width=1.6cm, valign=c, trim={1cm 1cm 1cm 1cm},clip]{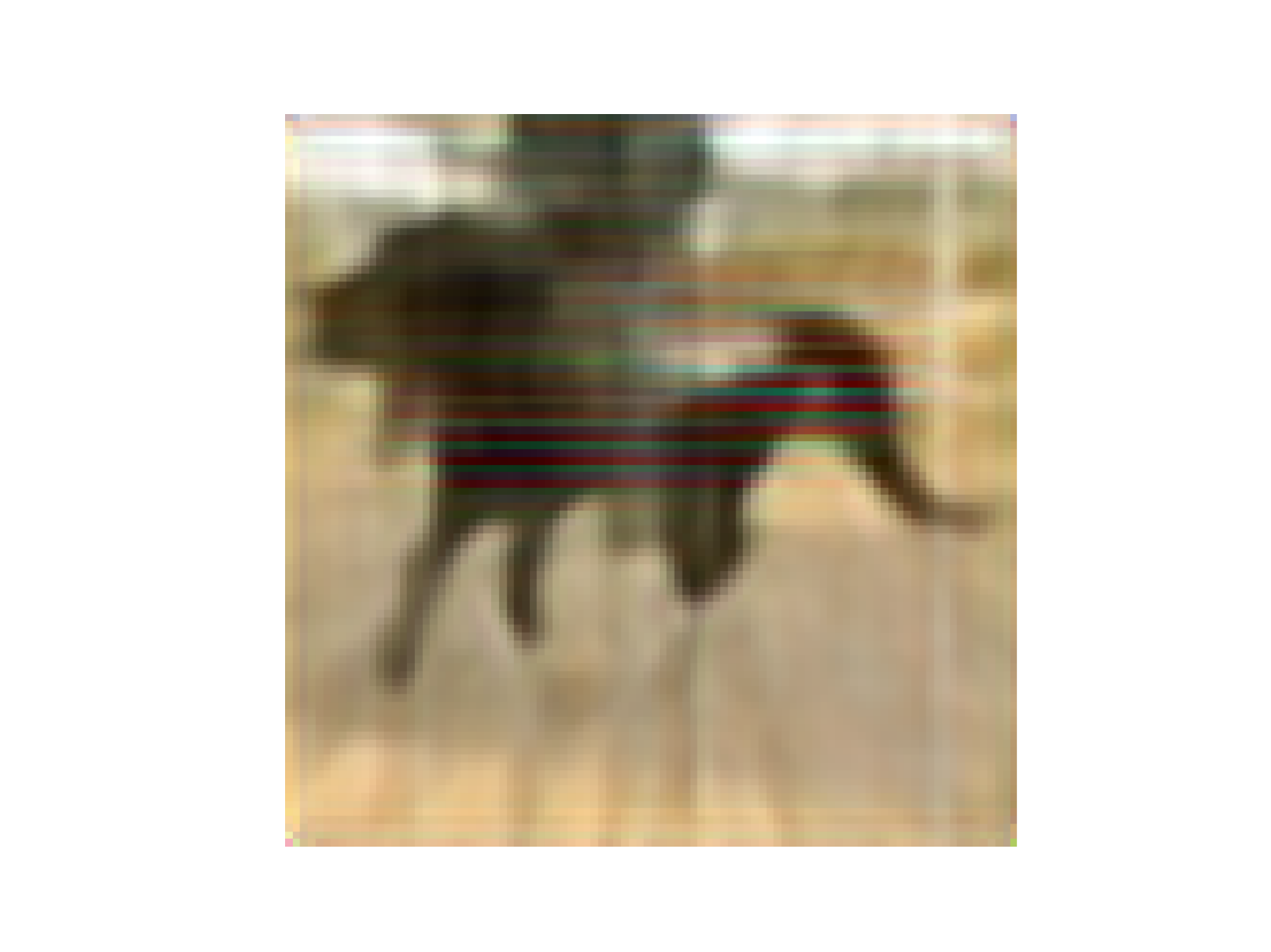}
     &
    \includegraphics[width=1.6cm, valign=c, trim={1cm 1cm 1cm 1cm},clip]{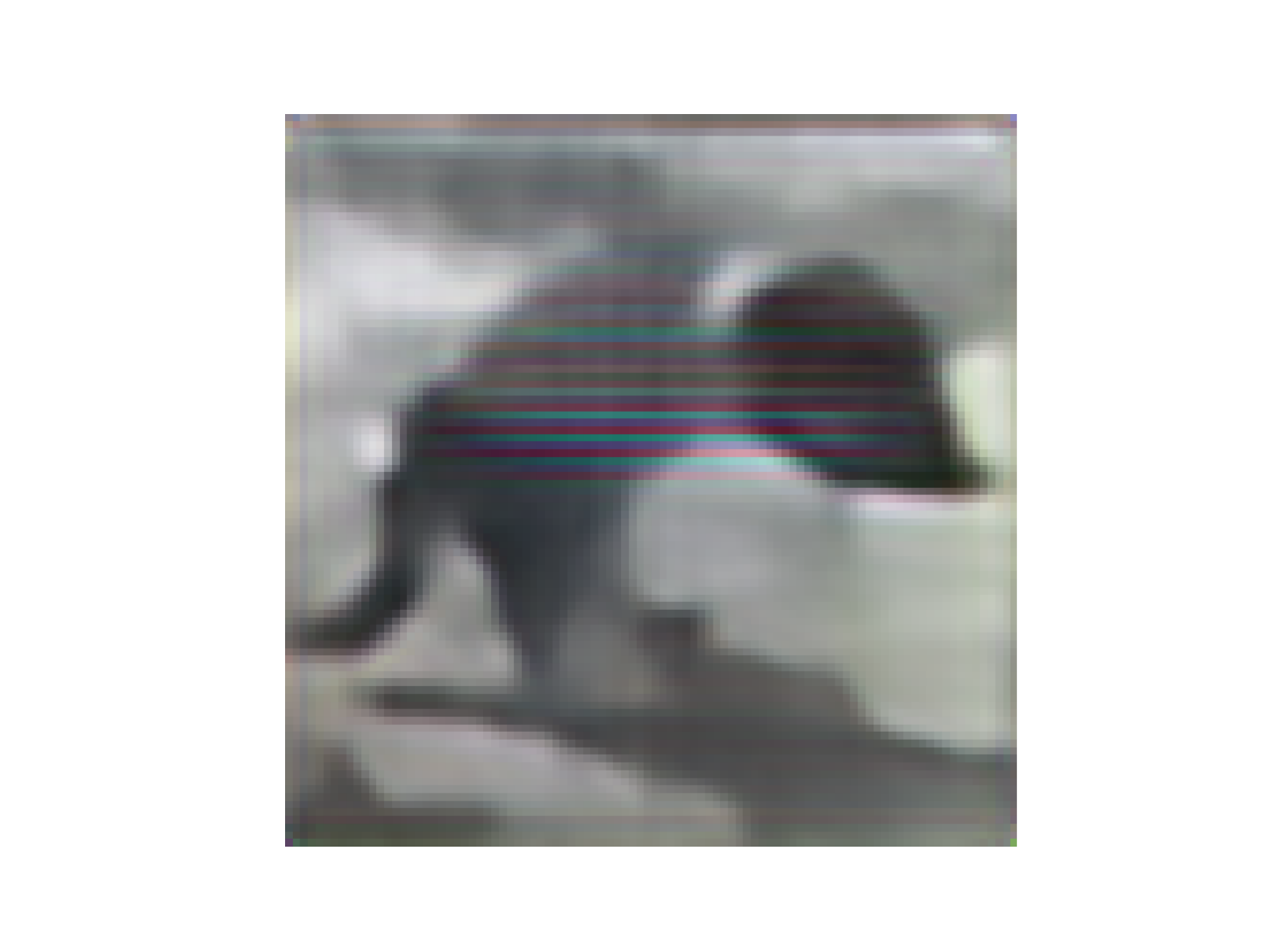} &
    \includegraphics[width=1.6cm, valign=c, trim={1cm 1cm 1cm 1cm},clip]{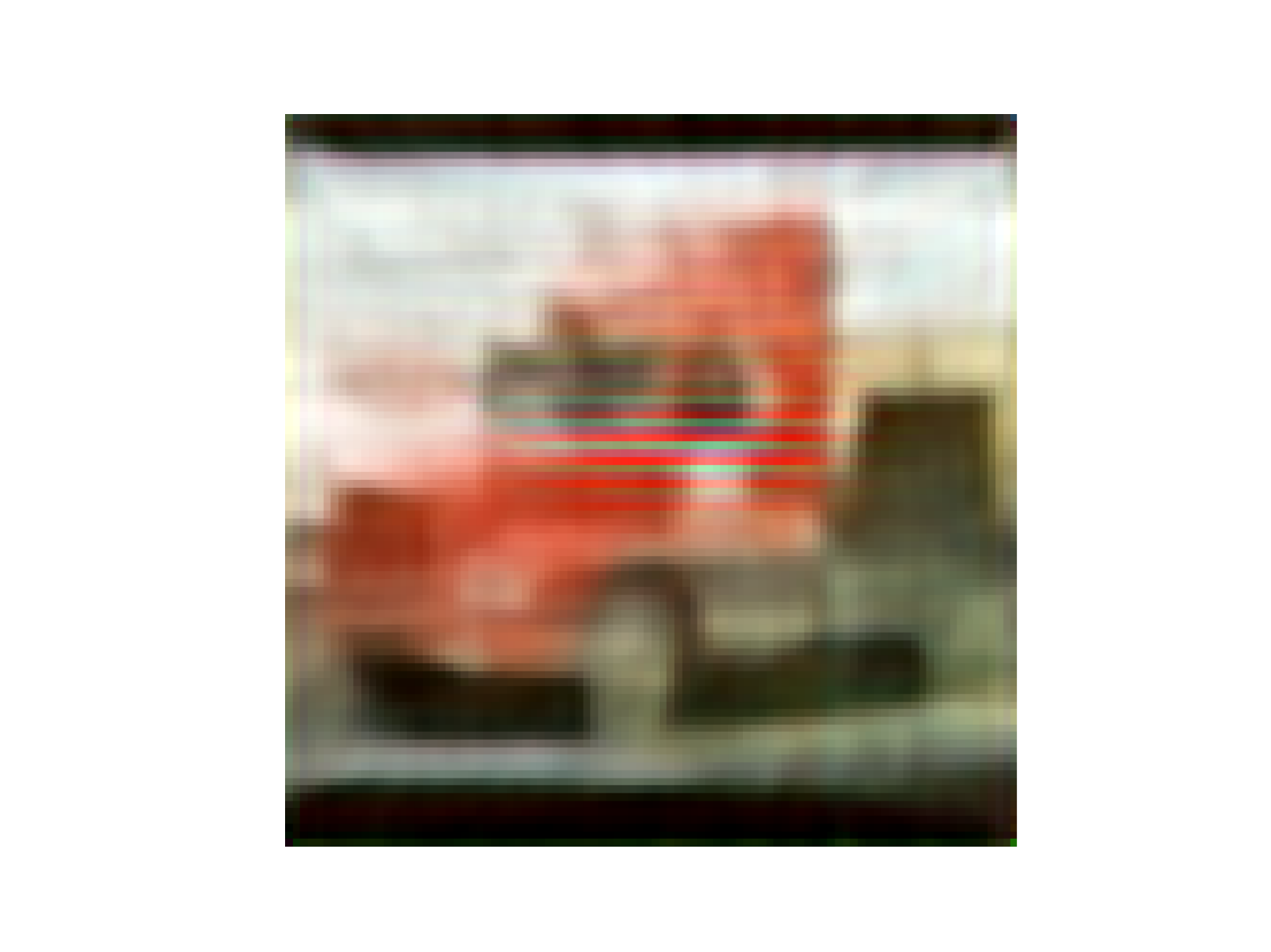} &
    \includegraphics[width=1.6cm, valign=c, trim={1cm 1cm 1cm 1cm},clip]{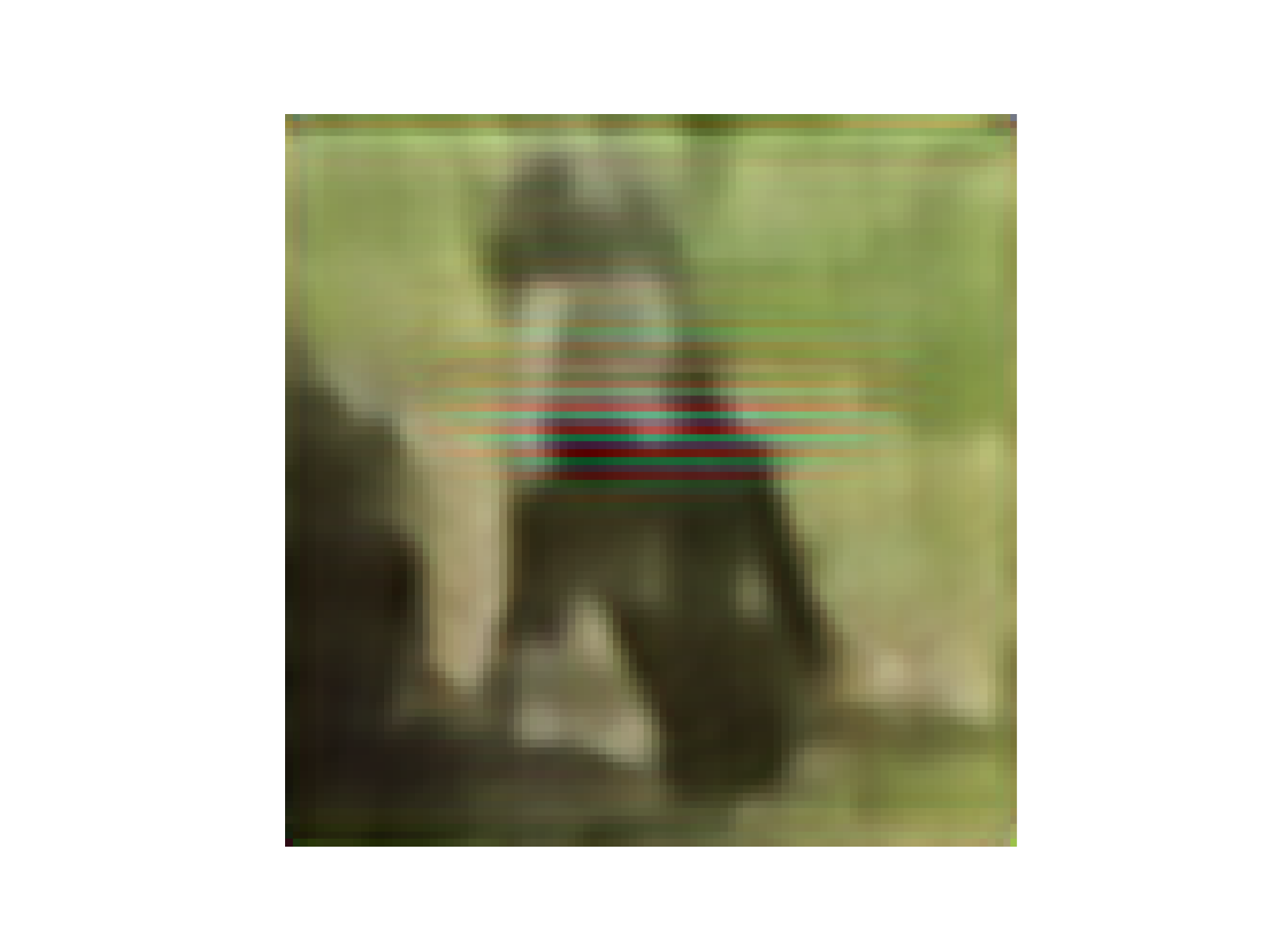} &
    \includegraphics[width=1.6cm, valign=c, trim={1cm 1cm 1cm 1cm},clip]{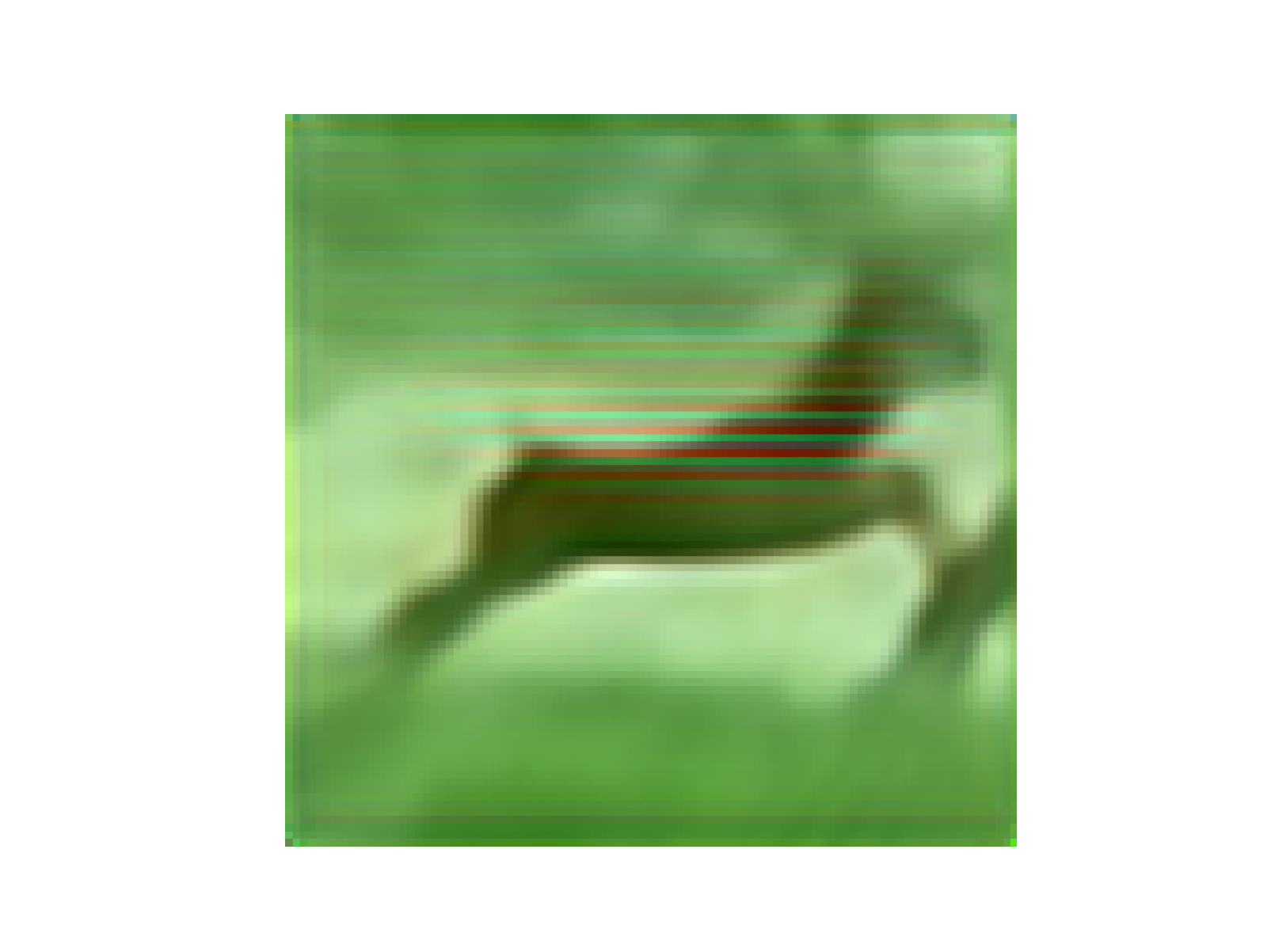}
    \\
        \hline

     \tiny{UNet} & \includegraphics[width=1.6cm, valign=c, trim={1cm 1cm 1cm 1cm},clip]{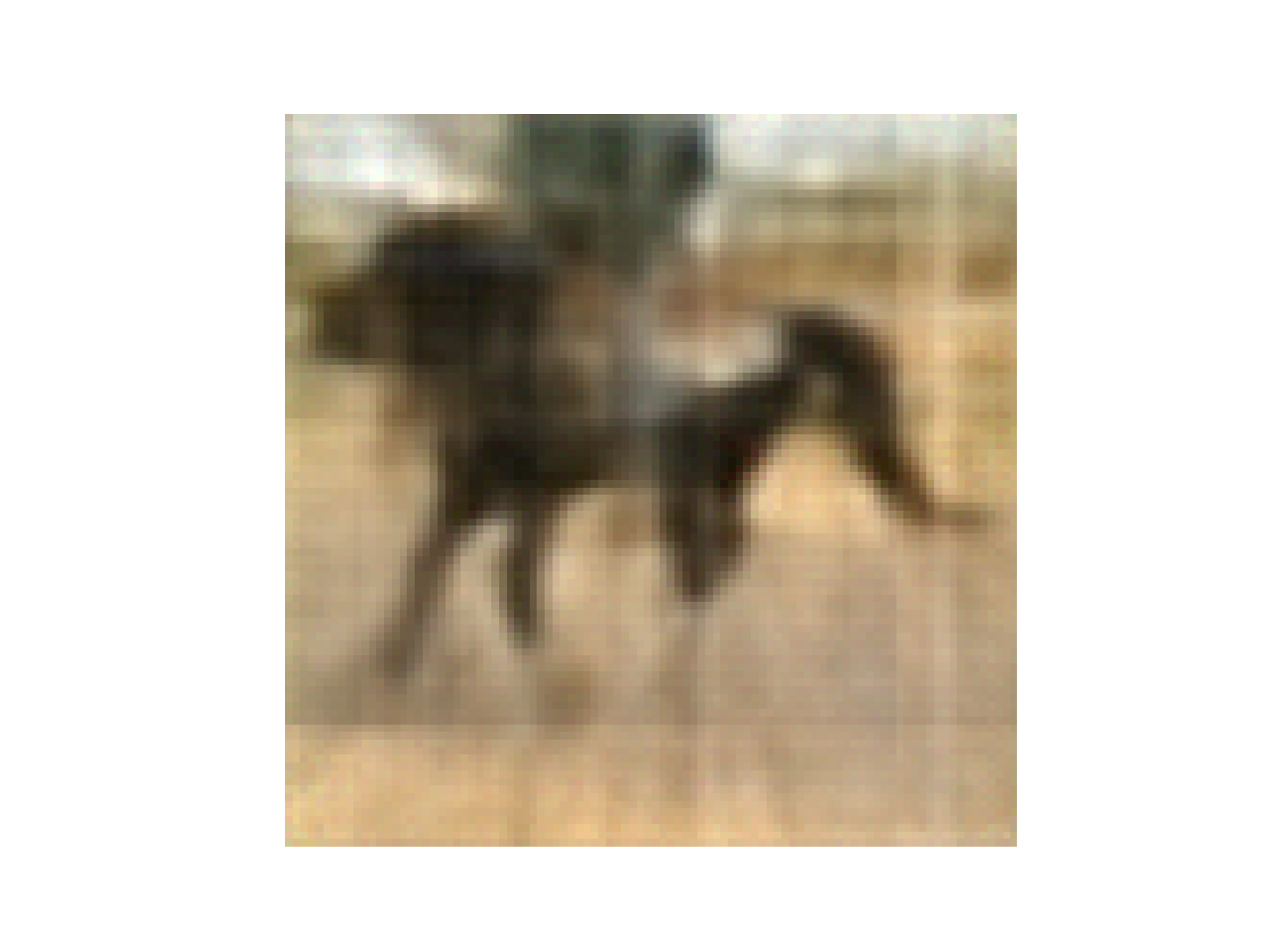}& \includegraphics[width=1.6cm, valign=c, trim={1cm 1cm 1cm 1cm},clip]{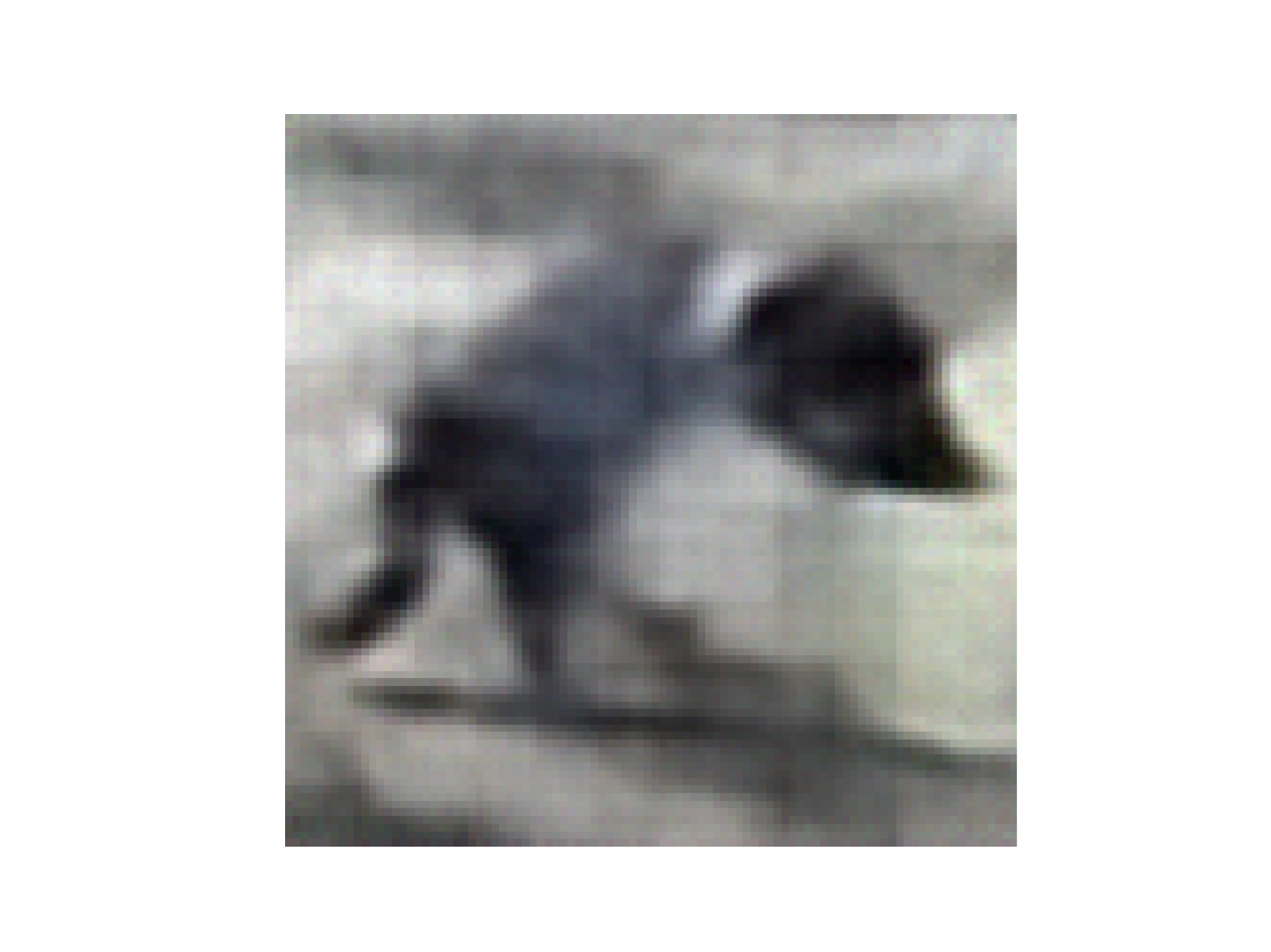}& \includegraphics[width=1.6cm, valign=c, trim={1cm 1cm 1cm 1cm},clip]{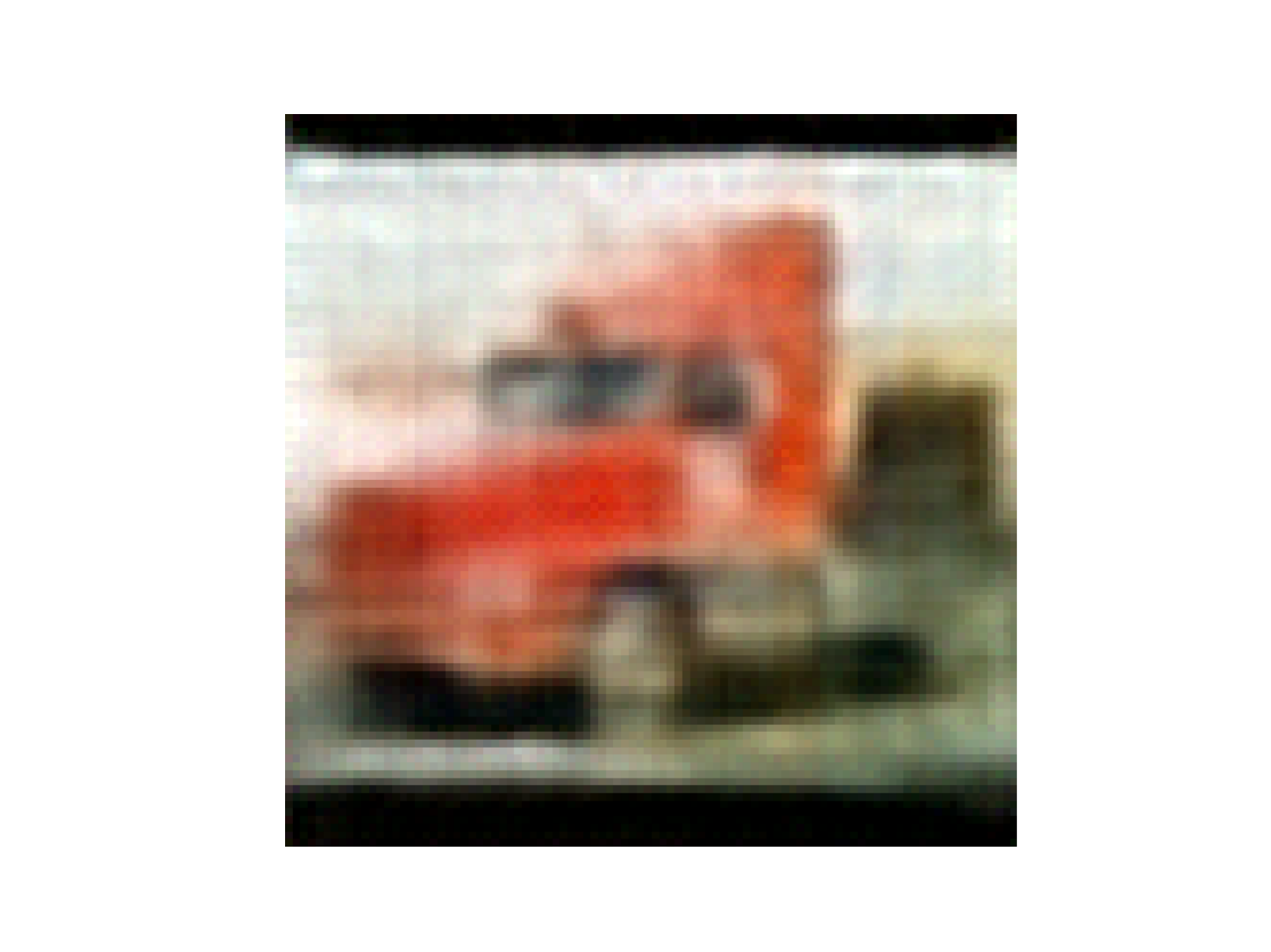}& \includegraphics[width=1.6cm, valign=c, trim={1cm 1cm 1cm 1cm},clip]{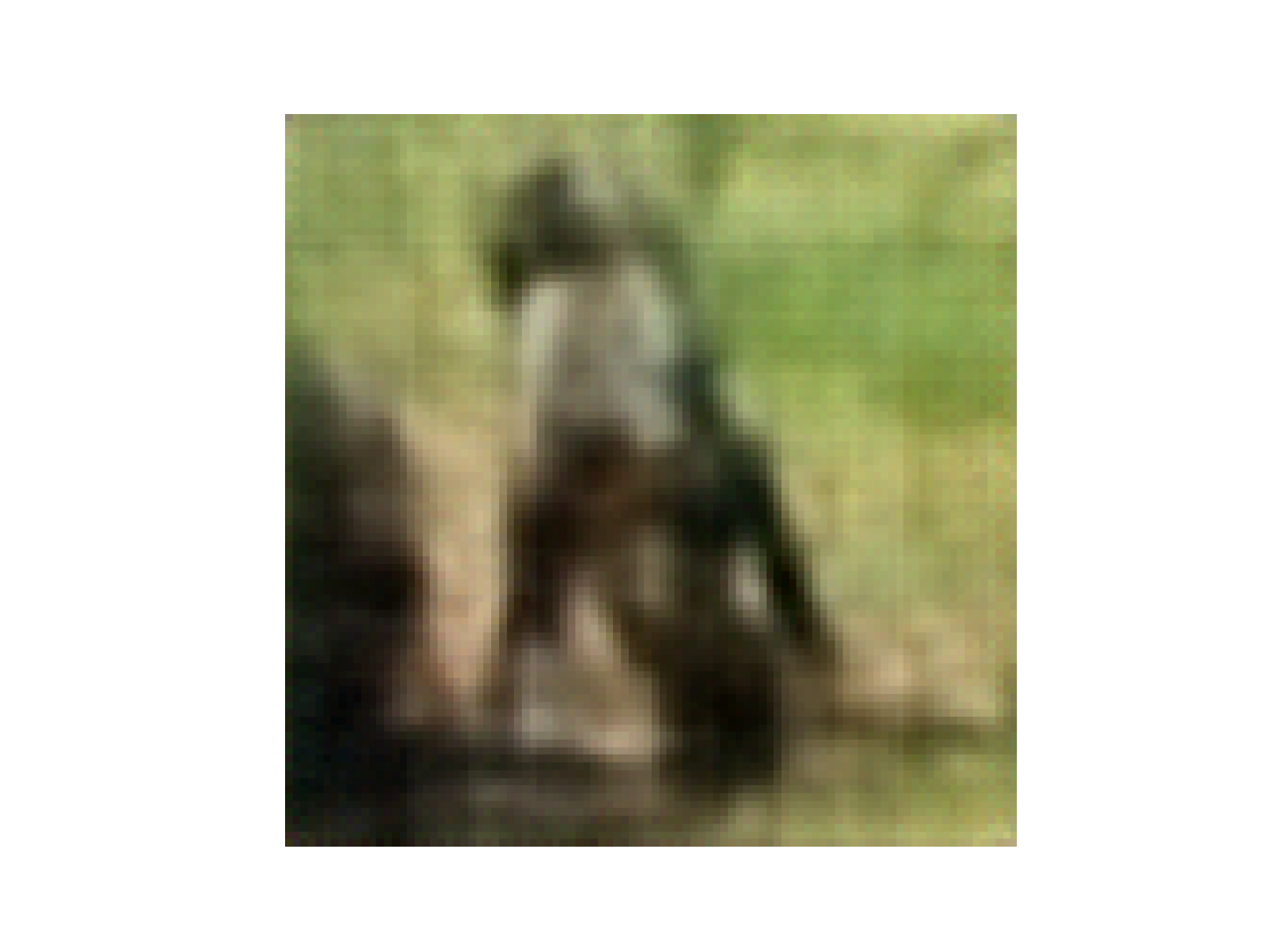}  & \includegraphics[width=1.6cm, valign=c, trim={1cm 1cm 1cm 1cm},clip]{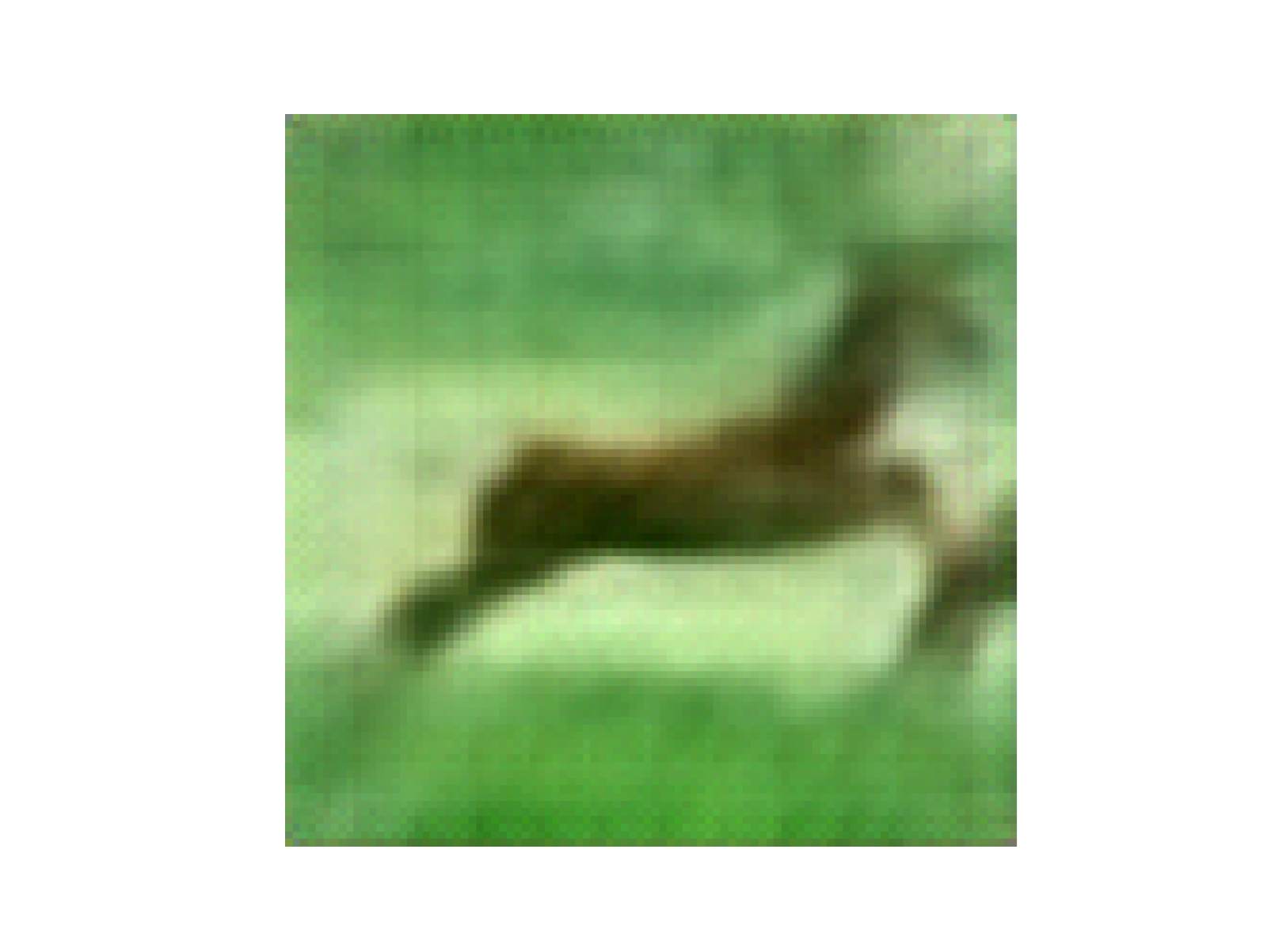}       \\
        \hline
     \tiny{ResNet}
     & \includegraphics[width=1.6cm, valign=c, trim={1cm 1cm 1cm 1cm},clip]{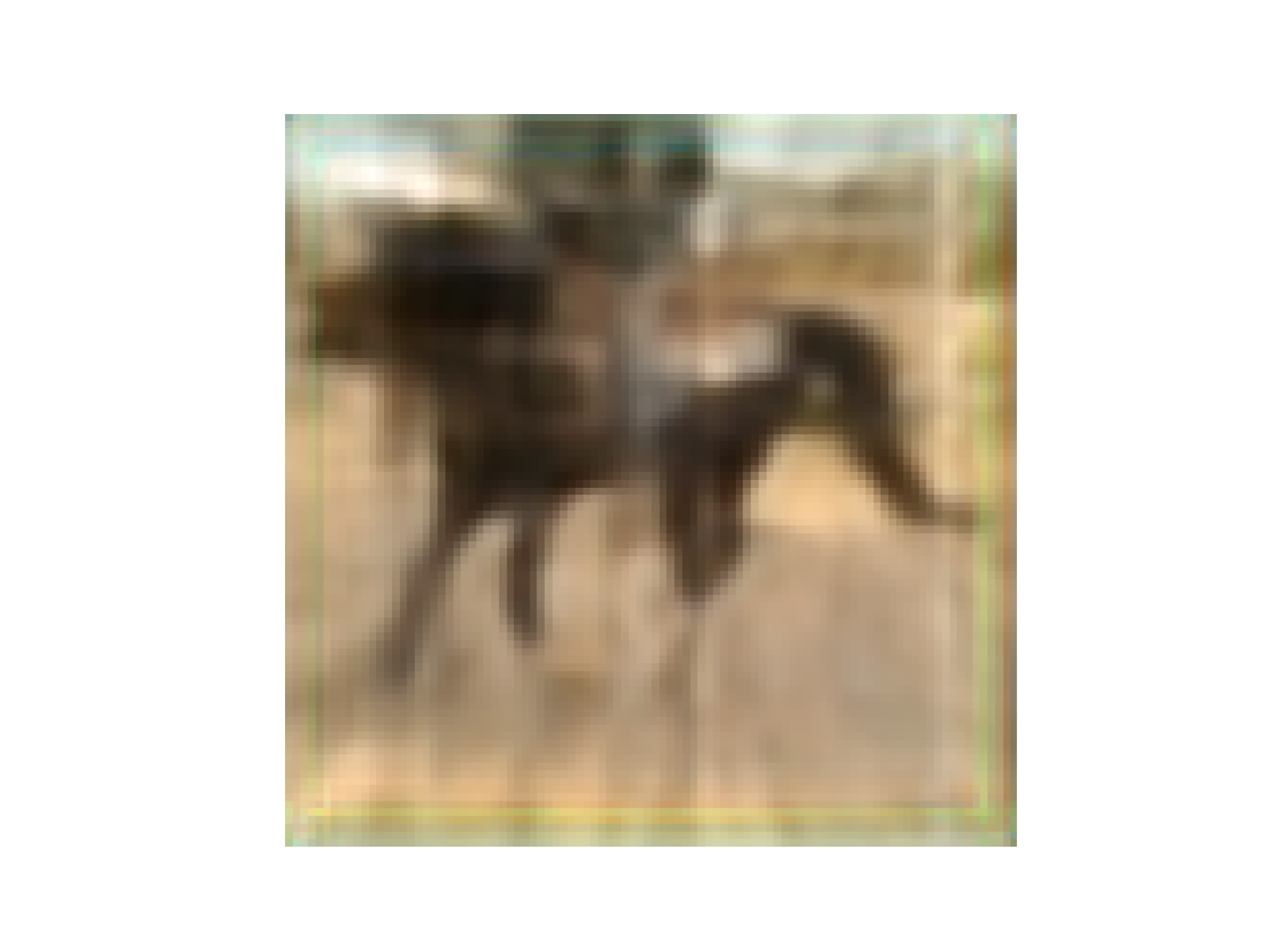}
     & \includegraphics[width=1.6cm, valign=c, trim={1cm 1cm 1cm 1cm},clip]{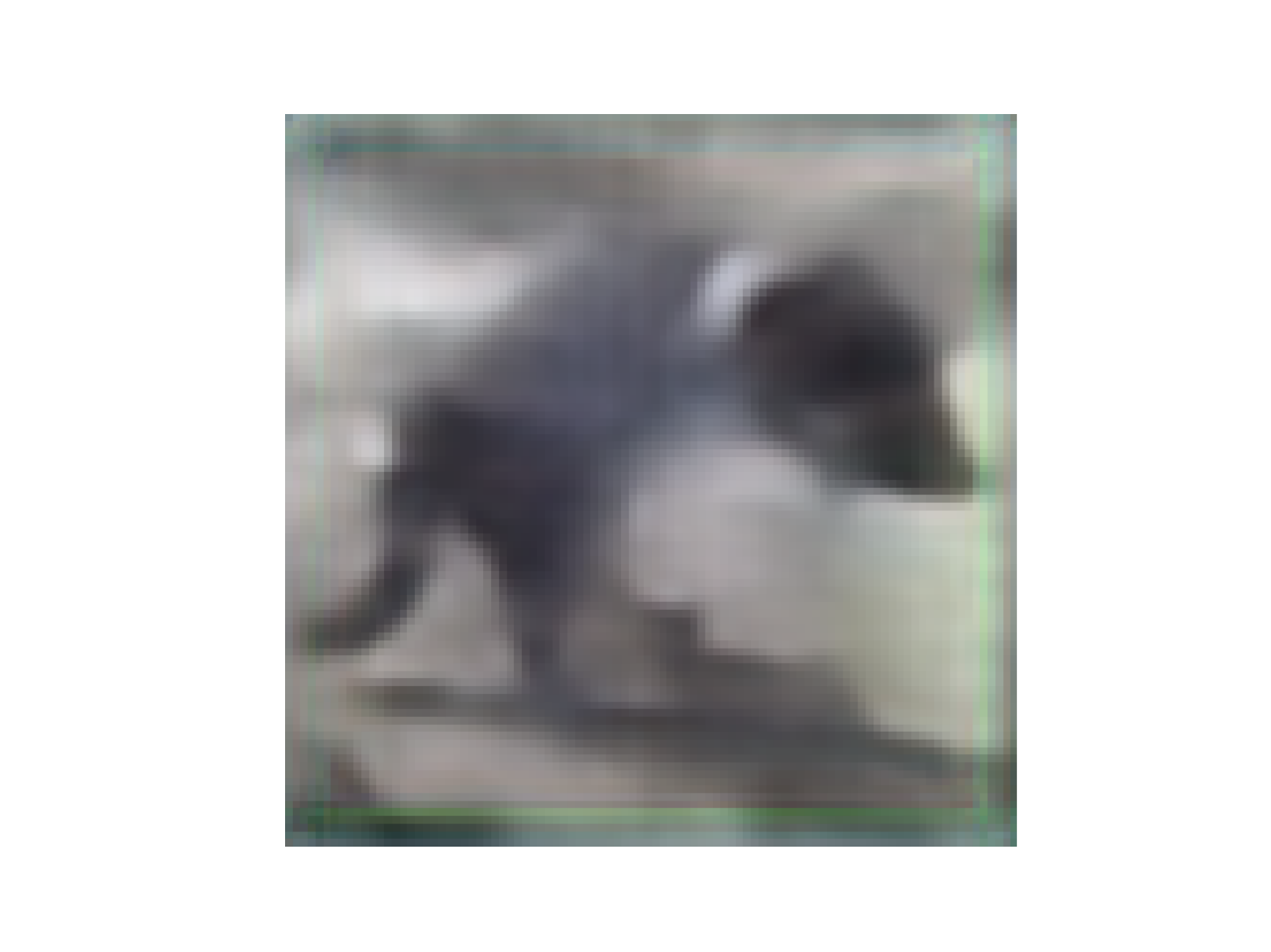}
     & \includegraphics[width=1.6cm, valign=c, trim={1cm 1cm 1cm 1cm},clip]{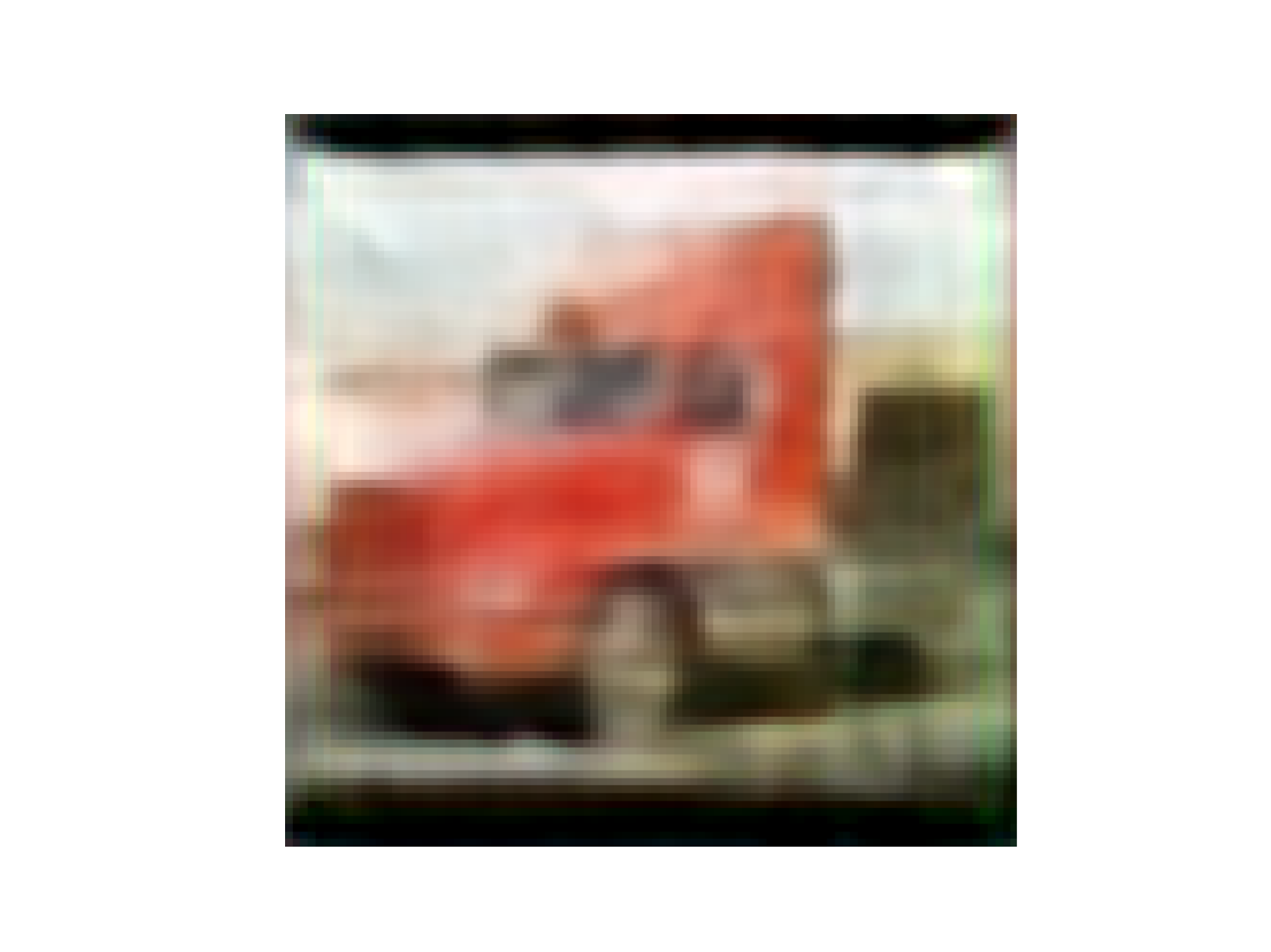}
     & \includegraphics[width=1.6cm, valign=c, trim={1cm 1cm 1cm 1cm},clip]{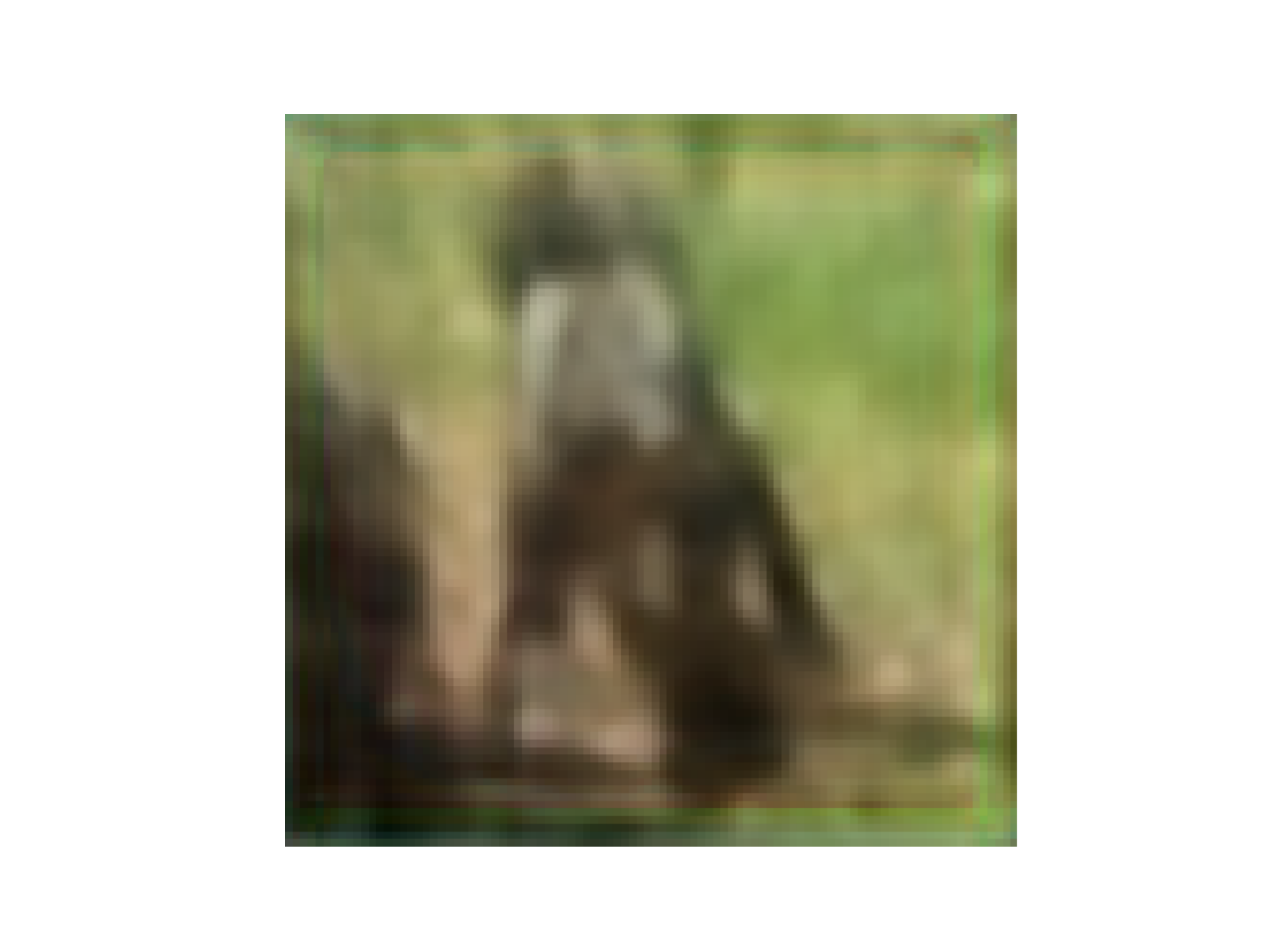}      & \includegraphics[width=1.6cm, valign=c, trim={1cm 1cm 1cm 1cm},clip]{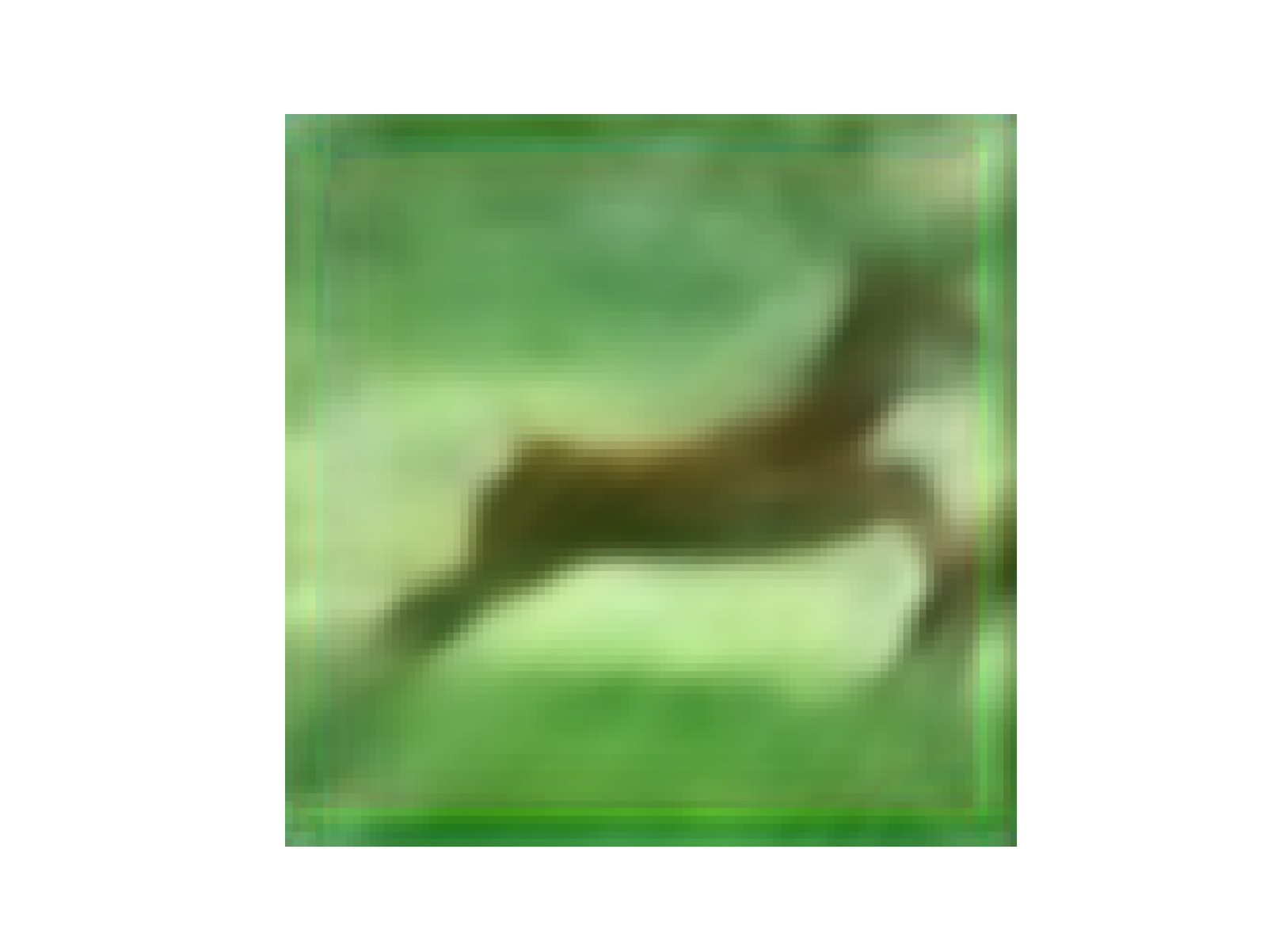}
     \\
     \hline
     \tiny{Ground-truth} & \includegraphics[width=1.6cm, valign=c, trim={1cm 1cm 1cm 1cm},clip]{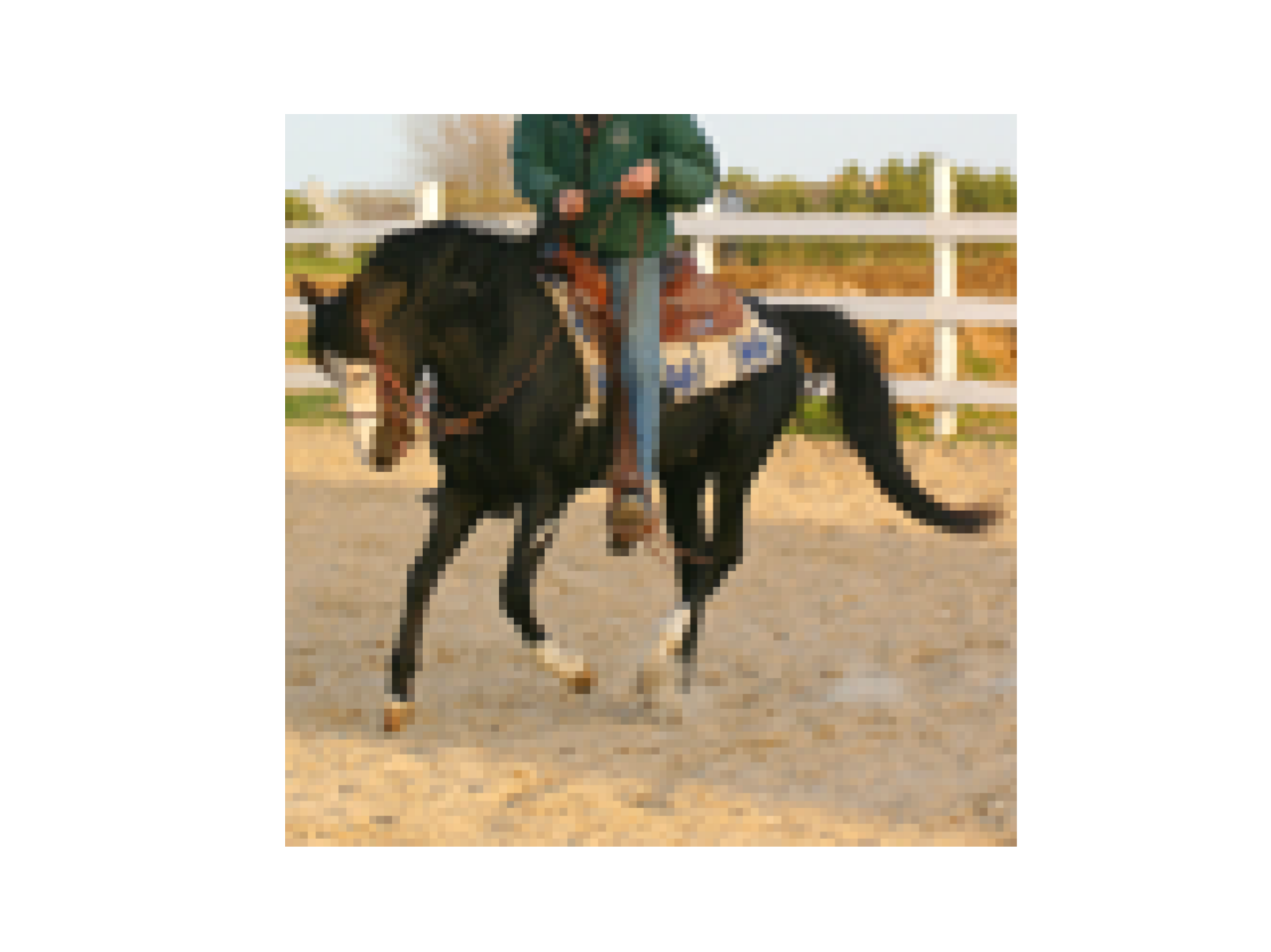} & \includegraphics[width=1.6cm, valign=c, trim={1cm 1cm 1cm 1cm},clip]{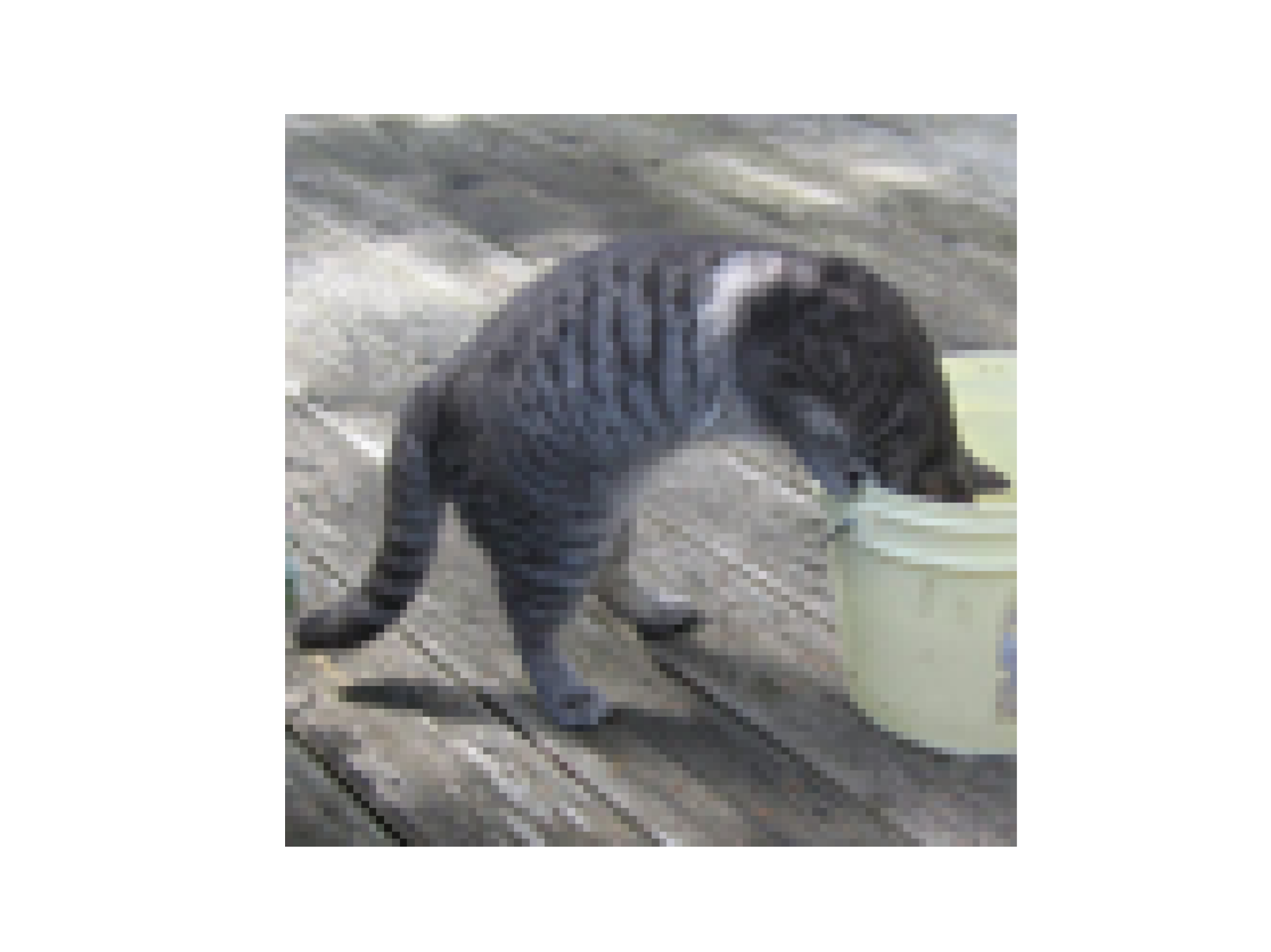}& \includegraphics[width=1.6cm, valign=c, trim={1cm 1cm 1cm 1cm},clip]{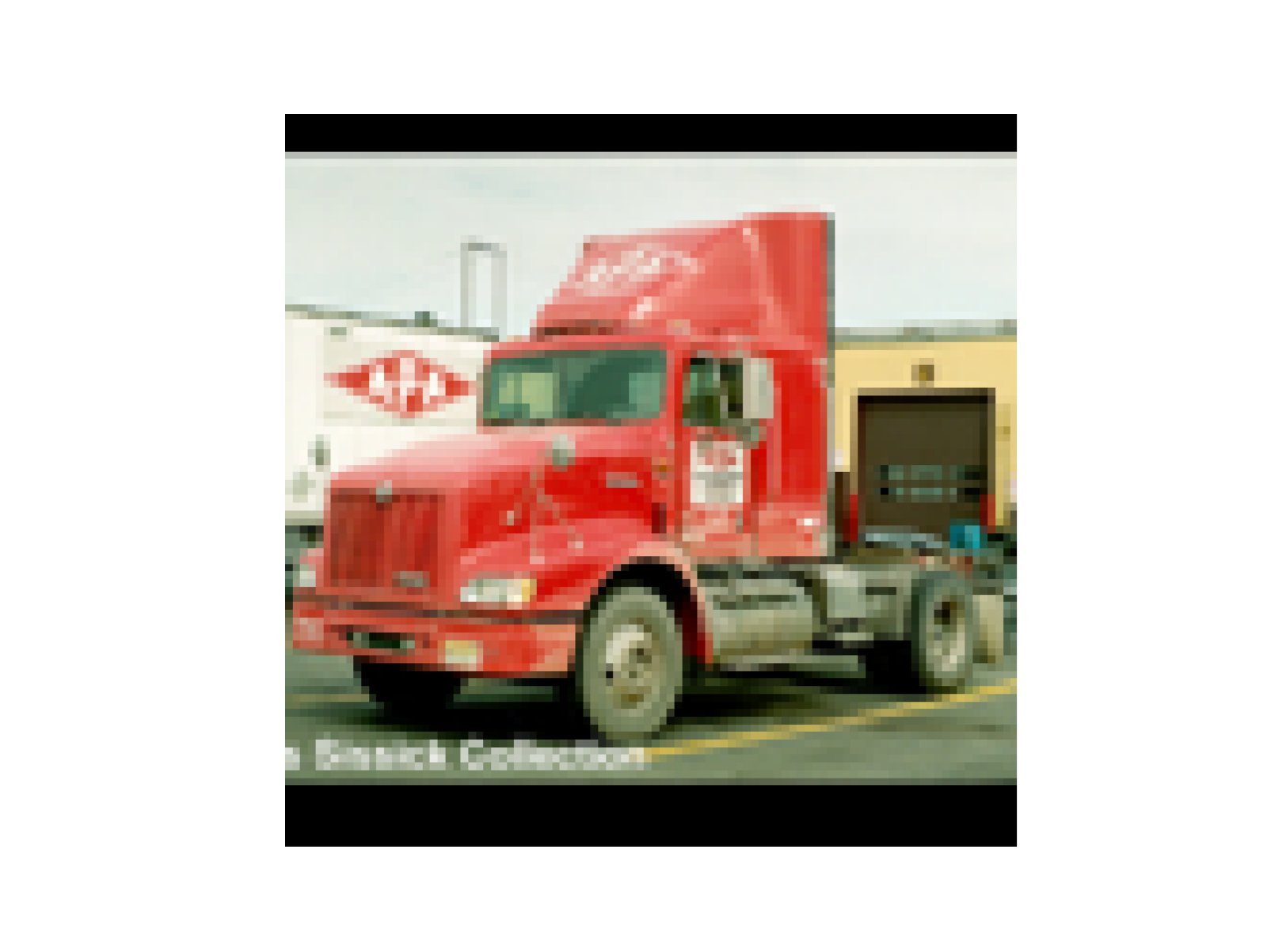} & \includegraphics[width=1.6cm, valign=c, trim={1cm 1cm 1cm 1cm},clip]{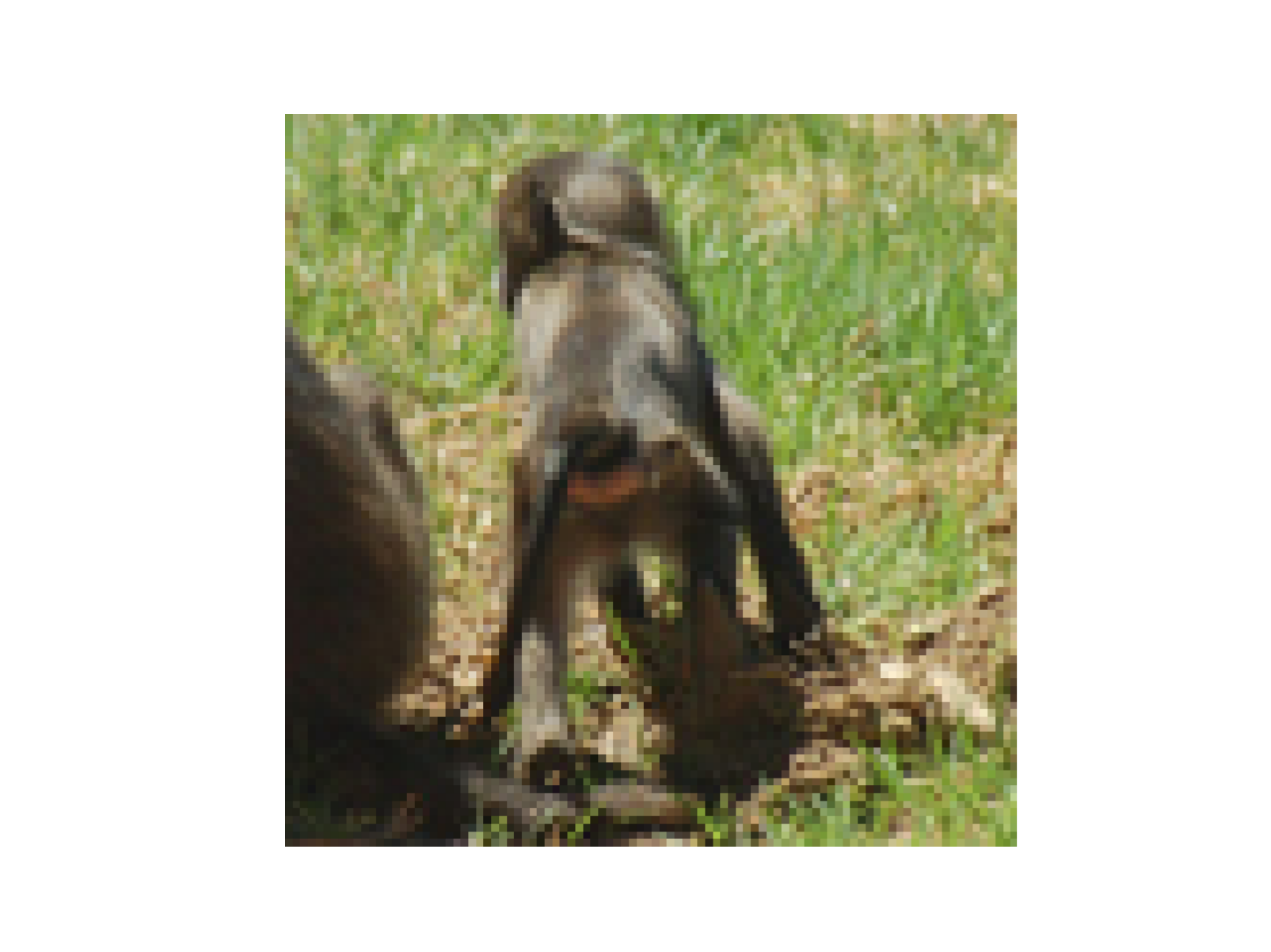} & \includegraphics[width=1.6cm, valign=c, trim={1cm 1cm 1cm 1cm},clip]{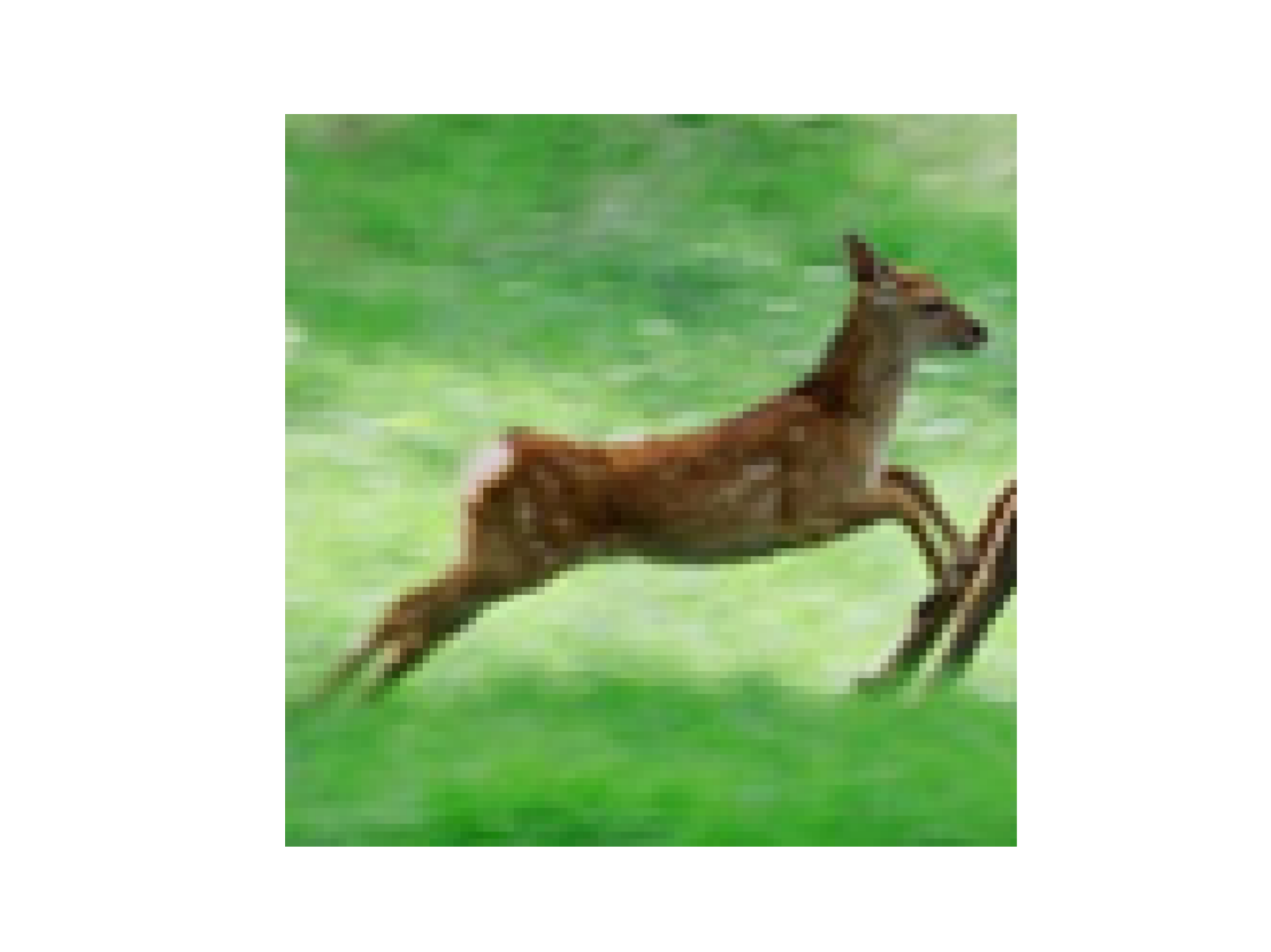}   \\
     \hline
    \end{tabular}
    \captionof{figure}{Examples of recovered images from the Tomography task { trained and tested on the STL-10 dataset}. While all models are able to reconstruct the images, our DRIP-based LA-Net and Hyper-ResNet architectures yield images with the lowest residual and error terms.}
    \label{fig:tomography_results}
\end{table}

\begin{table}
    \centering
    \renewcommand{\arraystretch}{2}
    \begin{tabular}{|cccccc|}
    \hline 
     \tiny{Data} & \includegraphics[width=1.6cm, valign=c, trim={1cm 1cm 1cm 1cm},clip]{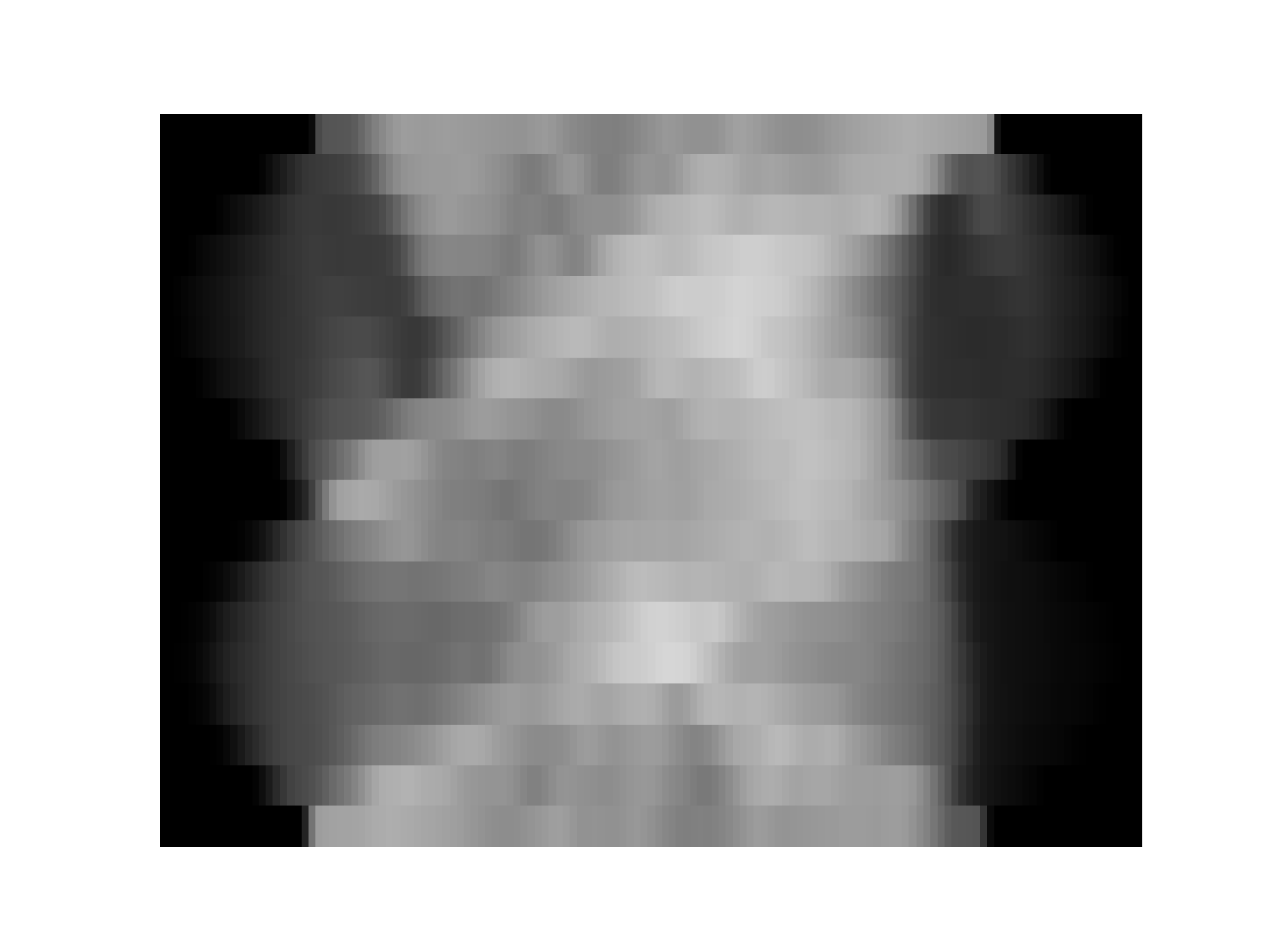} & \includegraphics[width=1.6cm, valign=c, trim={1cm 1cm 1cm 1cm},clip]{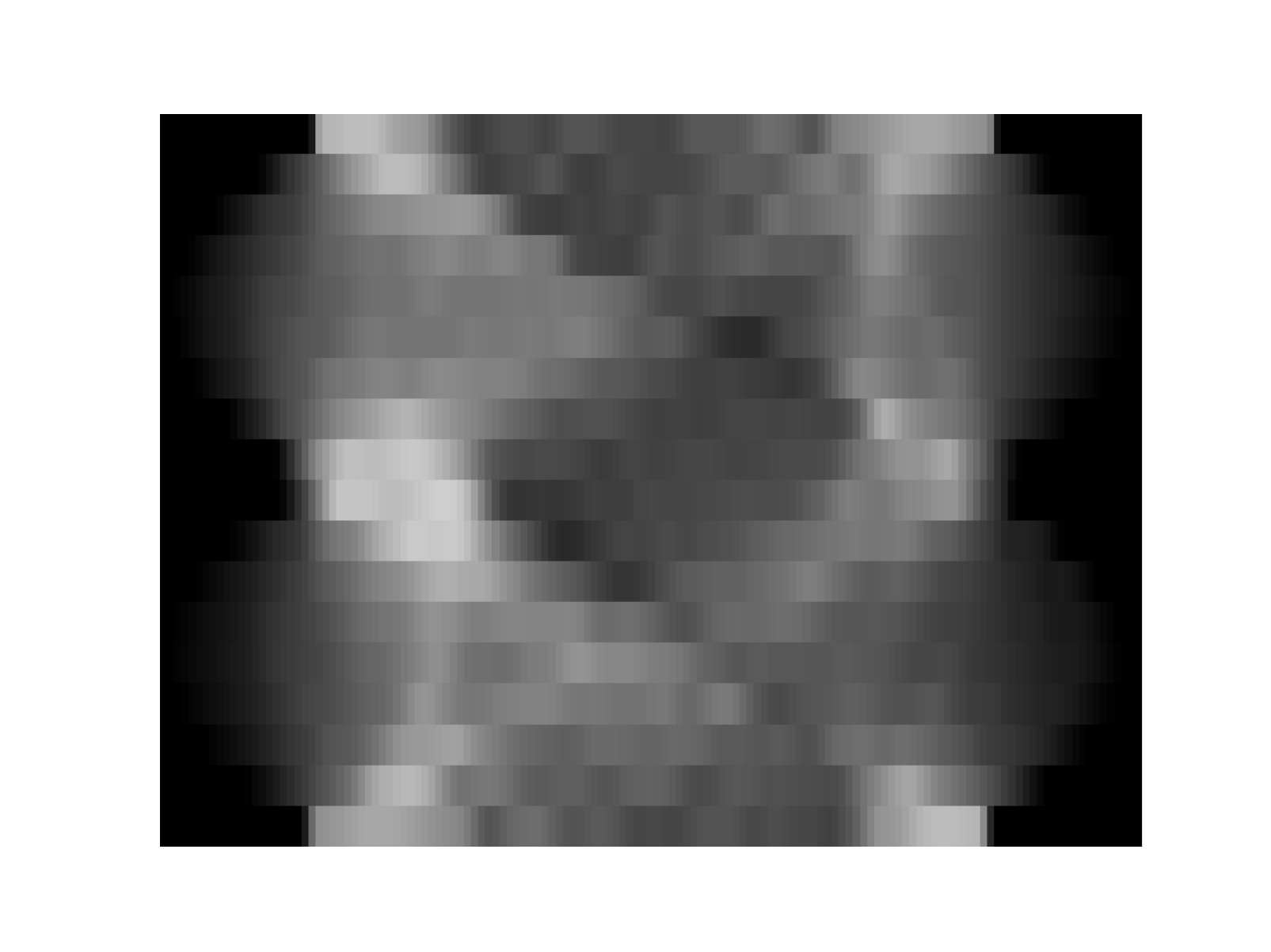} & \includegraphics[width=1.6cm, valign=c, trim={1cm 1cm 1cm 1cm},clip]{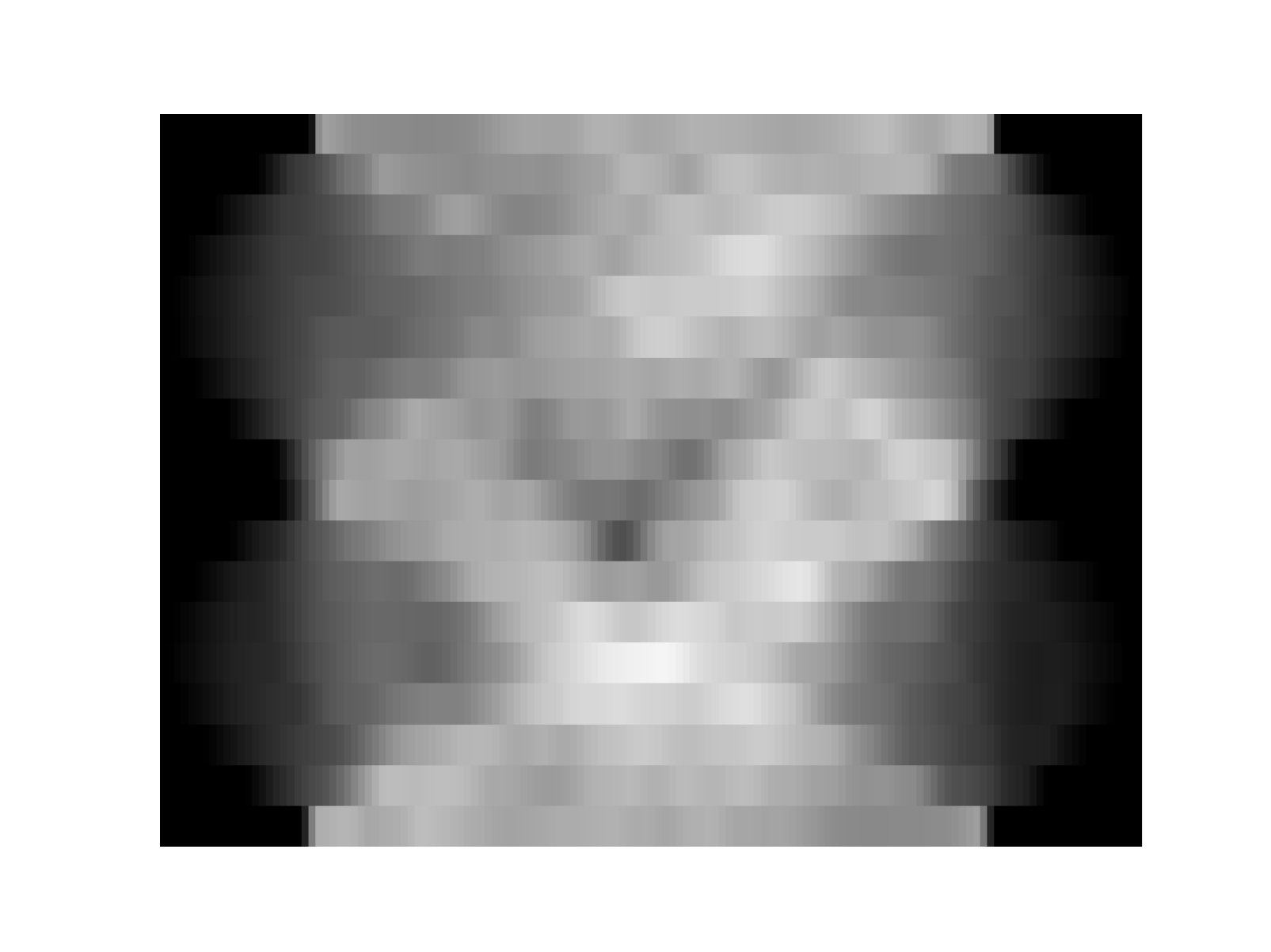} & \includegraphics[width=1.6cm, valign=c, trim={1cm 1cm 1cm 1cm},clip]{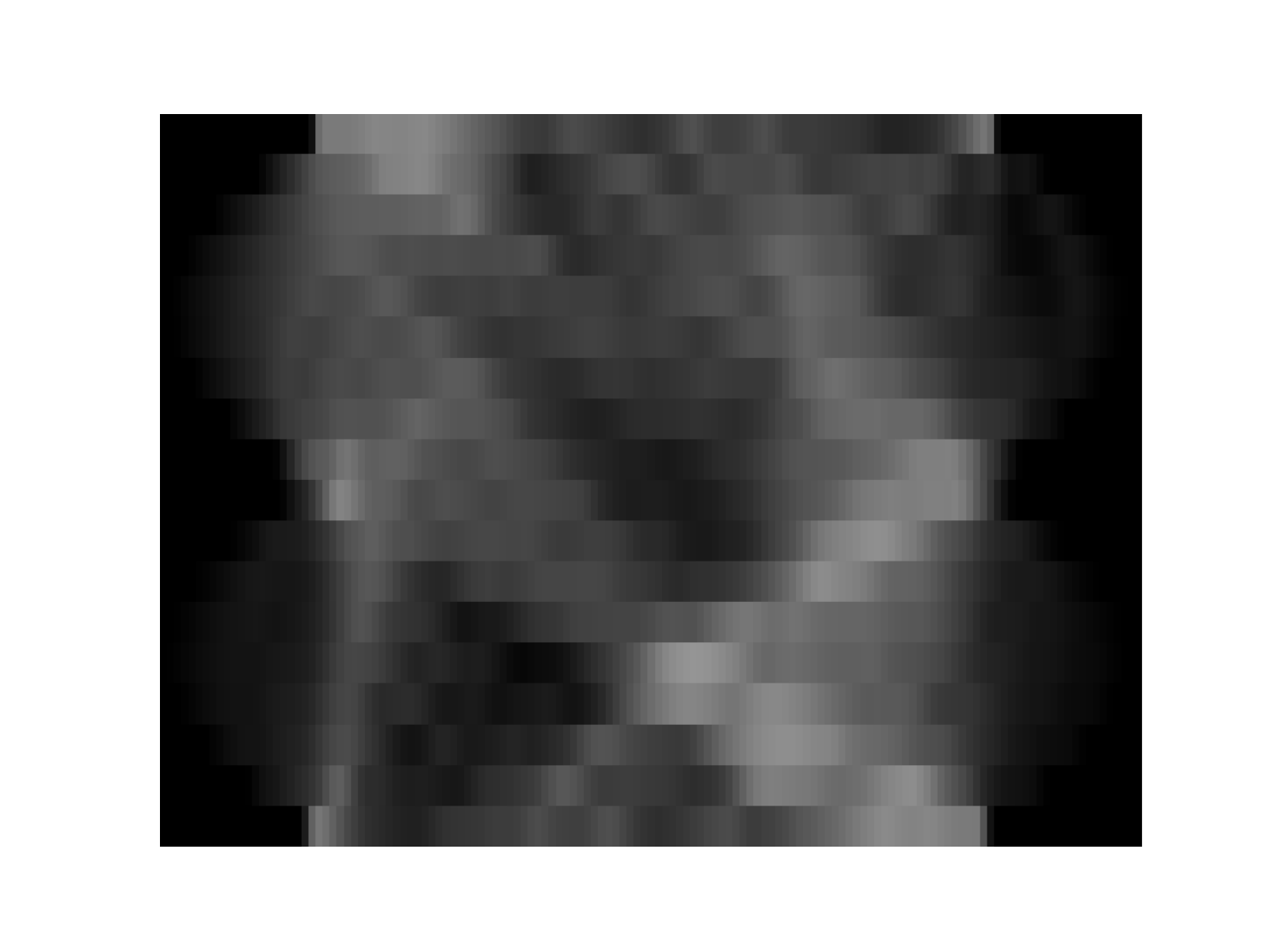} & \includegraphics[width=1.6cm, valign=c, trim={1cm 1cm 1cm 1cm},clip]{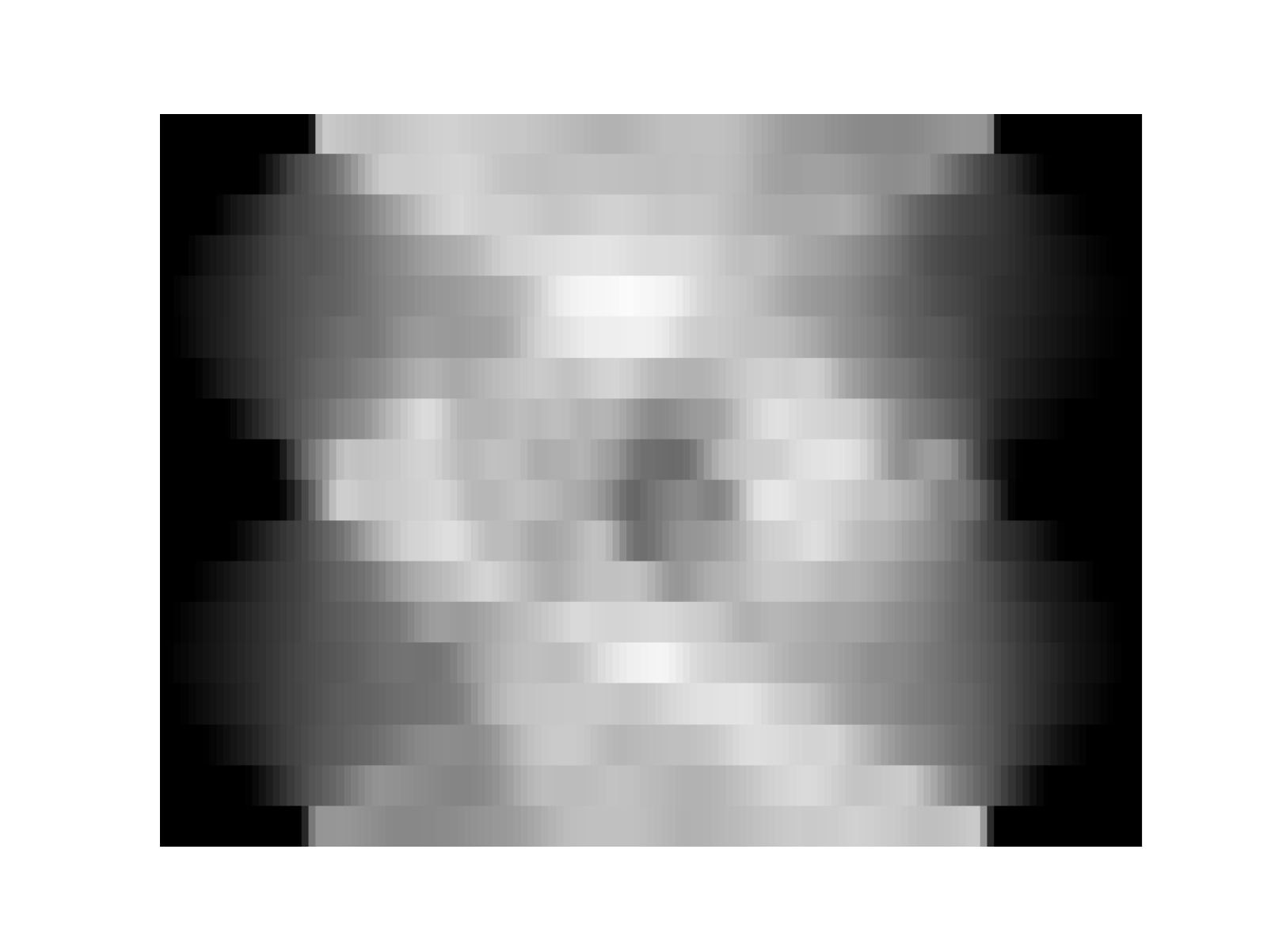} \\
     \hline
    \tiny{LA-Net} & 
    \includegraphics[width=1.6cm, valign=c, trim={1cm 1cm 1cm 1cm},clip]{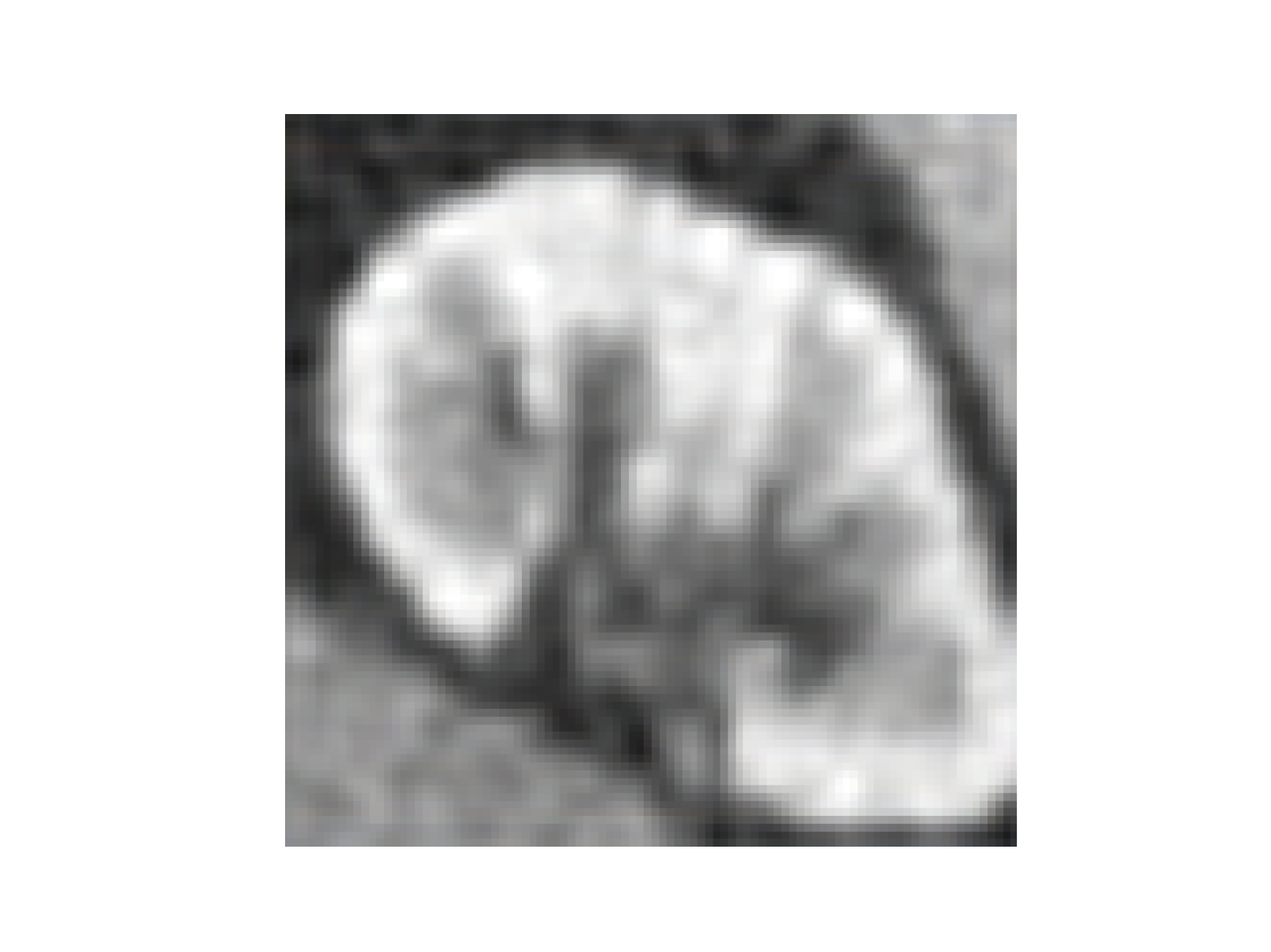}
     &
    \includegraphics[width=1.6cm, valign=c, trim={1cm 1cm 1cm 1cm},clip]{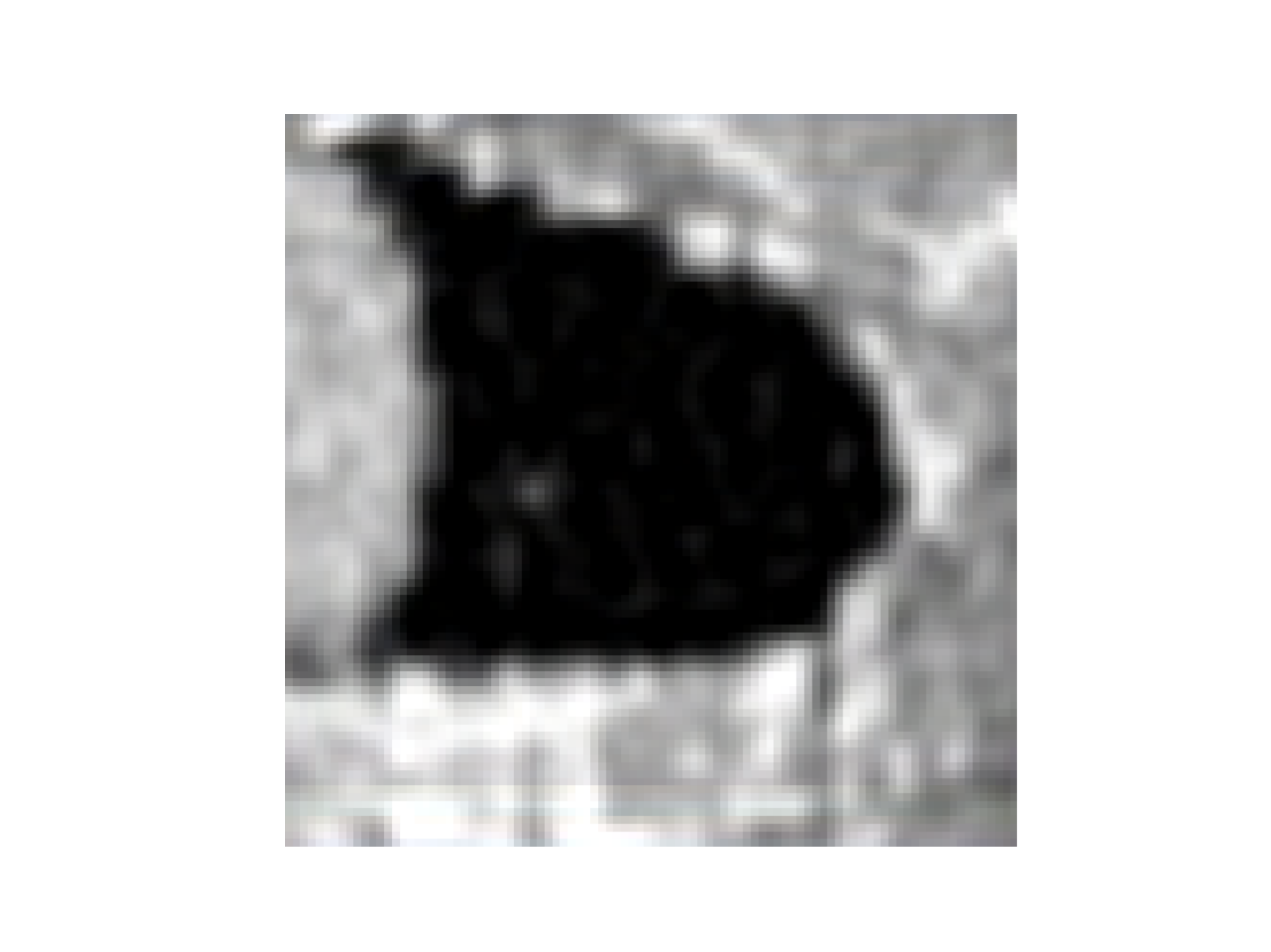} &
    \includegraphics[width=1.6cm, valign=c, trim={1cm 1cm 1cm 1cm},clip]{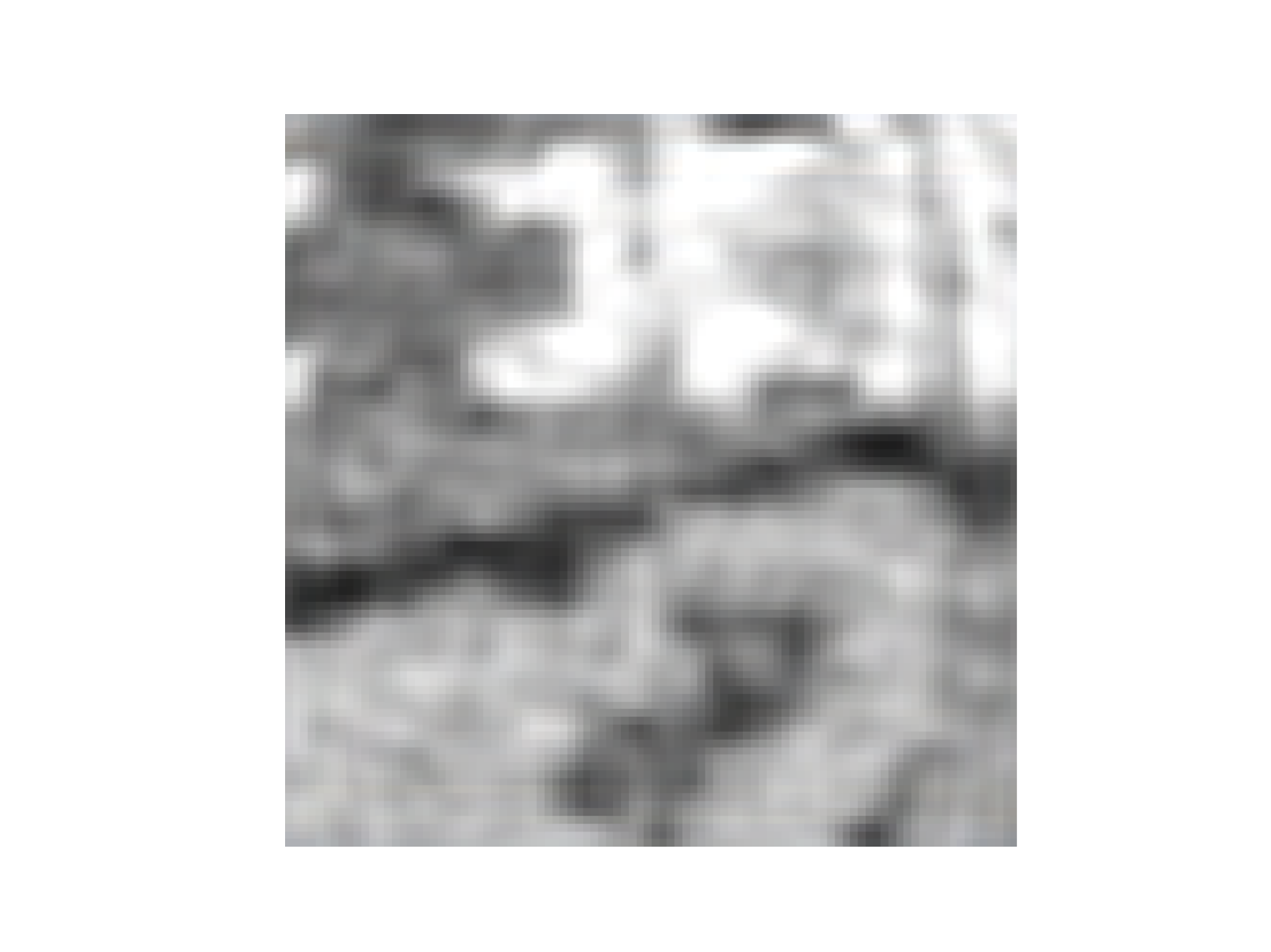} &
    \includegraphics[width=1.6cm, valign=c, trim={1cm 1cm 1cm 1cm},clip]{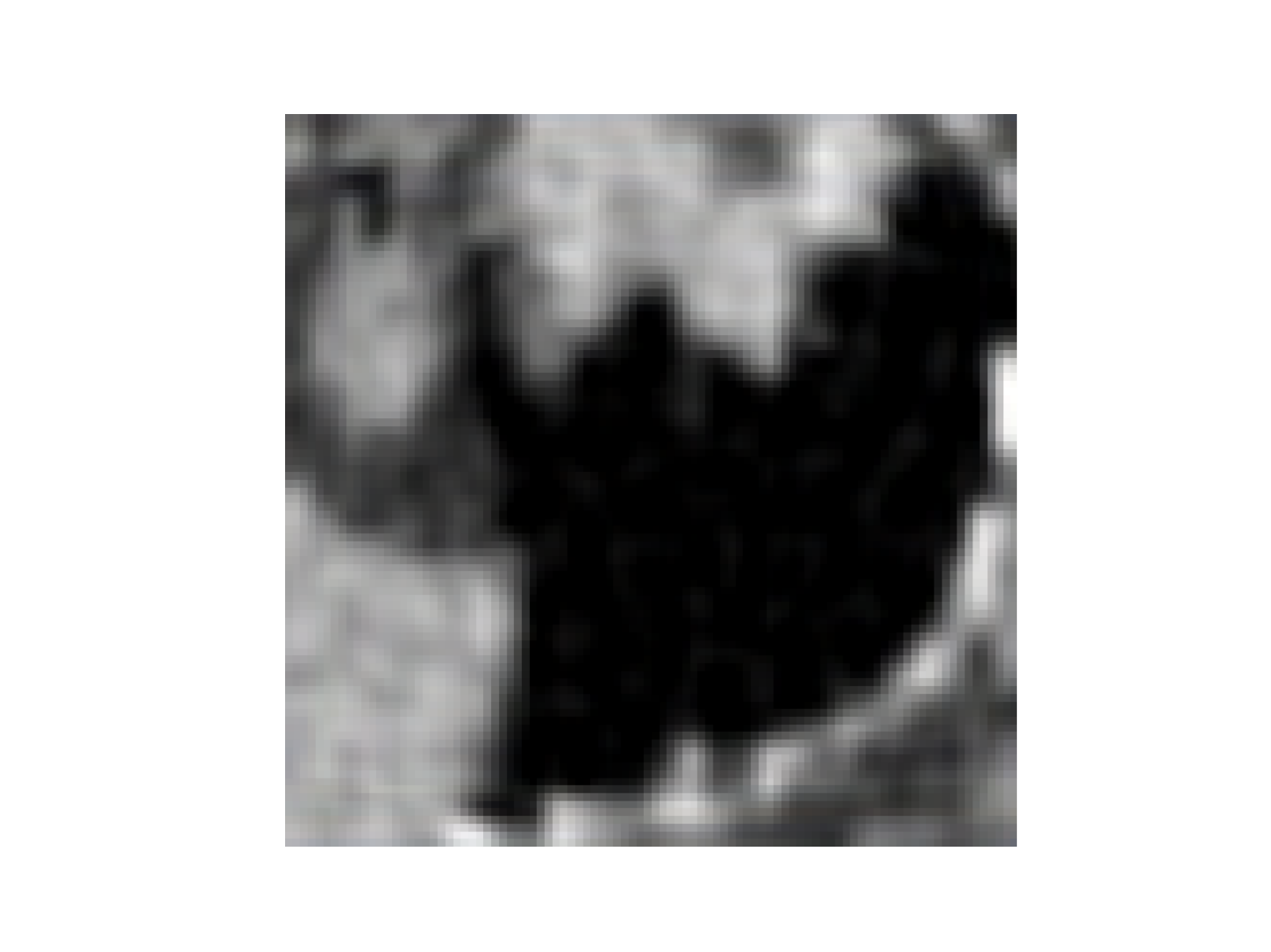} &
    \includegraphics[width=1.6cm, valign=c, trim={1cm 1cm 1cm 1cm},clip]{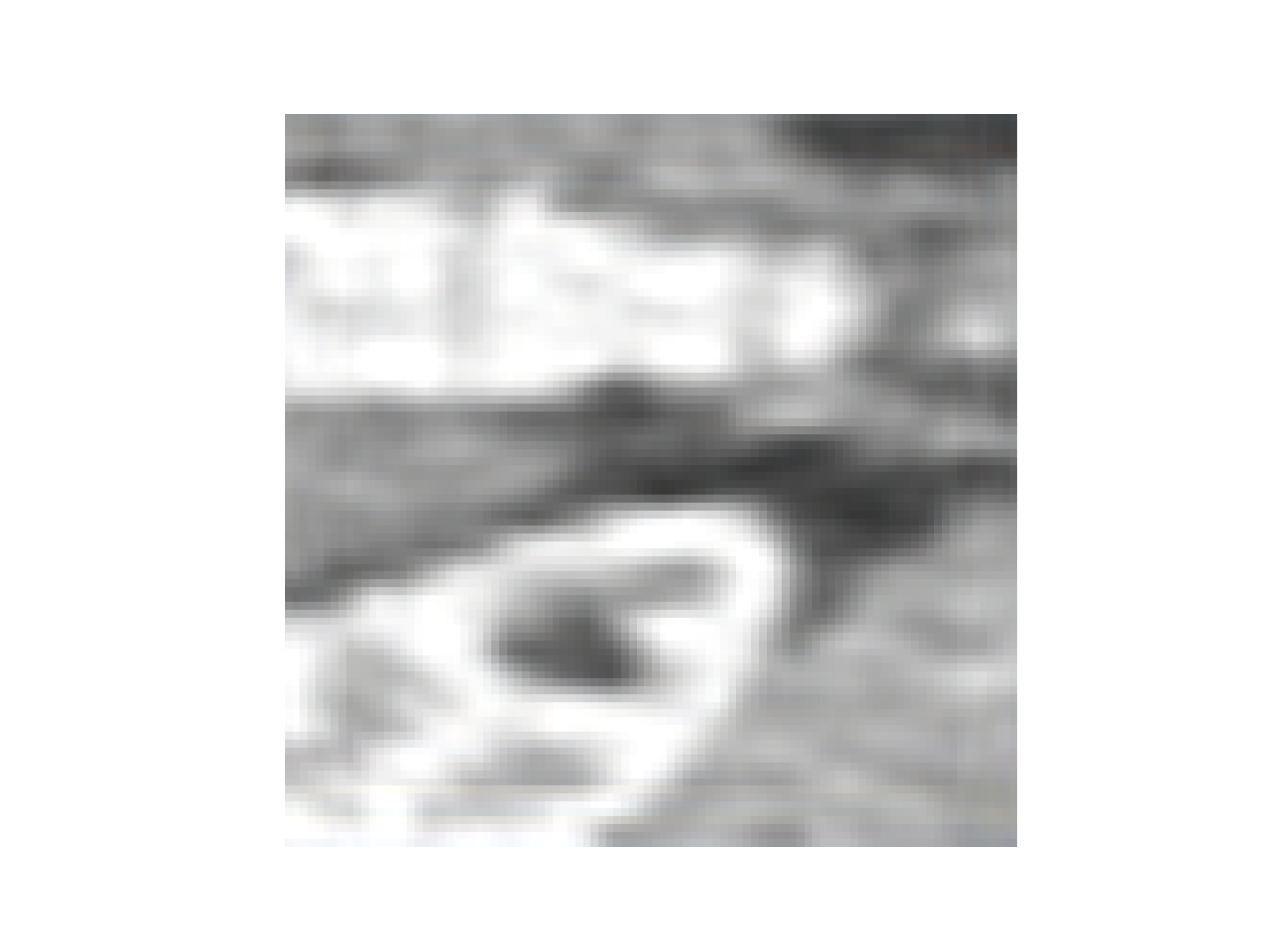}
    \\
        \hline
     
      \tiny{Hyper-ResNet} &  \includegraphics[width=1.6cm, valign=c, trim={1cm 1cm 1cm 1cm},clip]{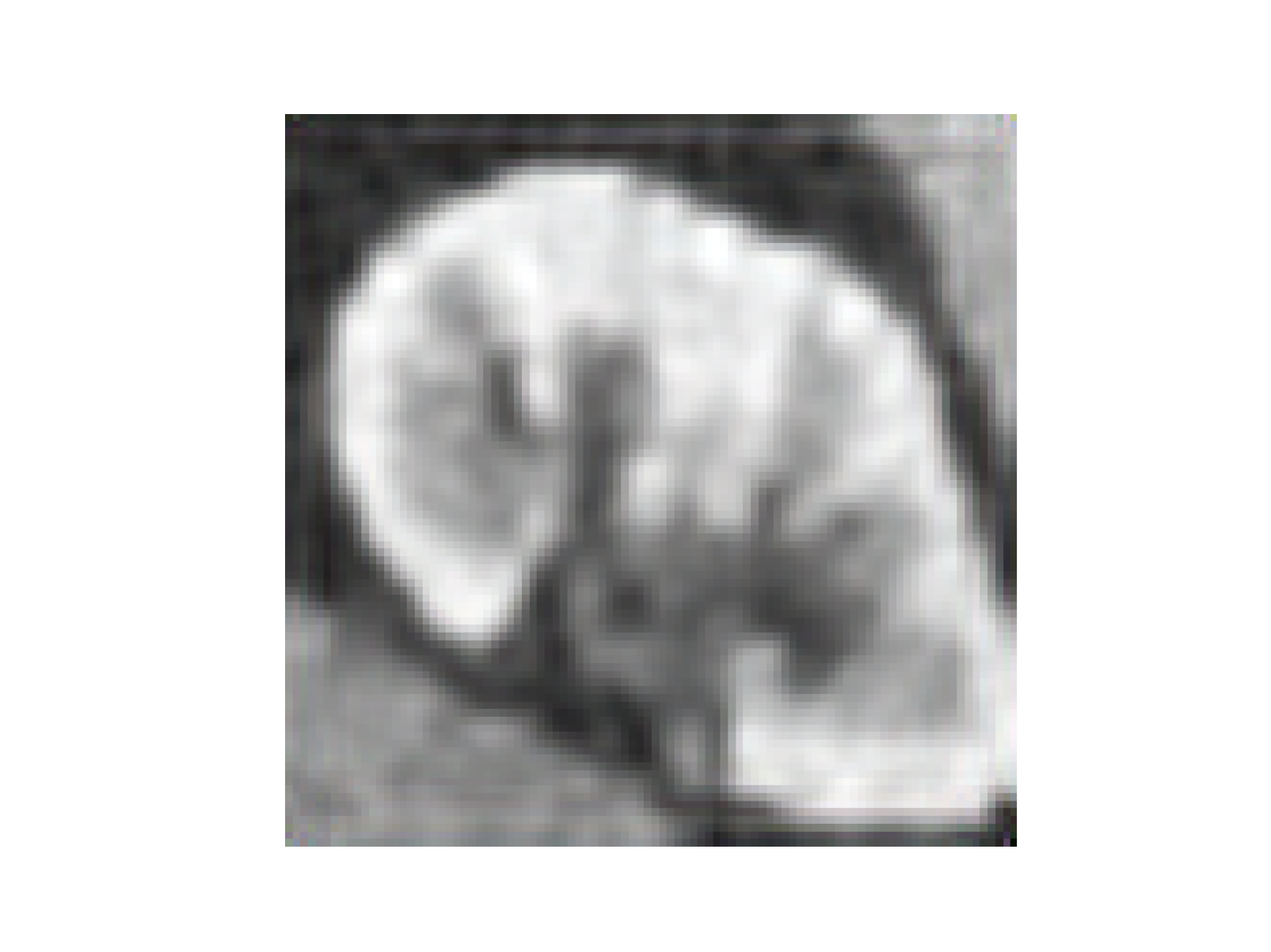}  &\includegraphics[width=1.6cm, valign=c, trim={1cm 1cm 1cm 1cm},clip]{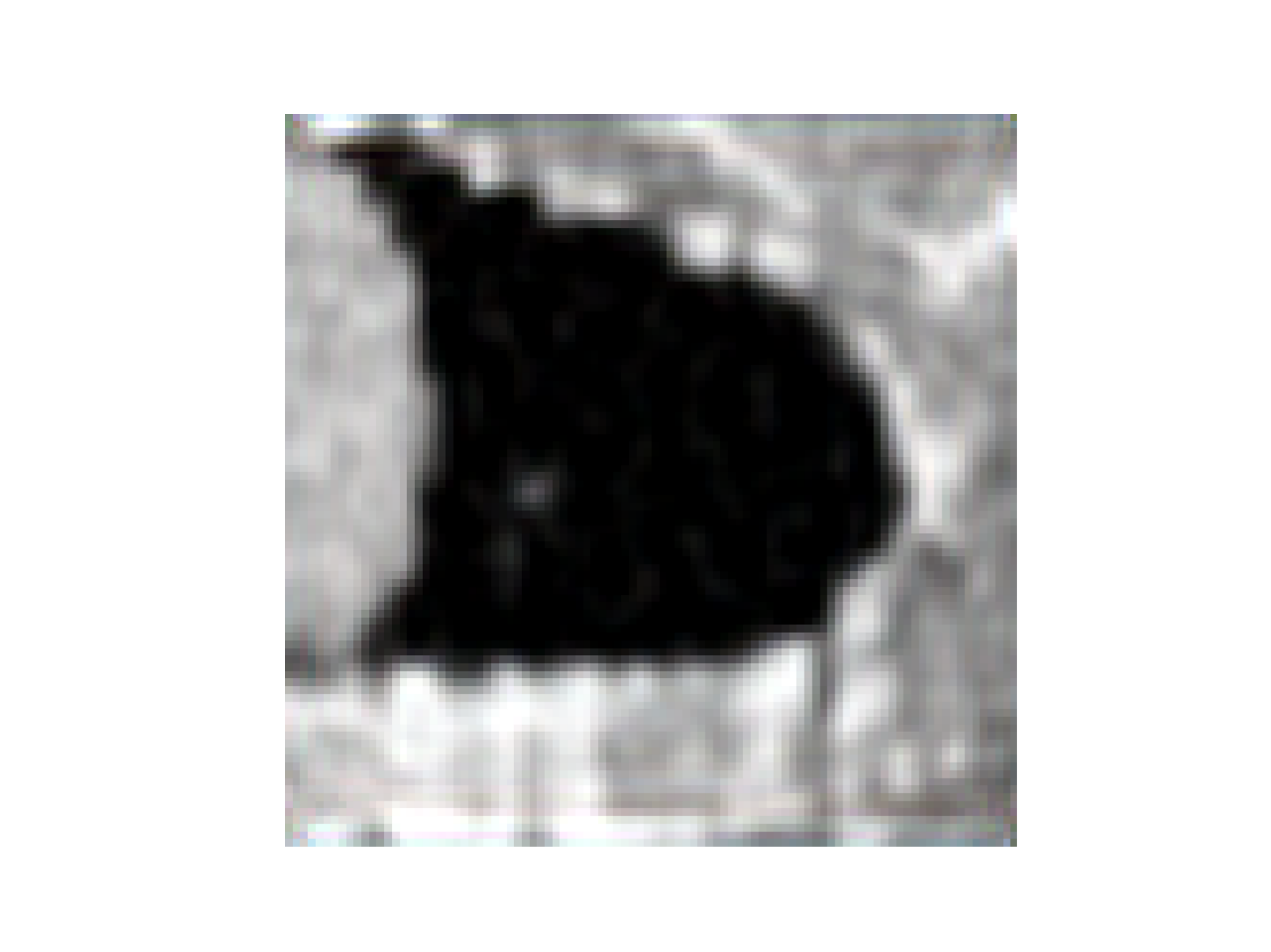} & \includegraphics[width=1.6cm, valign=c, trim={1cm 1cm 1cm 1cm},clip]{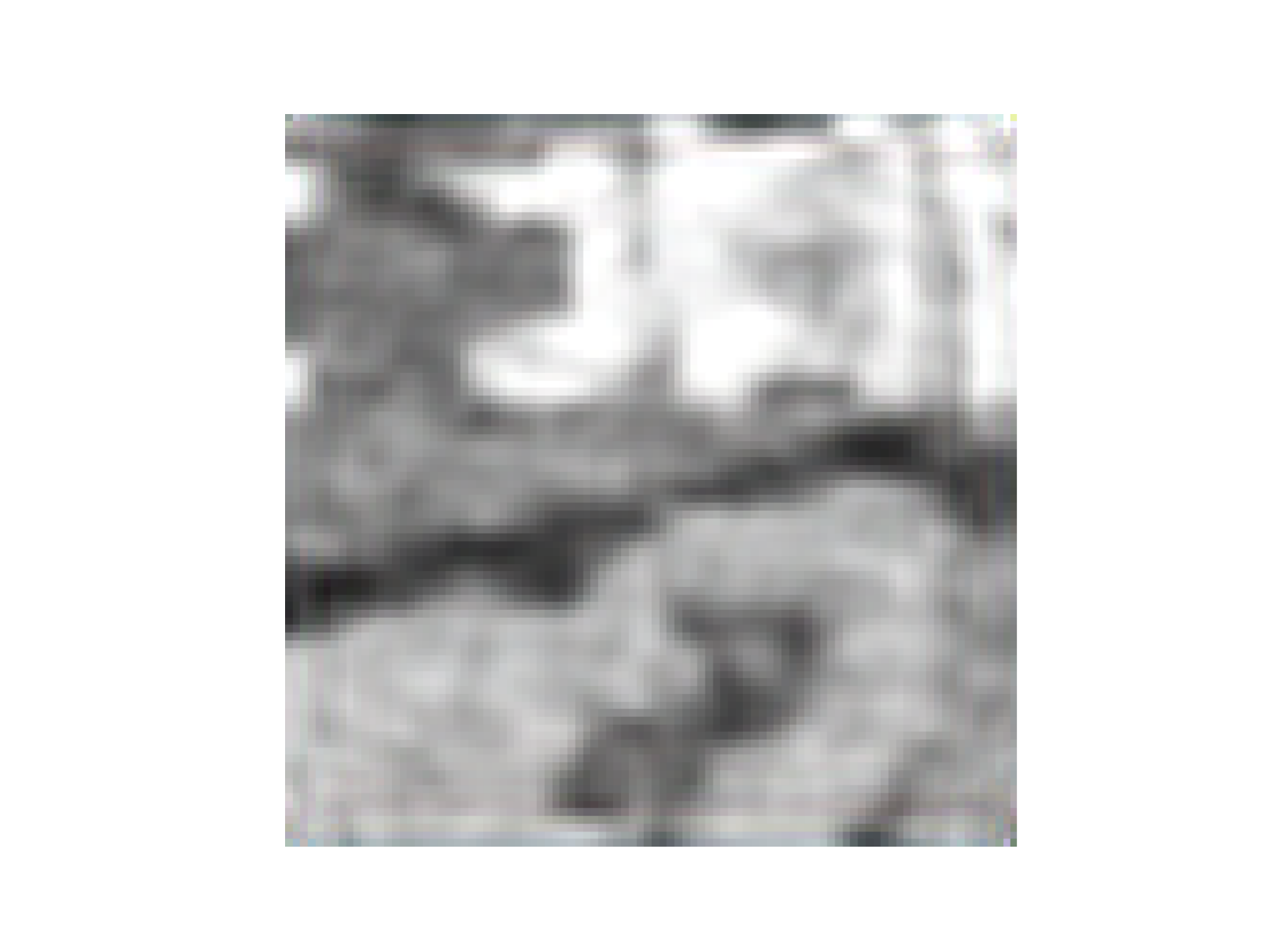} & \includegraphics[width=1.6cm, valign=c, trim={1cm 1cm 1cm 1cm},clip]{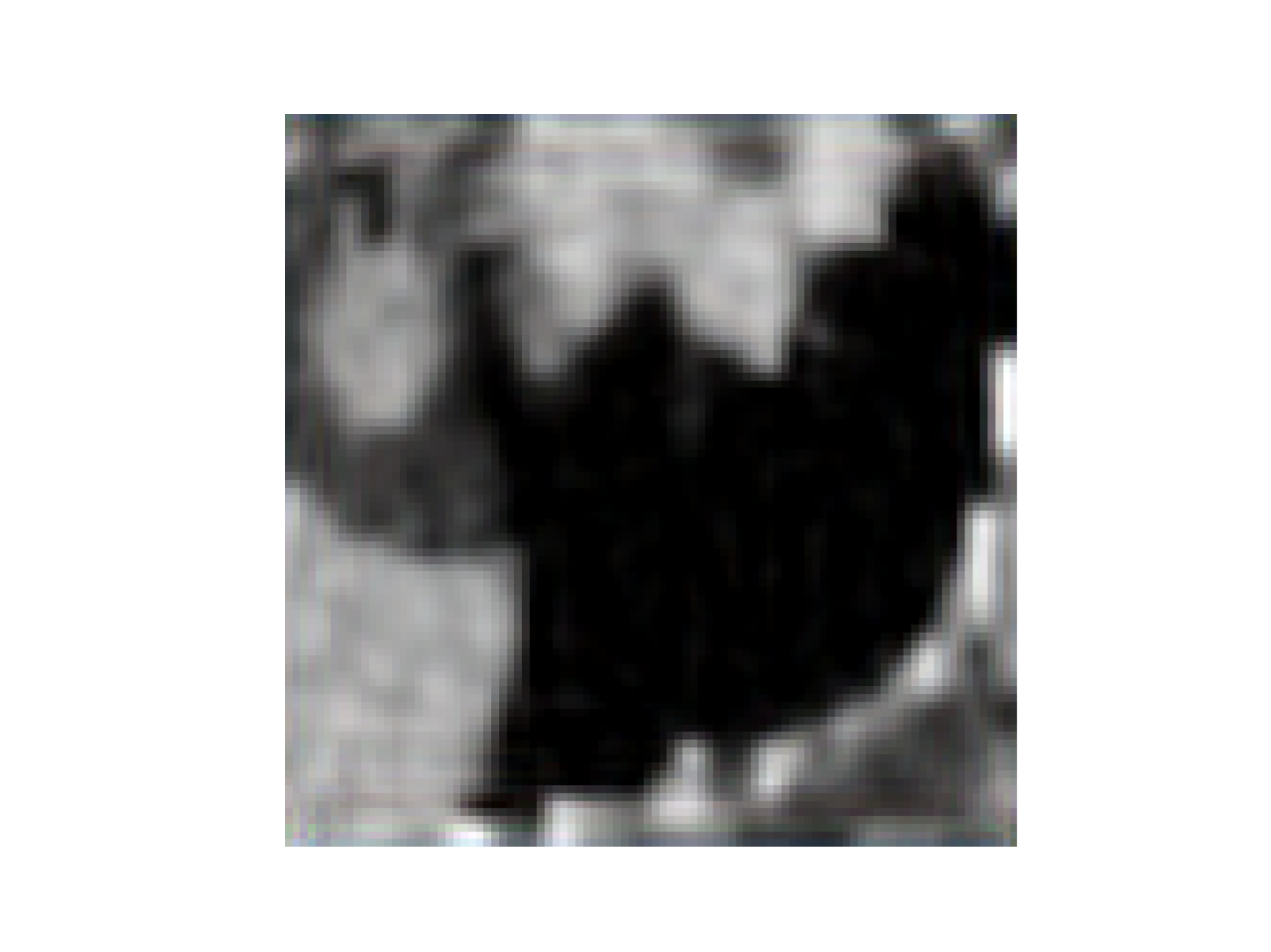} & \includegraphics[width=1.6cm, valign=c, trim={1cm 1cm 1cm 1cm},clip]{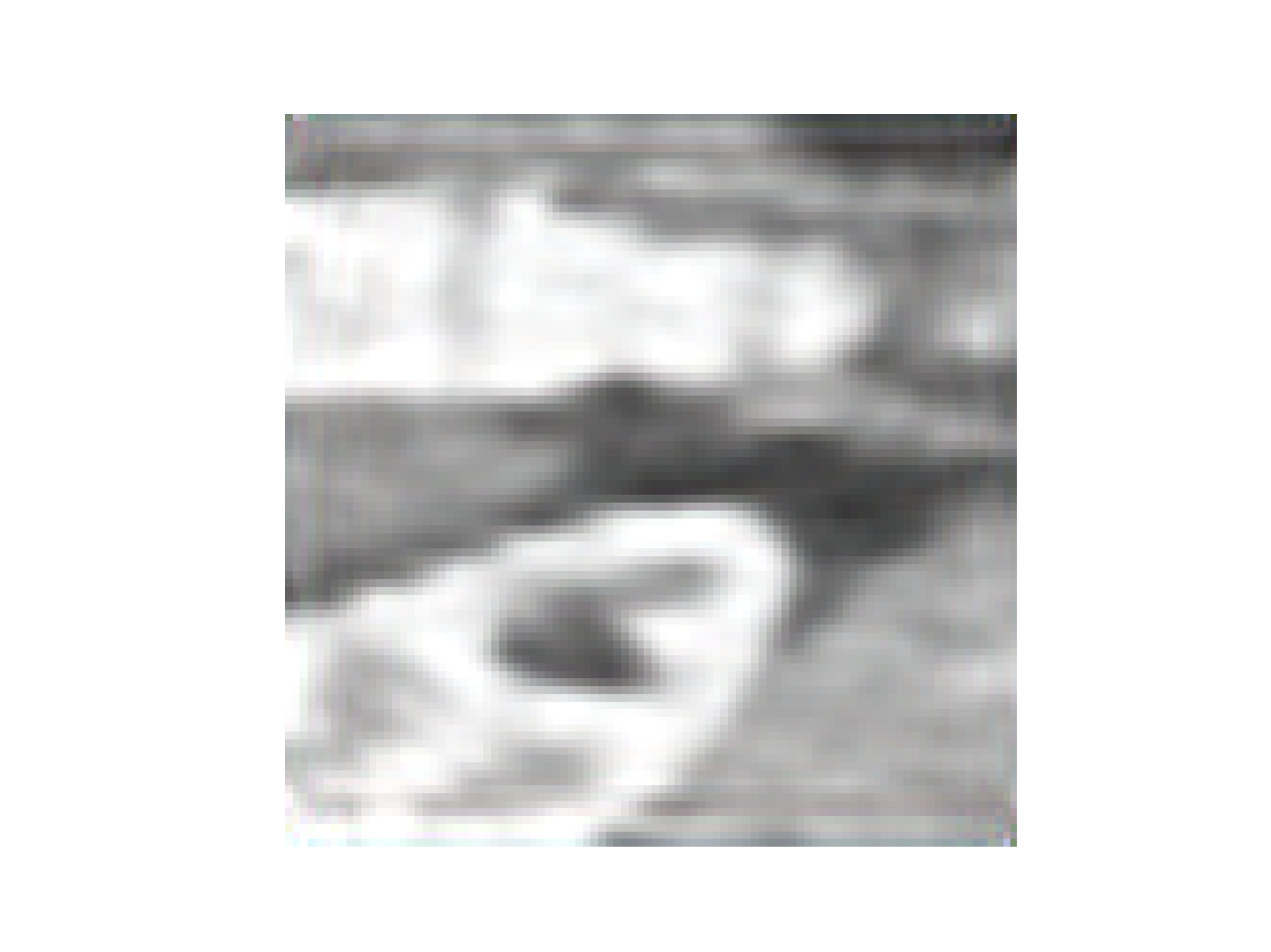}      \\
        \hline

            \tiny{{Neural-PGD}} & 
    \includegraphics[width=1.6cm, valign=c, trim={1cm 1cm 1cm 1cm},clip]{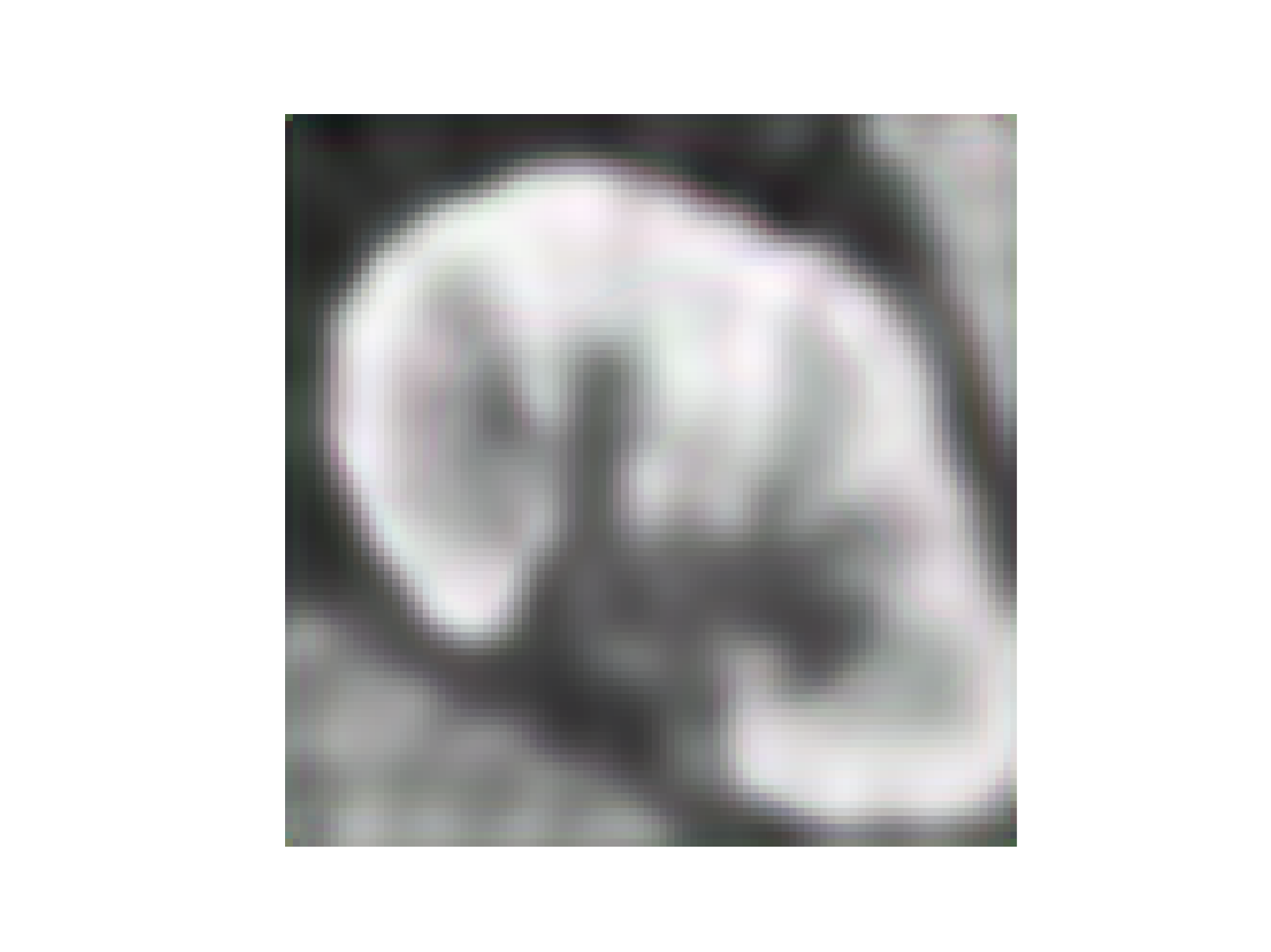}
     &
    \includegraphics[width=1.6cm, valign=c, trim={1cm 1cm 1cm 1cm},clip]{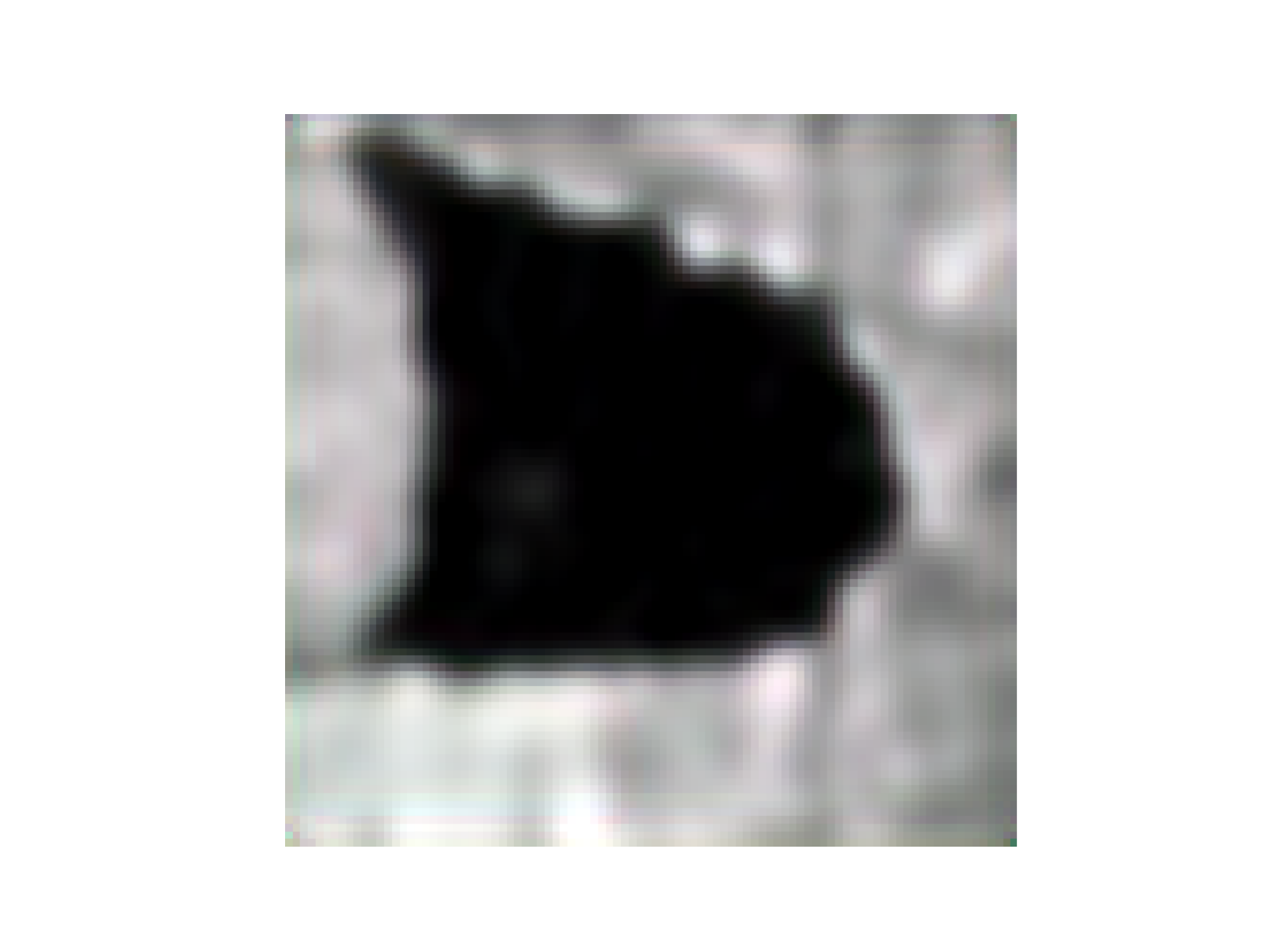} &
    \includegraphics[width=1.6cm, valign=c, trim={1cm 1cm 1cm 1cm},clip]{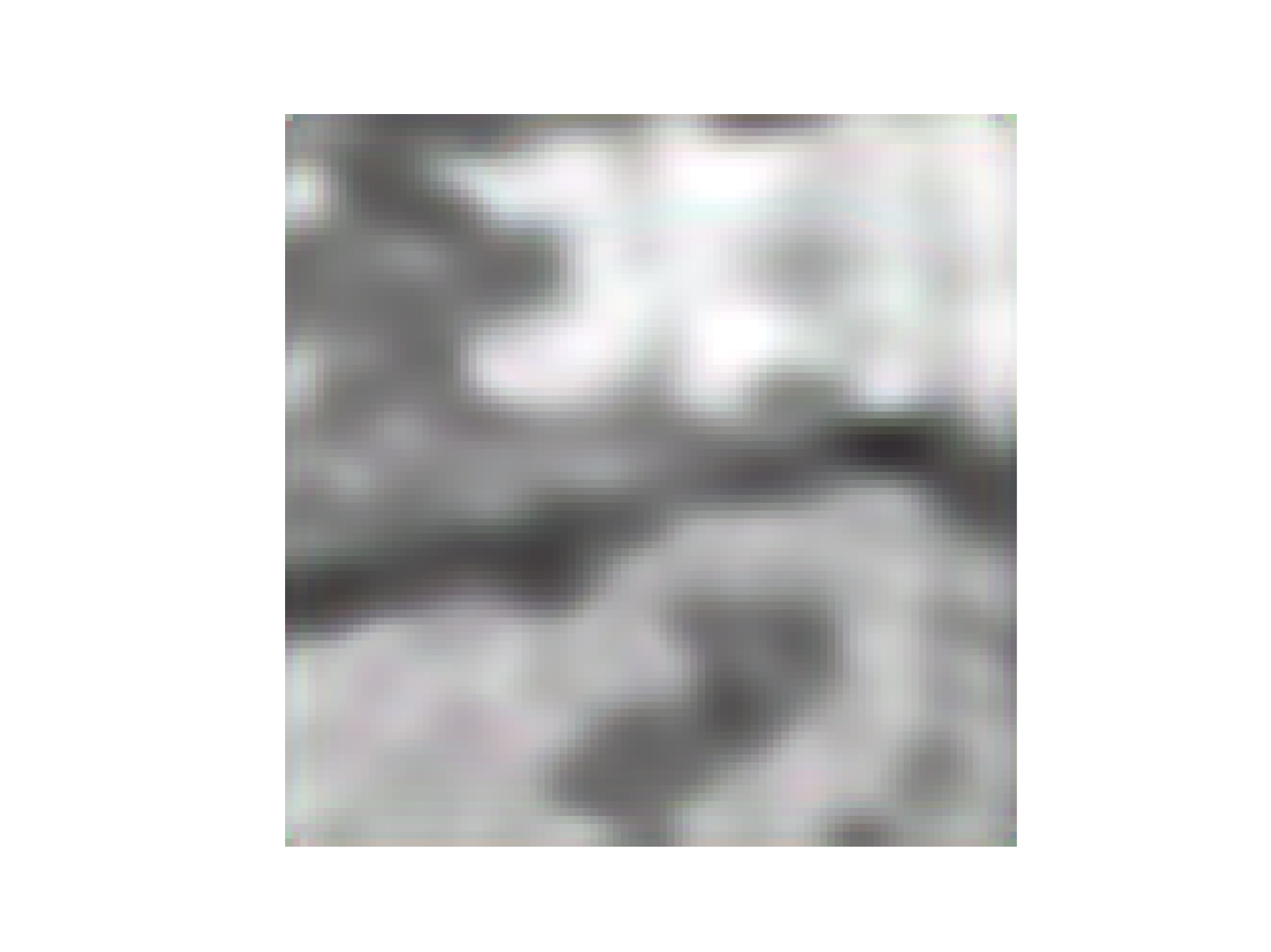} &
    \includegraphics[width=1.6cm, valign=c, trim={1cm 1cm 1cm 1cm},clip]{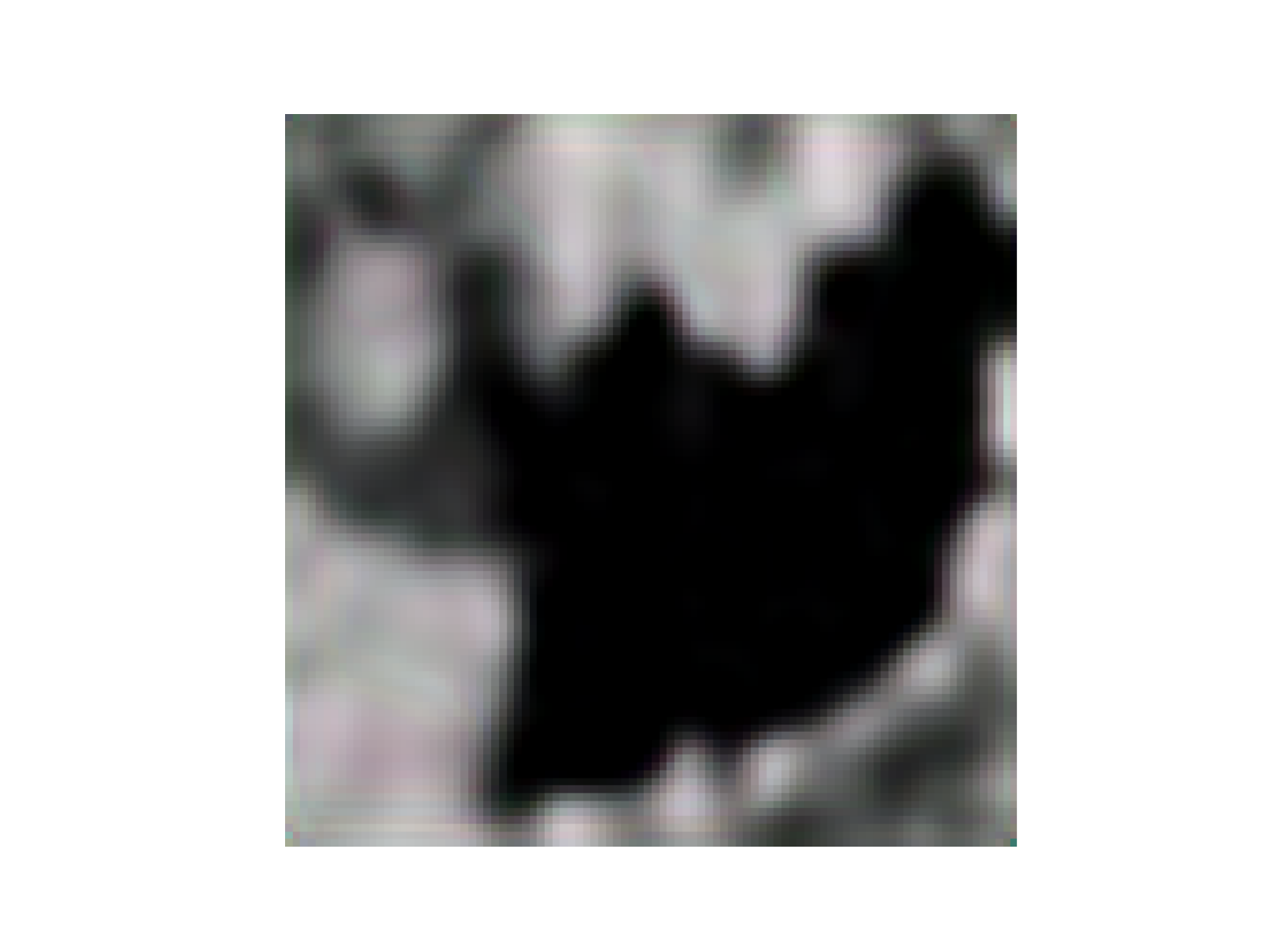} &
    \includegraphics[width=1.6cm, valign=c, trim={1cm 1cm 1cm 1cm},clip]{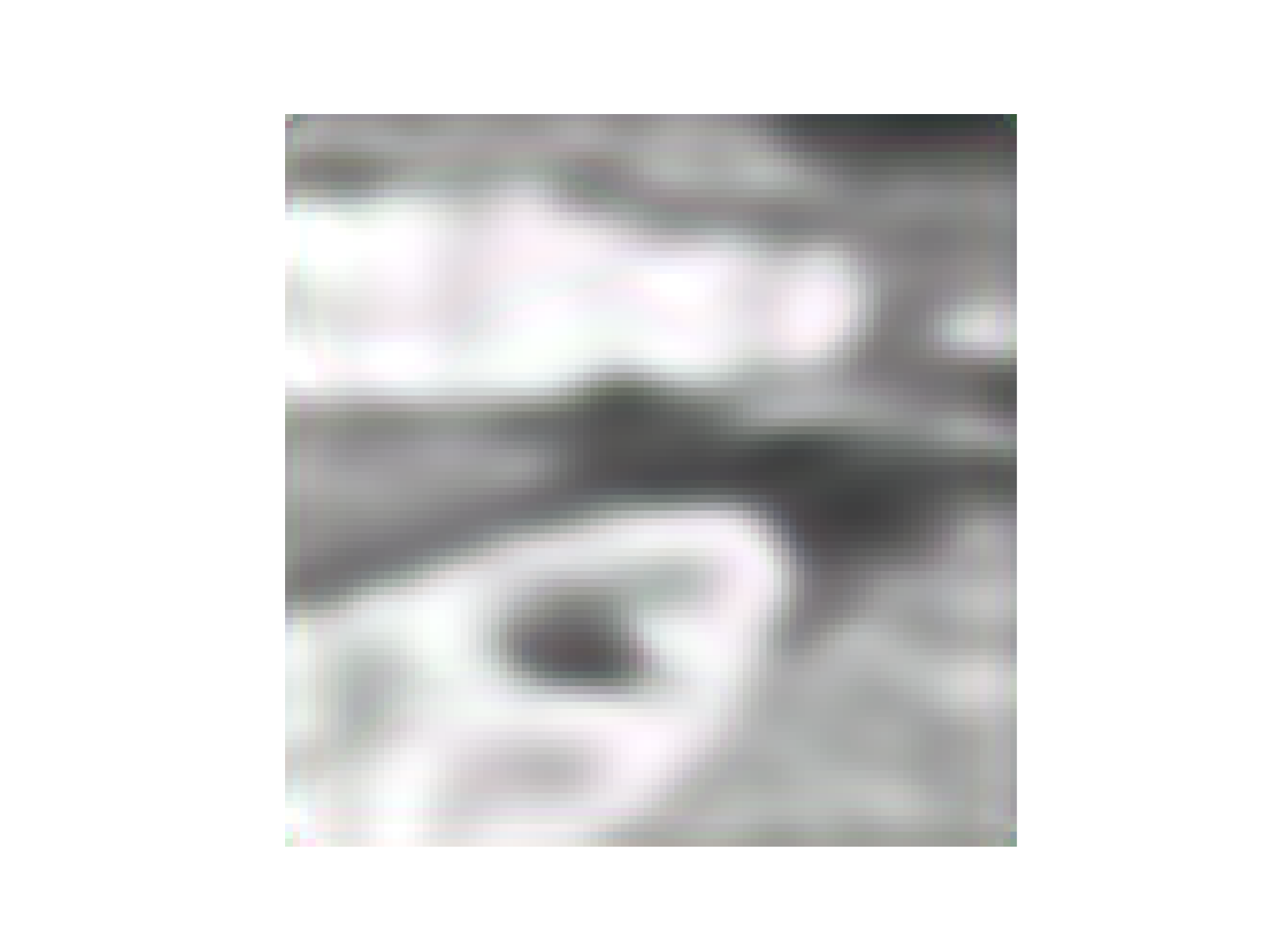}
    \\
        \hline

     \tiny{UNet} & \includegraphics[width=1.6cm, valign=c, trim={1cm 1cm 1cm 1cm},clip]{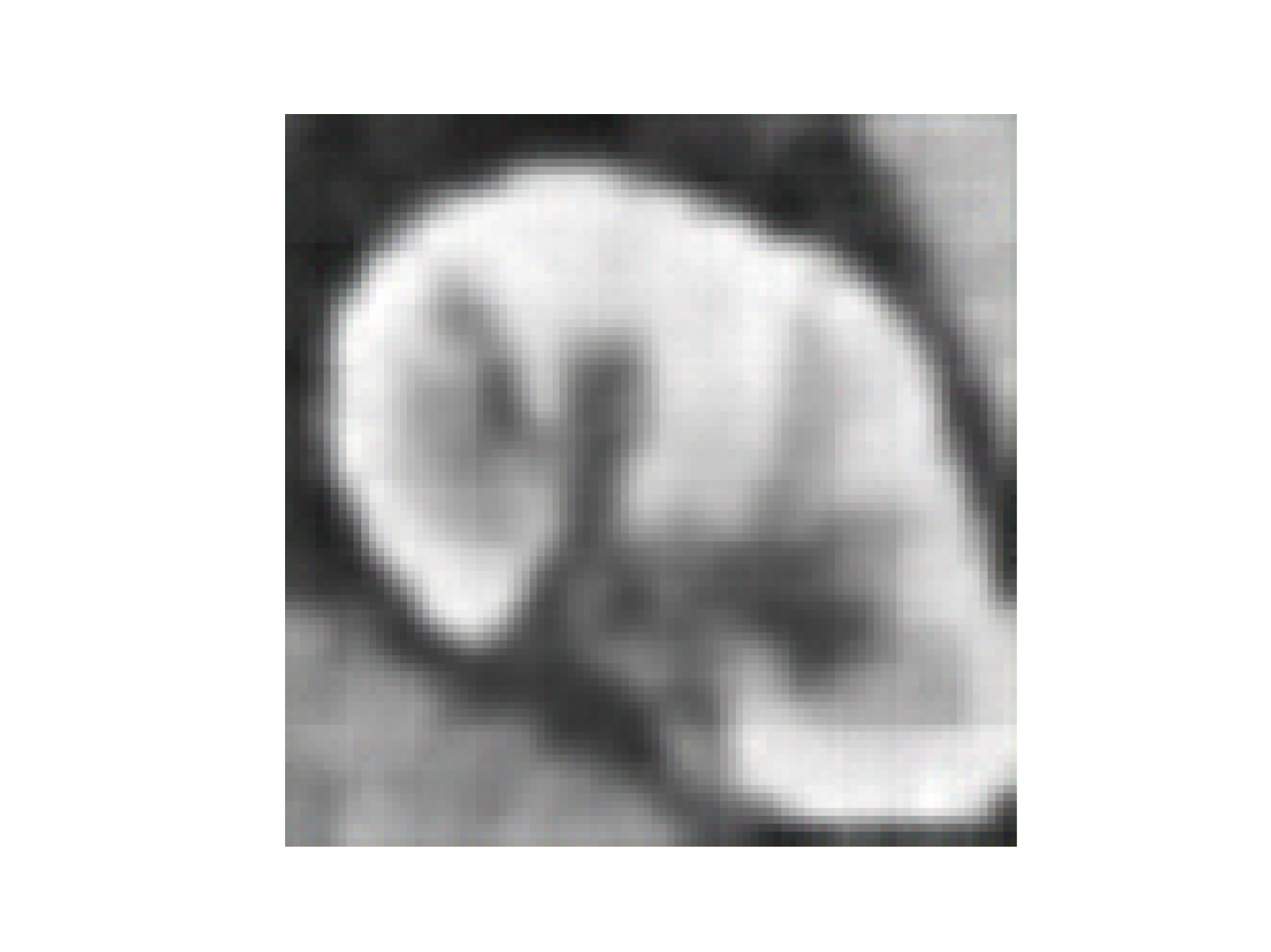}& \includegraphics[width=1.6cm, valign=c, trim={1cm 1cm 1cm 1cm},clip]{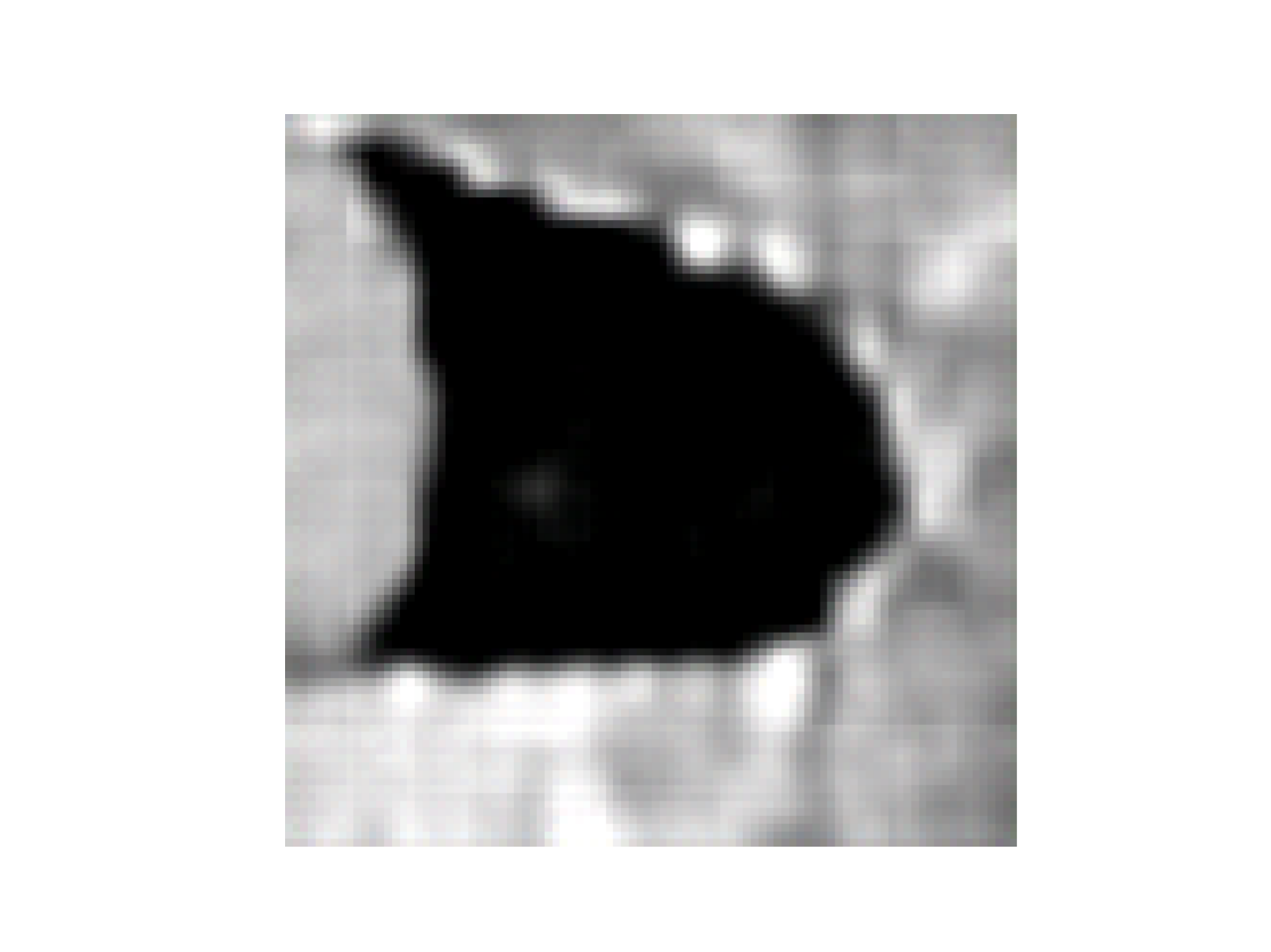}& \includegraphics[width=1.6cm, valign=c, trim={1cm 1cm 1cm 1cm},clip]{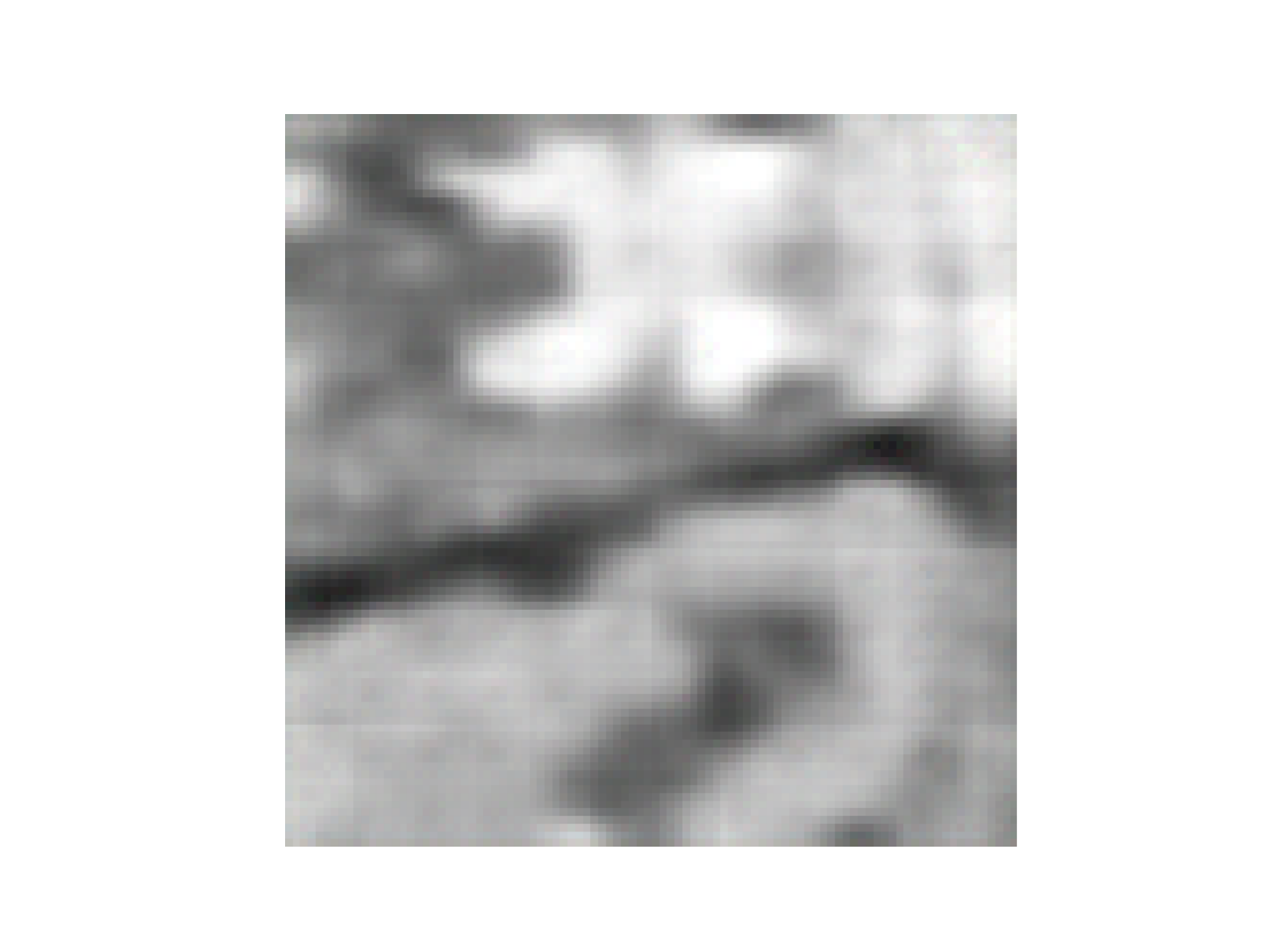}& \includegraphics[width=1.6cm, valign=c, trim={1cm 1cm 1cm 1cm},clip]{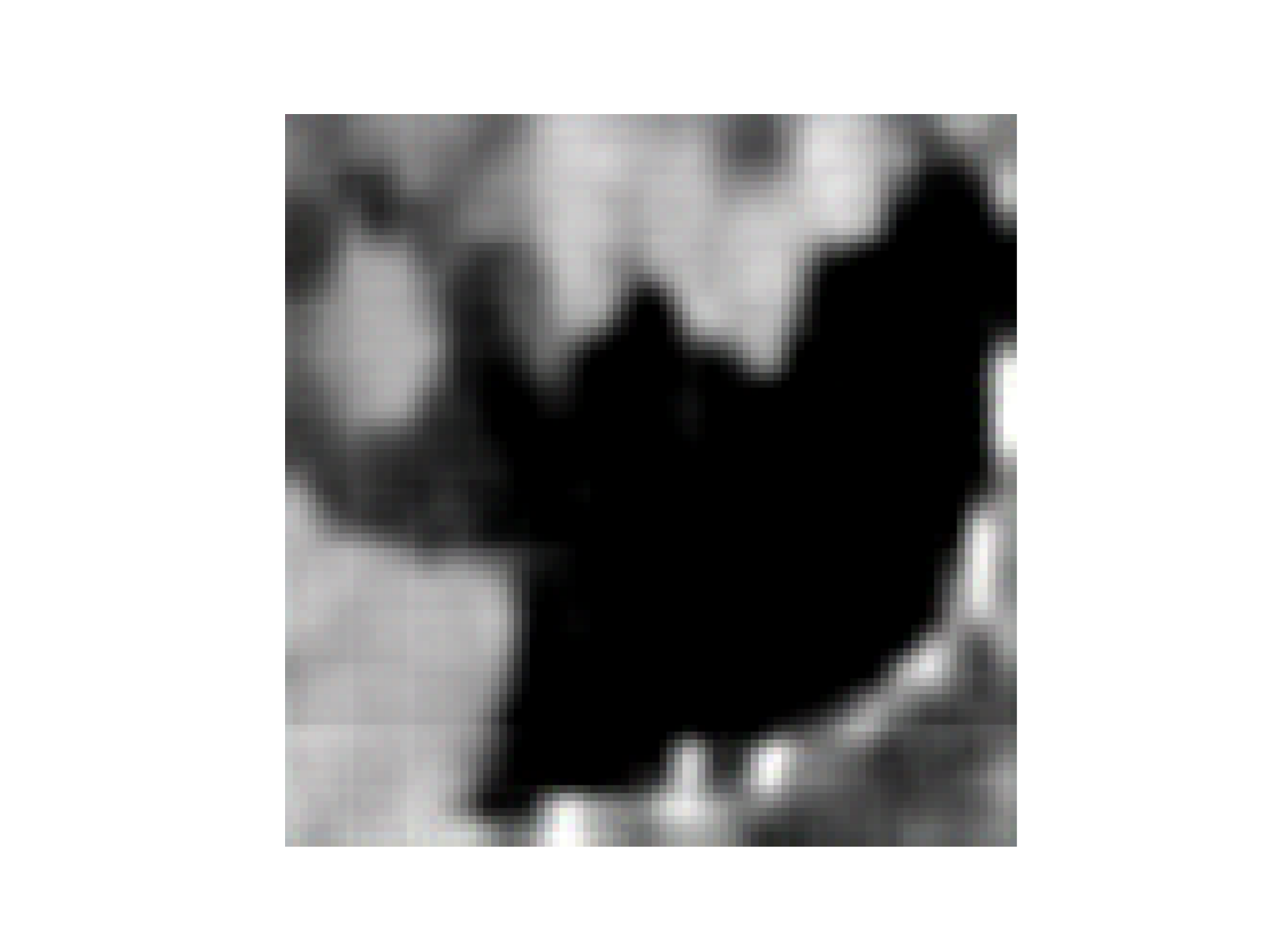}  & \includegraphics[width=1.6cm, valign=c, trim={1cm 1cm 1cm 1cm},clip]{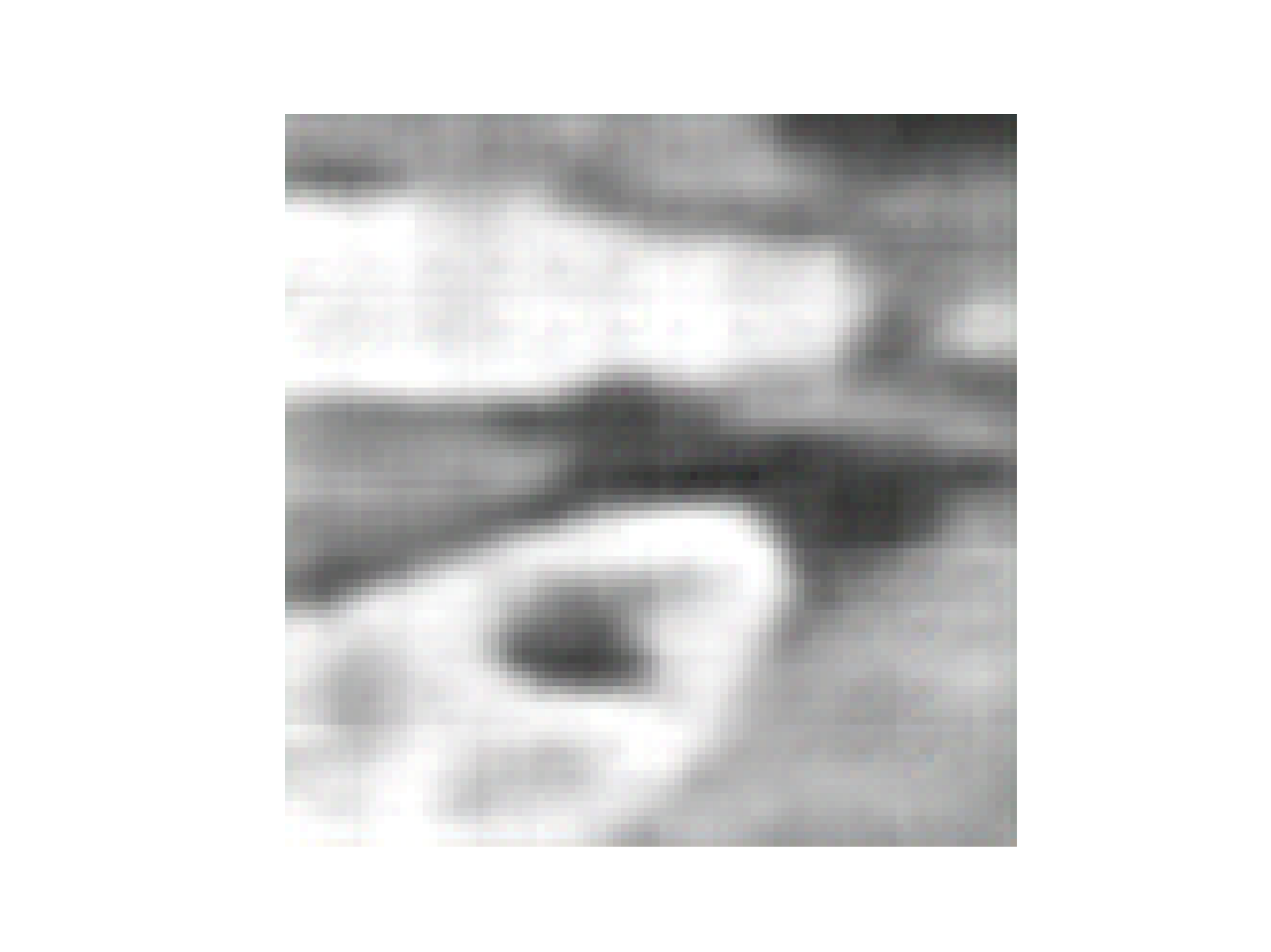}       \\
        \hline
     \tiny{ResNet}
     & \includegraphics[width=1.6cm, valign=c, trim={1cm 1cm 1cm 1cm},clip]{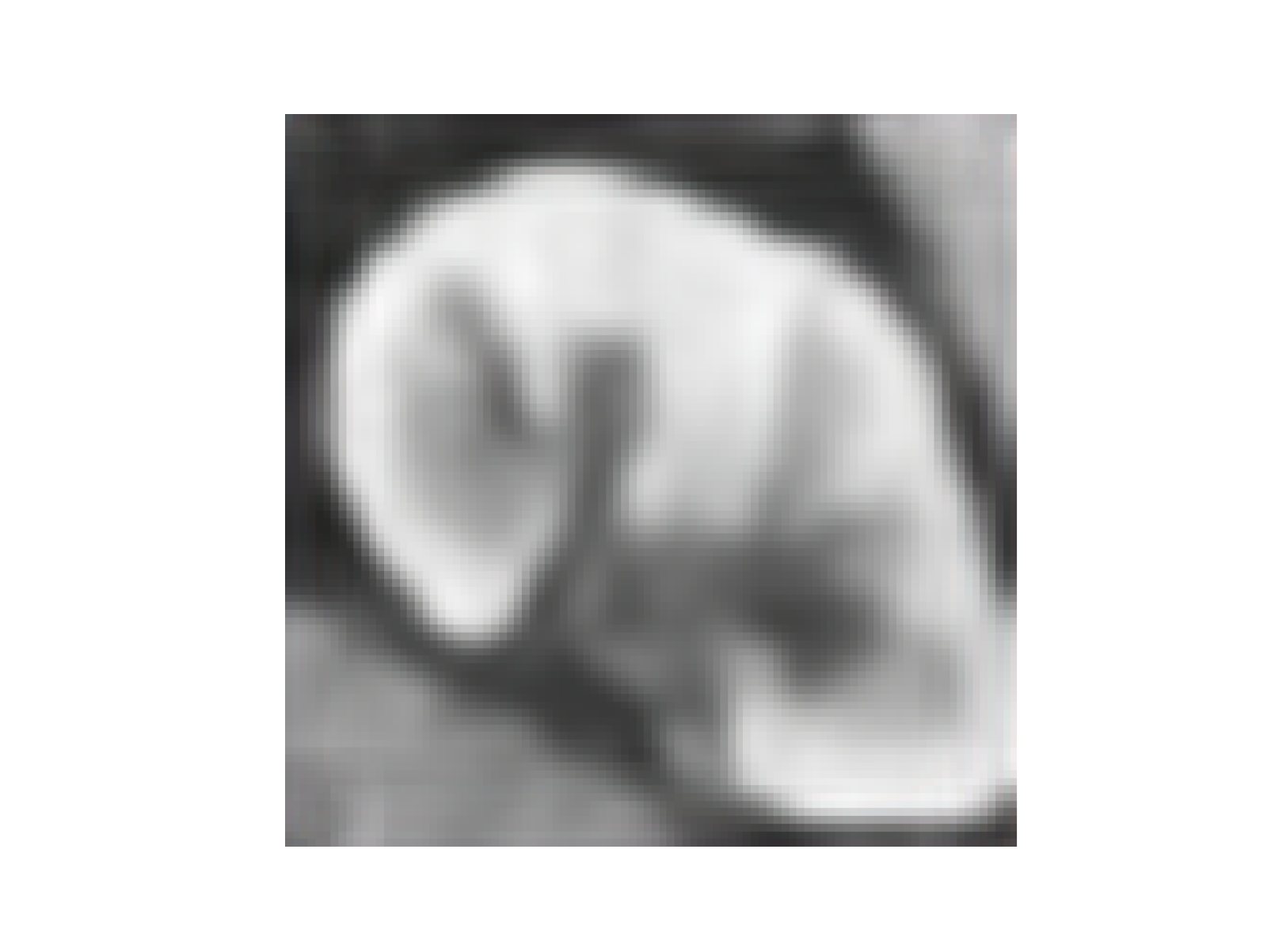}
     & \includegraphics[width=1.6cm, valign=c, trim={1cm 1cm 1cm 1cm},clip]{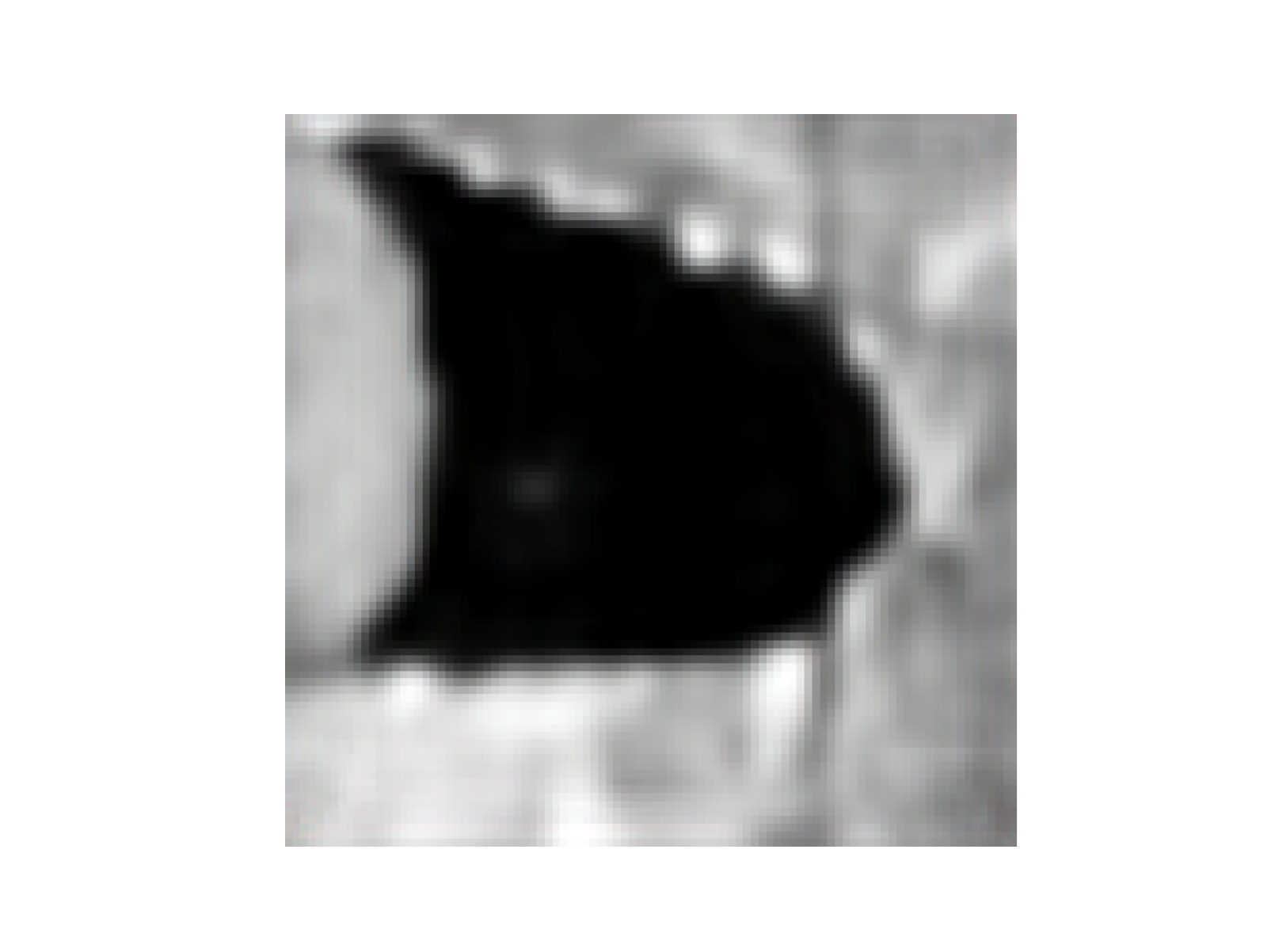}
     & \includegraphics[width=1.6cm, valign=c, trim={1cm 1cm 1cm 1cm},clip]{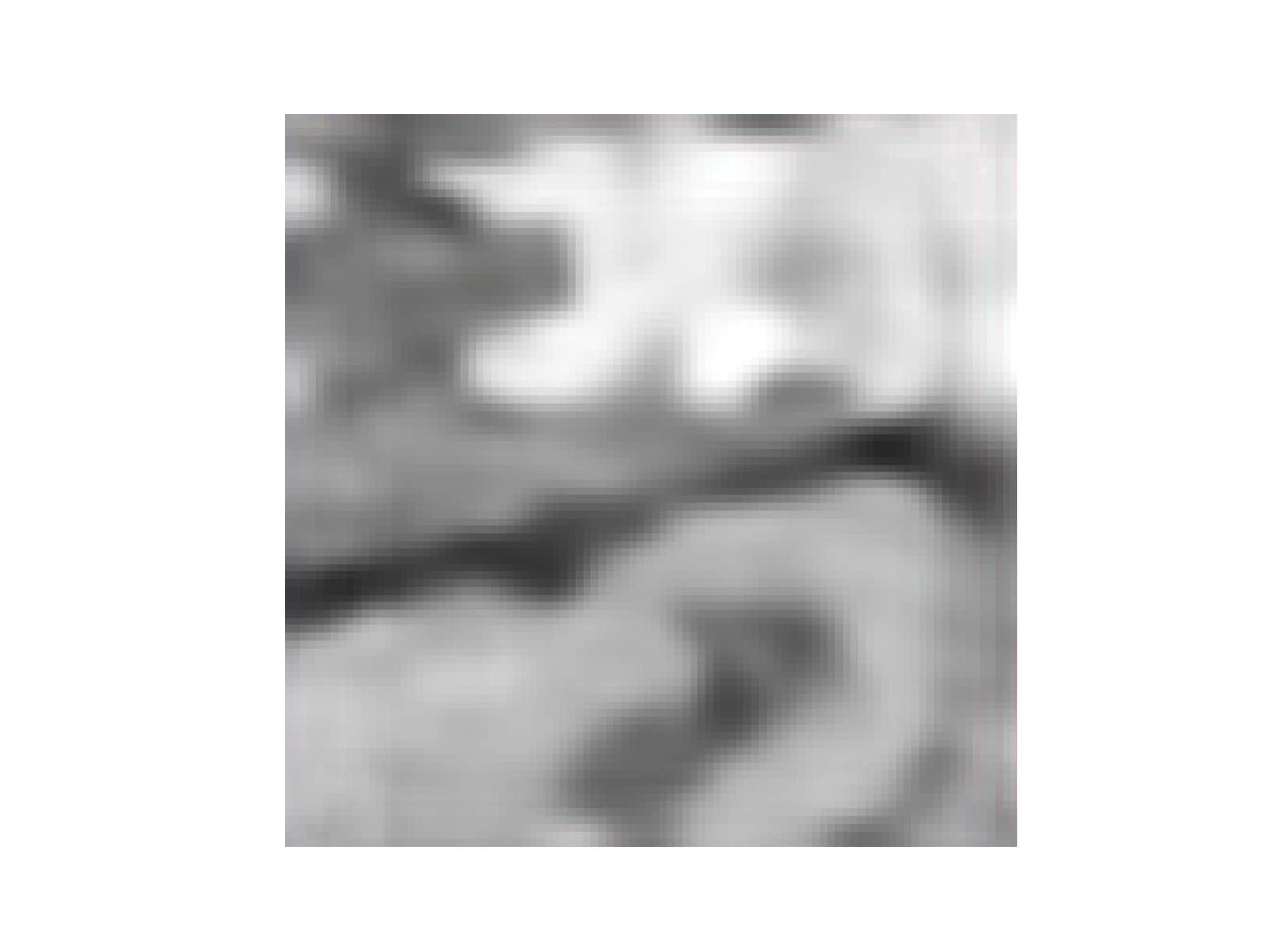}
     & \includegraphics[width=1.6cm, valign=c, trim={1cm 1cm 1cm 1cm},clip]{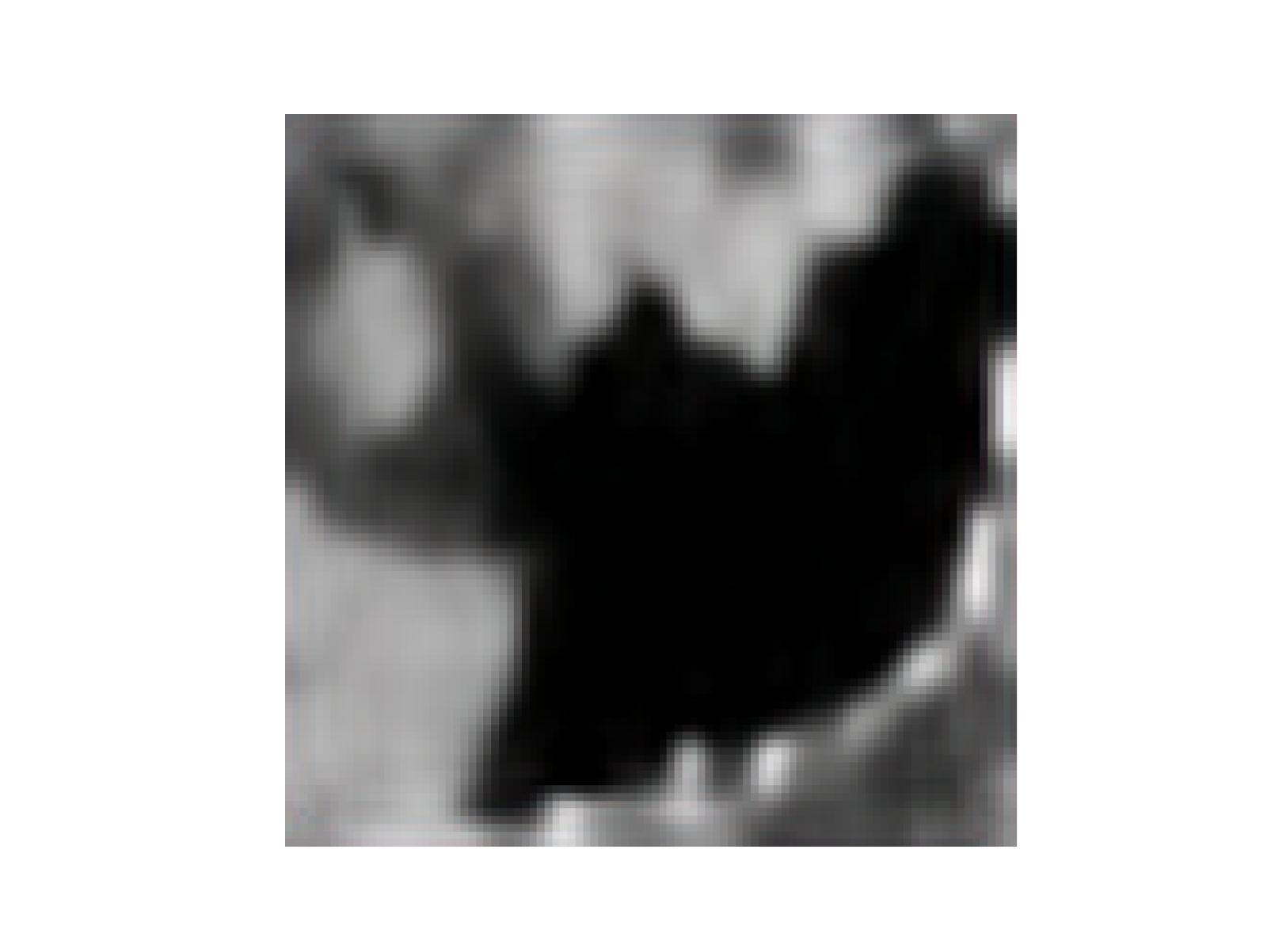}      & \includegraphics[width=1.6cm, valign=c, trim={1cm 1cm 1cm 1cm},clip]{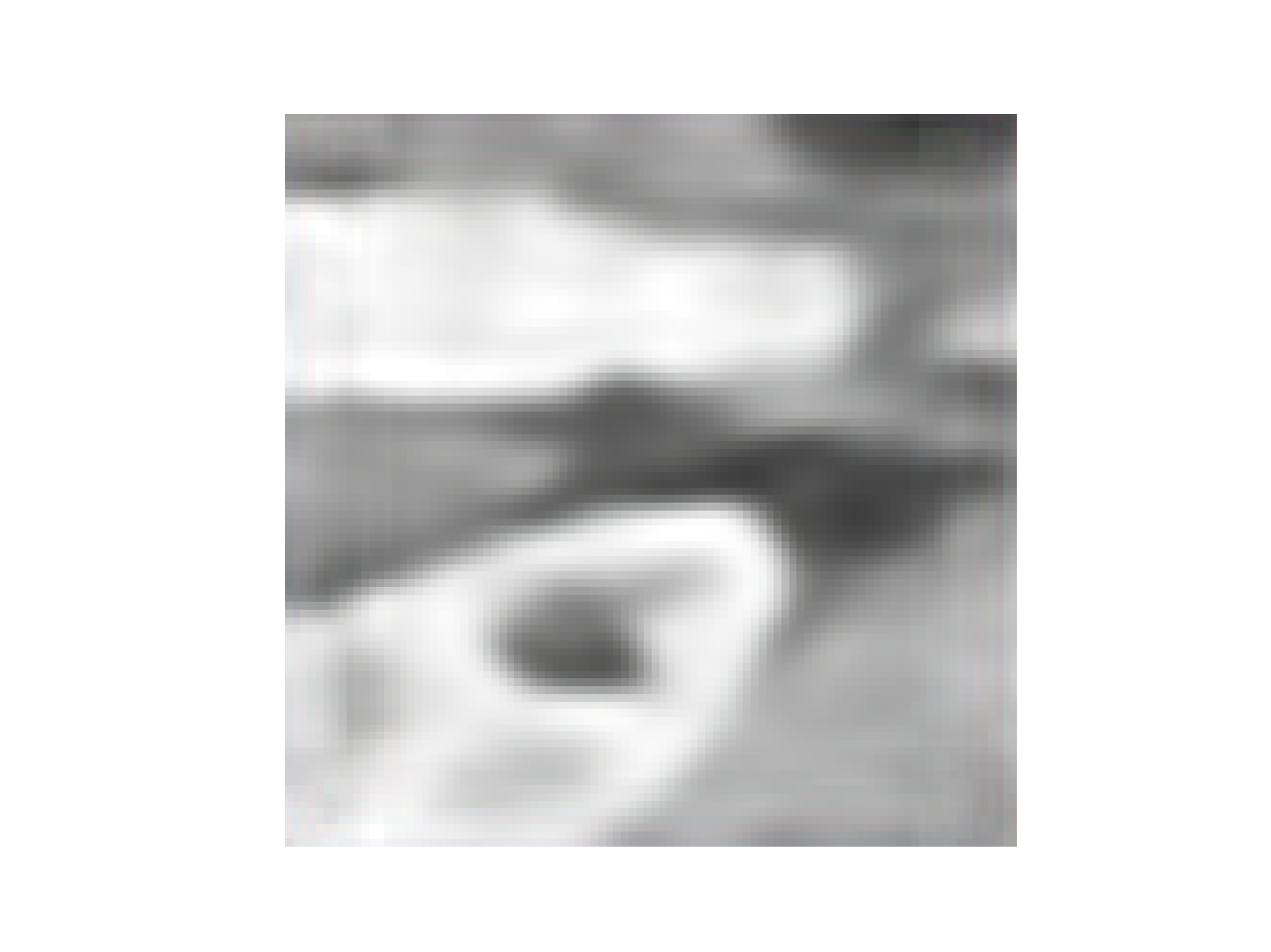}
     \\
     \hline
     \tiny{Ground-truth} & \includegraphics[width=1.6cm, valign=c, trim={1cm 1cm 1cm 1cm},clip]{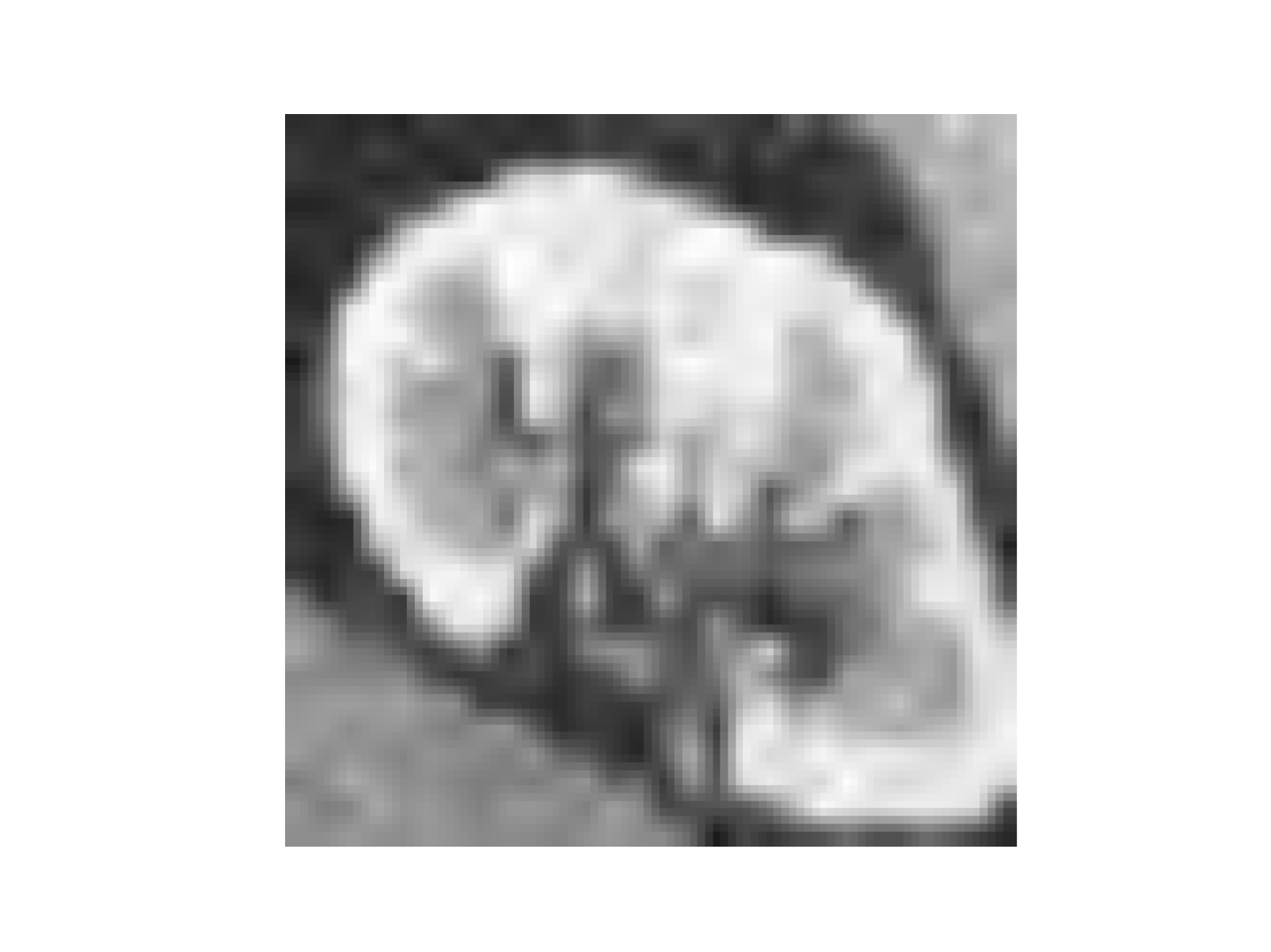} & \includegraphics[width=1.6cm, valign=c, trim={1cm 1cm 1cm 1cm},clip]{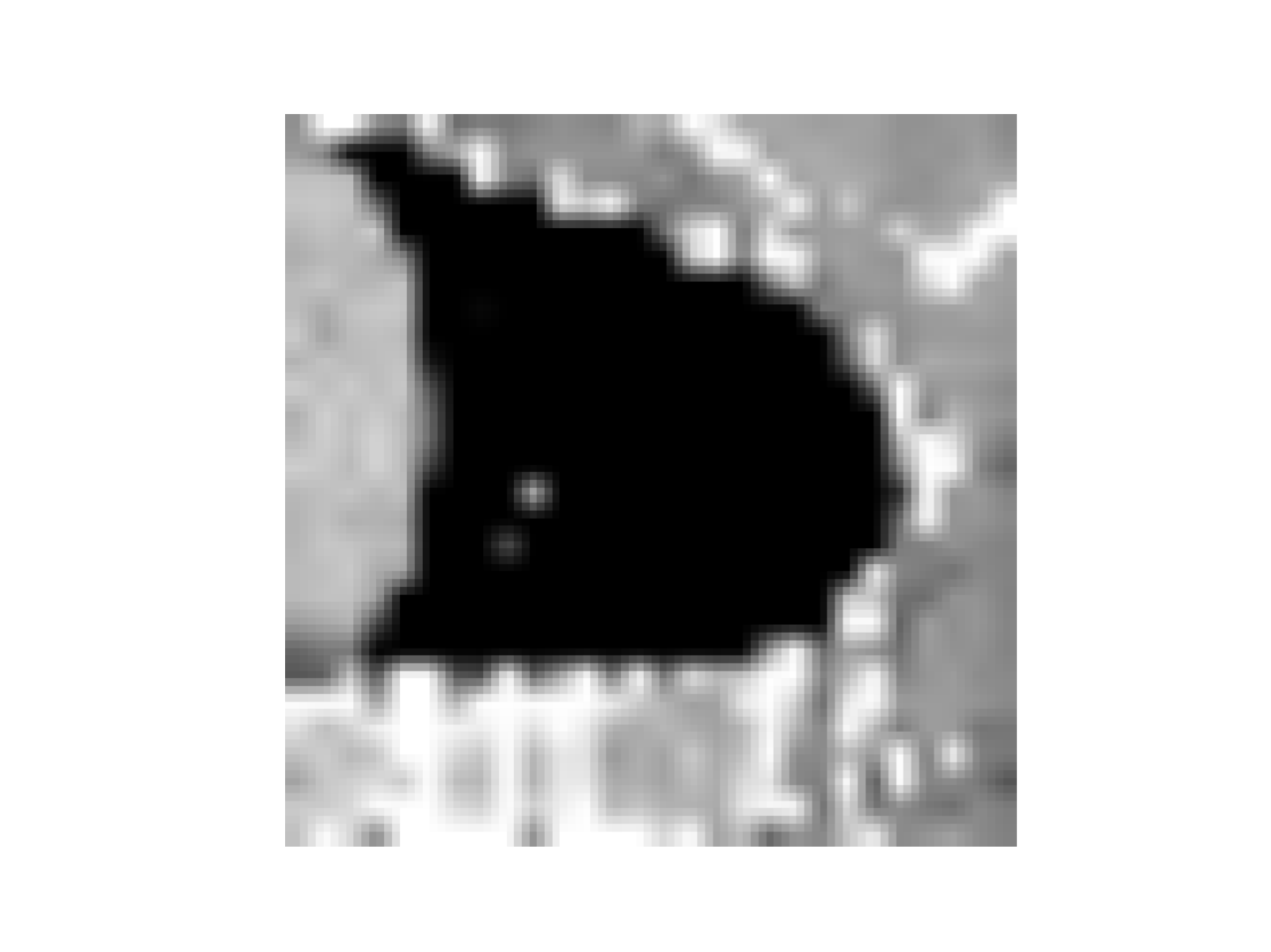}& \includegraphics[width=1.6cm, valign=c, trim={1cm 1cm 1cm 1cm},clip]{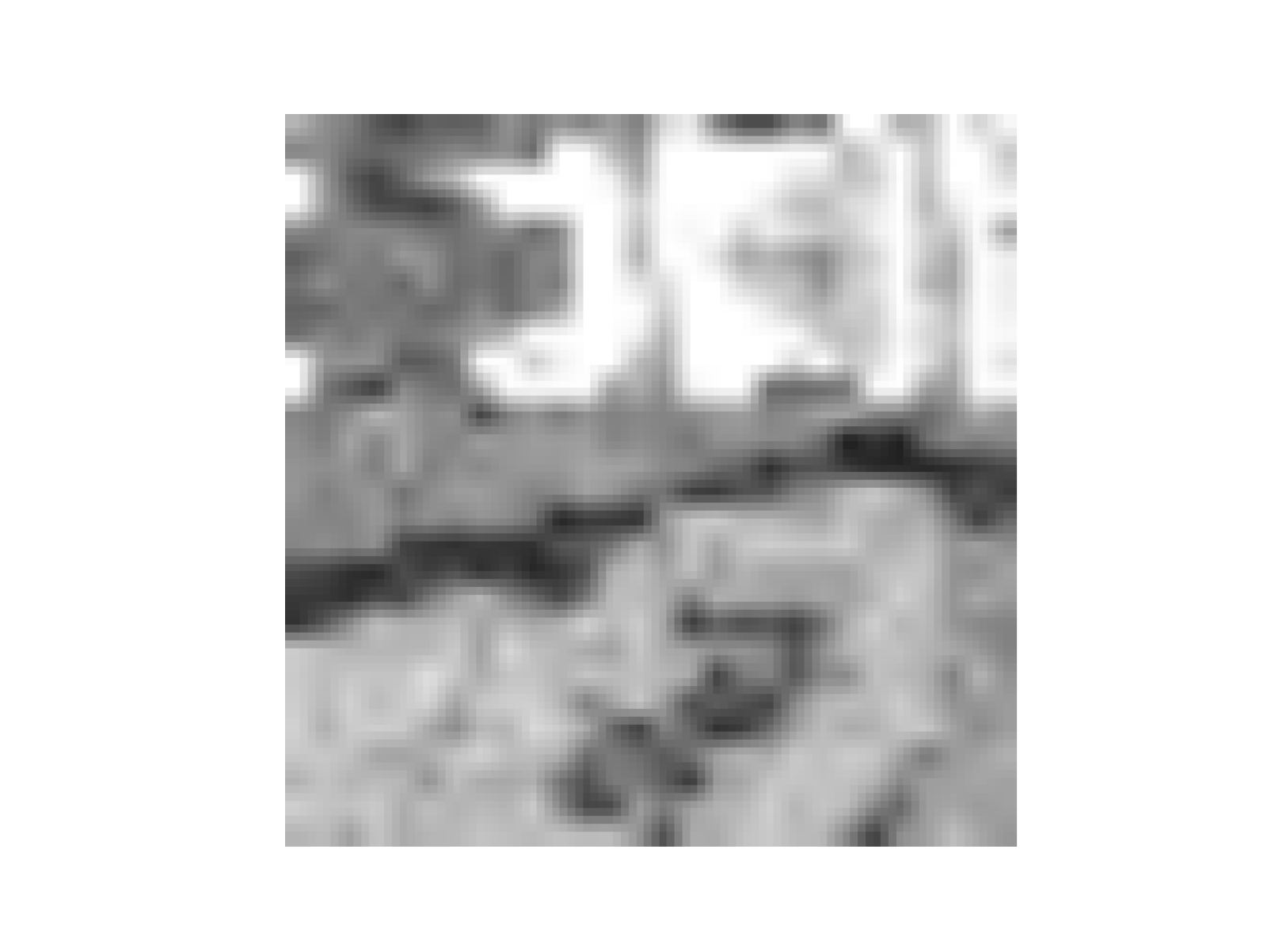} & \includegraphics[width=1.6cm, valign=c, trim={1cm 1cm 1cm 1cm},clip]{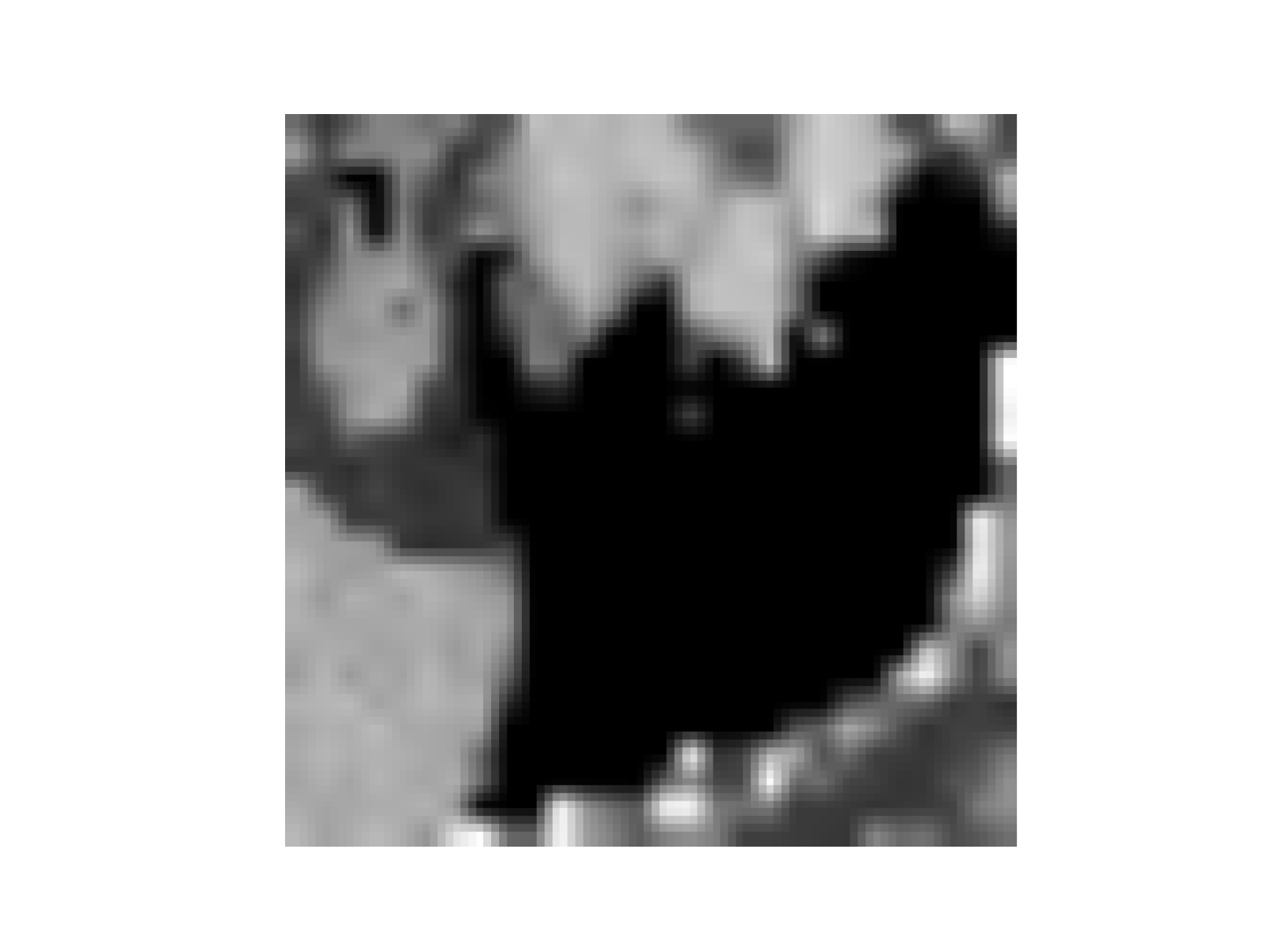} & \includegraphics[width=1.6cm, valign=c, trim={1cm 1cm 1cm 1cm},clip]{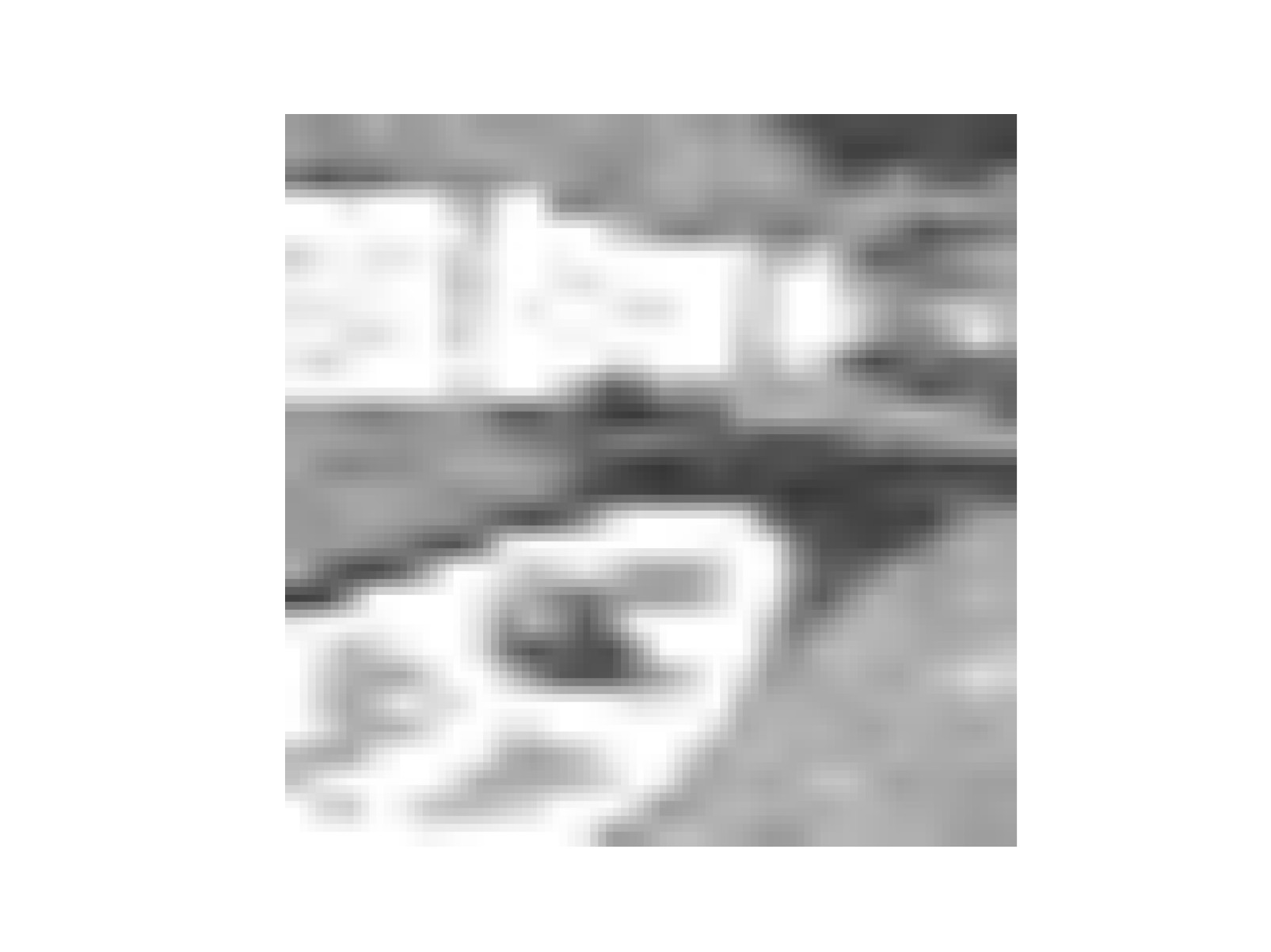}   \\
     \hline
    \end{tabular}
    \captionof{figure}{{Examples of recovered images from the Tomography task, trained and tested on the OrganCMNIST dataset}. While all models are able to reconstruct the images, our DRIP-based LA-Net and Hyper-ResNet architectures yield images with the lowest residual and error terms.}
    \label{fig:tomography_results_ct}
\end{table}

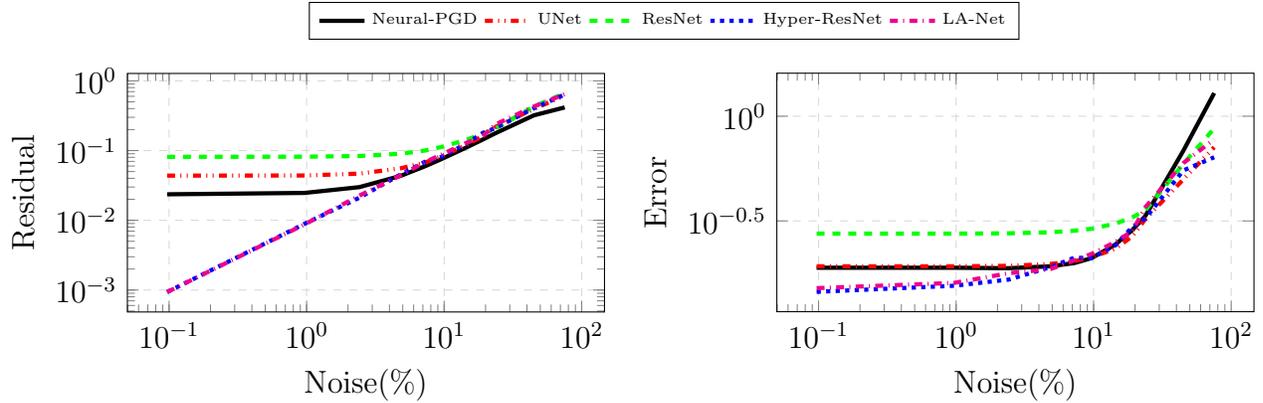
\begin{figure*}[h]
\centering
\begin{minipage}{0.48\textwidth}
\centering
\begin{tikzpicture}[define rgb/.code={\definecolor{mycolor}{RGB}{#1}},
                    rgb color/.style={define rgb={#1},mycolor}]
  \begin{axis}[
      width=1.0\linewidth, 
      height=0.6\linewidth,
      grid=major,
      grid style={dashed,gray!30},
      xlabel=Noise(\%),
      ylabel=Residual,
      ylabel near ticks,
      legend style={at={(1.125,1.3)},anchor=north,scale=1.0, cells={anchor=west}, font=\tiny,},
      legend columns=-1,
      yticklabel style={
        /pgf/number format/fixed,
        /pgf/number format/precision=3
      },
      scaled y ticks=false,
      every axis plot post/.style={ultra thick},
      xmode=log,
      ymode=log
    ]

    \addplot[color=black]
    table[x=noise,y=pgdresidual,col sep=comma] {data/ip_tomography.csv};
    
    \addplot[color=red, style=dashdotdotted]
table[x=noise,y=unetresidual,col sep=comma] {data/ip_tomography.csv};
    \addplot[color=green, style=dashed]
    table[x=noise,y=resnetresidual
    ,col sep=comma] {data/ip_tomography.csv};

    \addplot[color=blue, style=dotted]
    table[x=noise,y=hyperresidual
    ,col sep=comma] {data/ip_tomography.csv}; 
    
    \addplot[color=magenta, style=dashdotted]
    table[x=noise,y=laresidual
    ,col sep=comma] {data/ip_tomography.csv}; 
    \legend{{Neural-PGD}, UNet, ResNet, Hyper-ResNet, LA-Net}
    \end{axis}
\end{tikzpicture}
\end{minipage} \hspace{0.5em}
\begin{minipage}{.48\textwidth}
\centering
\begin{tikzpicture}[define rgb/.code={\definecolor{mycolor}{RGB}{#1}},
                    rgb color/.style={define rgb={#1},mycolor}]
  \begin{axis}[
      width=1.0\linewidth, 
      height=0.6\linewidth,
      grid=major,
      grid style={dashed,gray!30},
      xlabel=Noise(\%),
      ylabel=Error,
      ylabel near ticks,
      legend style={at={(10,1.3)},anchor=north,scale=1.0, draw=none, cells={anchor=west}, font=\tiny, fill=none},
      legend columns=-1,
      yticklabel style={
        /pgf/number format/fixed,
        /pgf/number format/precision=3
      },
      scaled y ticks=false,
      every axis plot post/.style={ultra thick},
      xmode=log,
      ymode=log
    ]

    \addplot[color=black]
    table[x=noise,y=pgderror,col sep=comma] {data/ip_tomography.csv};
    
    \addplot[color=red, style=dashdotdotted]
    table[x=noise,y=uneterror,col sep=comma] {data/ip_tomography.csv};
    \addplot[color=green, style=dashed]
    table[x=noise,y=resneterror
    ,col sep=comma] {data/ip_tomography.csv};

    \addplot[color=blue, style=dotted]
    table[x=noise,y=hypererror
    ,col sep=comma] {data/ip_tomography.csv}; 
    
    \addplot[color=magenta, style=dashdotted]
    table[x=noise,y=laerror
    ,col sep=comma] {data/ip_tomography.csv};

    \addlegendimage{}
    \addlegendimage{}
    \addlegendimage{}
    \addlegendentry{}
    \end{axis}
\end{tikzpicture}
\end{minipage}
\caption{A comparison of the obtained residual and errors vs. noise in the Tomography task on {STL-10 test set.}
}
\label{fig:residual_and_error_tomography}

\end{figure*}

\begin{figure*}[h]
\centering
\begin{minipage}{0.48\textwidth}
\centering
\begin{tikzpicture}[define rgb/.code={\definecolor{mycolor}{RGB}{#1}},
                    rgb color/.style={define rgb={#1},mycolor}]
  \begin{axis}[
      width=1.0\linewidth, 
      height=0.6\linewidth,
      grid=major,
      grid style={dashed,gray!30},
      xlabel=Noise(\%),
      ylabel=Residual,
      ylabel near ticks,
      legend style={at={(1.125,1.3)},anchor=north,scale=1.0, cells={anchor=west}, font=\tiny,},
      legend columns=-1,
      yticklabel style={
        /pgf/number format/fixed,
        /pgf/number format/precision=3
      },
      scaled y ticks=false,
      every axis plot post/.style={ultra thick},
      xmode=log,
      ymode=log
    ]

    \addplot[color=black]
    table[x=noise,y=pgdresidual,col sep=comma] {data/ip_tomography_ctmnist.csv};
    
    \addplot[color=red, style=dashdotdotted]
table[x=noise,y=unetresidual,col sep=comma] {data/ip_tomography_ctmnist.csv};
    \addplot[color=green, style=dashed]
    table[x=noise,y=resnetresidual
    ,col sep=comma] {data/ip_tomography_ctmnist.csv};

    \addplot[color=blue, style=dotted]
    table[x=noise,y=hyperresidual
    ,col sep=comma] {data/ip_tomography_ctmnist.csv}; 
    
    \addplot[color=magenta, style=dashdotted]
    table[x=noise,y=laresidual
    ,col sep=comma] {data/ip_tomography_ctmnist.csv}; 
    \legend{{Neural-PGD}, UNet, ResNet, Hyper-ResNet, LA-Net}
    \end{axis}
\end{tikzpicture}
\end{minipage} \hspace{0.5em}
\begin{minipage}{.48\textwidth}
\centering
\begin{tikzpicture}[define rgb/.code={\definecolor{mycolor}{RGB}{#1}},
                    rgb color/.style={define rgb={#1},mycolor}]
  \begin{axis}[
      width=1.0\linewidth, 
      height=0.6\linewidth,
      grid=major,
      grid style={dashed,gray!30},
      xlabel=Noise(\%),
      ylabel=Error,
      ylabel near ticks,
      legend style={at={(10,1.3)},anchor=north,scale=1.0, draw=none, cells={anchor=west}, font=\tiny, fill=none},
      legend columns=-1,
      yticklabel style={
        /pgf/number format/fixed,
        /pgf/number format/precision=3
      },
      scaled y ticks=false,
      every axis plot post/.style={ultra thick},
      xmode=log,
      ymode=log
    ]

    \addplot[color=black]
    table[x=noise,y=pgderror,col sep=comma] {data/ip_tomography_ctmnist.csv};
    
    \addplot[color=red, style=dashdotdotted]
    table[x=noise,y=uneterror,col sep=comma] {data/ip_tomography_ctmnist.csv};
    \addplot[color=green, style=dashed]
    table[x=noise,y=resneterror
    ,col sep=comma] {data/ip_tomography_ctmnist.csv};

    \addplot[color=blue, style=dotted]
    table[x=noise,y=hypererror
    ,col sep=comma] {data/ip_tomography_ctmnist.csv}; 
    
    \addplot[color=magenta, style=dashdotted]
    table[x=noise,y=laerror
    ,col sep=comma] {data/ip_tomography_ctmnist.csv};

    \addlegendimage{}
    \addlegendimage{}
    \addlegendimage{}
    \addlegendentry{}
    \end{axis}
\end{tikzpicture}
\end{minipage}
\caption{{A comparison of the obtained residual and errors vs. noise in the Tomography task on OrganCMNIST test set.}
}
\label{fig:residual_and_error_tomography_CTorgan}

\end{figure*}
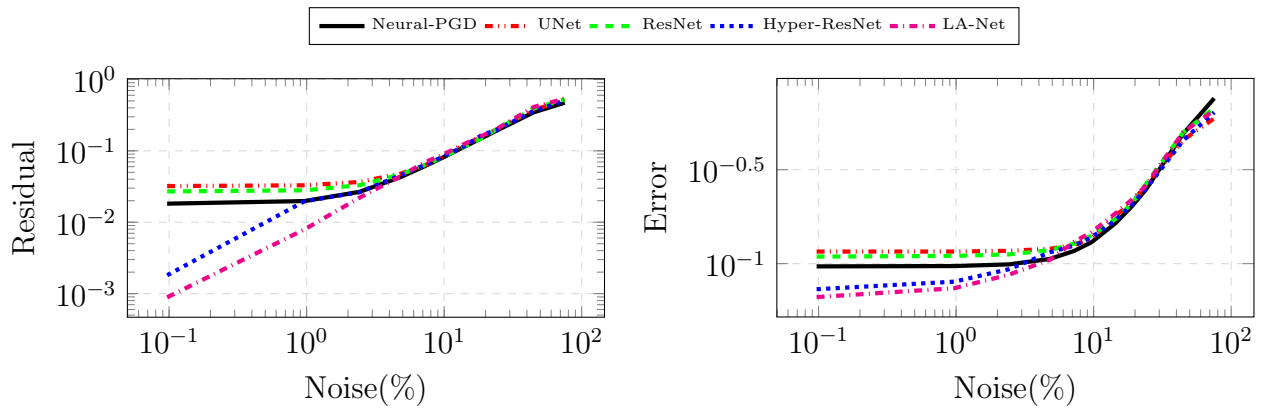

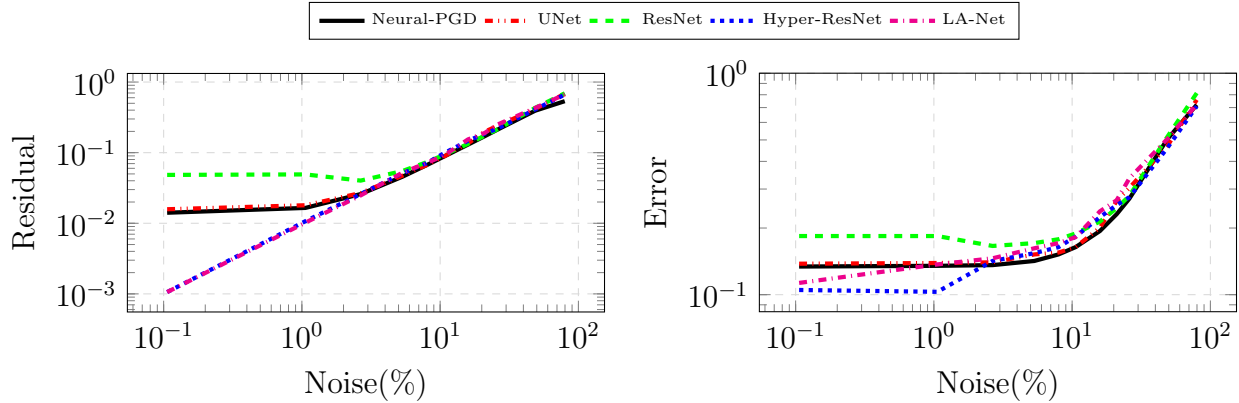
\begin{figure*}[h]
\centering
\begin{minipage}{0.48\textwidth}
\centering
\begin{tikzpicture}[define rgb/.code={\definecolor{mycolor}{RGB}{#1}},
                    rgb color/.style={define rgb={#1},mycolor}]
  \begin{axis}[
      width=1.0\linewidth, 
      height=0.6\linewidth,
      grid=major,
      grid style={dashed,gray!30},
      xlabel=Noise(\%),
      ylabel=Residual,
      ylabel near ticks,
      legend style={at={(1.125,1.3)},anchor=north,scale=1.0, cells={anchor=west}, font=\tiny,},
      legend columns=-1,
      yticklabel style={
        /pgf/number format/fixed,
        /pgf/number format/precision=3
      },
      scaled y ticks=false,
      every axis plot post/.style={ultra thick},
      xmode=log,
      ymode=log
    ]
    \addplot[color=black]
    table[x=noise,y=pgdresidual,col sep=comma] {data/ip_deblur.csv};
    
    \addplot[color=red, style=dashdotdotted]
    table[x=noise,y=unetresidual,col sep=comma] {data/ip_deblur.csv};
    \addplot[color=green, style=dashed]
    table[x=noise,y=resnetresidual
    ,col sep=comma] {data/ip_deblur.csv};

    \addplot[color=blue, style=dotted]
    table[x=noise,y=hyperresidual
    ,col sep=comma] {data/ip_deblur.csv}; 
    
    \addplot[color=magenta, style=dashdotted]
    table[x=noise,y=laresidual
    ,col sep=comma] {data/ip_deblur.csv}; 
    \legend{{Neural-PGD}, UNet, ResNet, Hyper-ResNet, LA-Net}
    \end{axis}
\end{tikzpicture}
\end{minipage} \hspace{0.5em}
\begin{minipage}{.48\textwidth}
\centering
\begin{tikzpicture}[define rgb/.code={\definecolor{mycolor}{RGB}{#1}},
                    rgb color/.style={define rgb={#1},mycolor}]
  \begin{axis}[
      width=1.0\linewidth, 
      height=0.6\linewidth,
      grid=major,
      grid style={dashed,gray!30},
      xlabel=Noise(\%),
      ylabel=Error,
      ylabel near ticks,
      legend style={at={(10,1.3)},anchor=north,scale=1.0, draw=none, cells={anchor=west}, font=\tiny, fill=none},
      legend columns=-1,
      yticklabel style={
        /pgf/number format/fixed,
        /pgf/number format/precision=3
      },
      scaled y ticks=false,
      every axis plot post/.style={ultra thick},
      xmode=log,
      ymode=log
    ]

    \addplot[color=black]
    table[x=noise,y=pgderror,col sep=comma] {data/ip_deblur.csv};
    
    \addplot[color=red, style=dashdotdotted]
    table[x=noise,y=uneterror,col sep=comma] {data/ip_deblur.csv};
    \addplot[color=green, style=dashed]
    table[x=noise,y=resneterror
    ,col sep=comma] {data/ip_deblur.csv};

    \addplot[color=blue, style=dotted]
    table[x=noise,y=hypererror
    ,col sep=comma] {data/ip_deblur.csv}; 
    
    \addplot[color=magenta, style=dashdotted]
    table[x=noise,y=laerror
    ,col sep=comma] {data/ip_deblur.csv};

    \addlegendimage{}
    \addlegendimage{}
    \addlegendimage{}
    \addlegendentry{}
    \end{axis}
\end{tikzpicture}
\end{minipage}
\caption{A comparison of the obtained residual and errors vs. noise in the Deblurring task on {STL-10 test set}. 
}
\label{fig:residual_and_error_deblur}

\end{figure*}

We now turn to show the advantages of our DRIP approach compared to the neural proximal methods. First, we show that although both DRIP-based LA-Net and Hyper-ResNet, as well as standard networks like {UNet, ResNet, and Neural-PGD} offer qualitatively similar results if evaluated under the same conditions and settings as used during the training phase, our LA-Net and Hyper-ResNet networks still offer slightly better reconstructions. That holds both qualitatively, as can be seen in Figure \ref{fig:tomography_results} and quantitatively as shown in Figures \ref{fig:residual_and_error_tomography}, \ref{fig:residual_and_error_tomography_CTorgan}, and \ref{fig:residual_and_error_deblur}.

%%%%%%%%%%

%%%%%%%%%%

\begin{figure*}[h]
\centering
\begin{minipage}{0.48\textwidth}
\centering
\begin{tikzpicture}[define rgb/.code={\definecolor{mycolor}{RGB}{#1}},
                    rgb color/.style={define rgb={#1},mycolor}]
  \begin{axis}[
      width=1.0\linewidth, 
      height=0.6\linewidth,
      grid=major,
      grid style={dashed,gray!30},
      xlabel=Iterations,
      ylabel=Residual,
      ylabel near ticks,
      legend style={at={(1.125,1.3)},anchor=north,scale=1.0, cells={anchor=west}, font=\tiny,},
      legend columns=-1,
      yticklabel style={
        /pgf/number format/fixed,
        /pgf/number format/precision=3
      },
      scaled y ticks=false,
      every axis plot post/.style={ultra thick},
      xmode=log,
      ymode=log
    ]

    \addplot[color=black]
    table[x=iters,y=pgdresidual
    ,col sep=comma] {data/ip_tomography_ood.csv};
    \addplot[color=red, style=dashdotdotted]
table[x=iters,y=unetresidual,col sep=comma] {data/ip_tomography_ood.csv};
    \addplot[color=green, style=dashed]
    table[x=iters,y=resnetresidual
    ,col sep=comma] {data/ip_tomography_ood.csv};

    \addplot[color=blue, style=dotted]
    table[x=iters,y=hyperresidual
    ,col sep=comma] {data/ip_tomography_ood.csv}; 
    
    \addplot[color=magenta, style=dashdotted]
    table[x=iters,y=laresidual
    ,col sep=comma] {data/ip_tomography_ood.csv}; 
    \legend{{Neural-PGD}, UNet, ResNet, Hyper-ResNet, LA-Net}
    \end{axis}
\end{tikzpicture}
\end{minipage} \hspace{0.5em}
\begin{minipage}{.48\textwidth}
\centering
\begin{tikzpicture}[define rgb/.code={\definecolor{mycolor}{RGB}{#1}},
                    rgb color/.style={define rgb={#1},mycolor}]
  \begin{axis}[
      width=1.0\linewidth, 
      height=0.6\linewidth,
      grid=major,
      grid style={dashed,gray!30},
      xlabel=Iterations,
      ylabel=Error,
      ylabel near ticks,
      legend style={at={(10,1.3)},anchor=north,scale=1.0, draw=none, cells={anchor=west}, font=\tiny, fill=none},
      legend columns=-1,
      yticklabel style={
        /pgf/number format/fixed,
        /pgf/number format/precision=3
      },
      scaled y ticks=false,
      every axis plot post/.style={ultra thick},
      xmode=log,
      ymode=log
    ]

    \addplot[color=black]
    table[x=iters,y=pgderror
    ,col sep=comma] {data/ip_tomography_ood.csv};
    
    \addplot[color=red, style=dashdotdotted]
    table[x=iters,y=uneterror,col sep=comma] {data/ip_tomography_ood.csv};
    \addplot[color=green, style=dashed]
    table[x=iters,y=resneterror
    ,col sep=comma] {data/ip_tomography_ood.csv};

    \addplot[color=blue, style=dotted]
    table[x=iters,y=hypererror
    ,col sep=comma] {data/ip_tomography_ood.csv}; 
    
    \addplot[color=magenta, style=dashdotted]
    table[x=iters,y=laerror
    ,col sep=comma] {data/ip_tomography_ood.csv}; 
    \addlegendimage{}
    \addlegendimage{}
    \addlegendimage{}
    \addlegendentry{}
    \end{axis}
\end{tikzpicture}
\end{minipage}
\caption{A comparison of the obtained residual and errors vs. iterations in the Tomography task {on the STL-10 test set}. 
}
\label{fig:residual_and_error_tomography_ood}

\end{figure*}
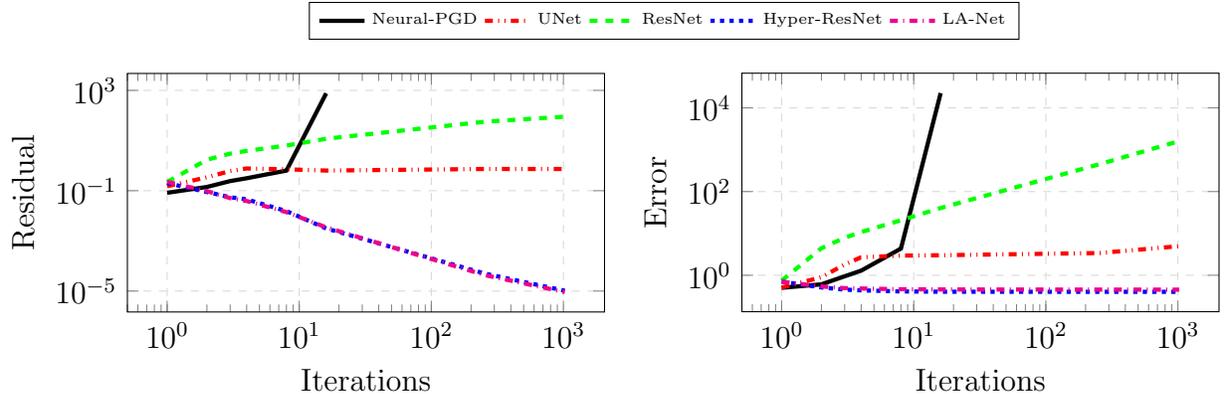

Moreover, from Figures \ref{fig:residual_and_error_tomography}, \ref{fig:residual_and_error_tomography_CTorgan}, and \ref{fig:residual_and_error_deblur}, we learn that our DRIP does not degrade (relative to the expected noise levels) as the noise regime varies, while the baseline models offer worse performance. Specifically, we find that as the noise levels decrease, both the data fitting (residual) and error terms offered by our DRIP-based LA-Net and Hyper-ResNet networks decrease, as expected, thanks to our data fitting mechanism. That is in contrast to the observations previously discussed regarding proximal methods, which do not fit the data to the desired accuracy in such cases. To further demonstrate that, we provide visual comparisons on an out-of-distribution phantom object to be reconstructed from tomographic projections, to learn about the qualitative difference between our DRIP approach and a proximal-based neural network, in Figure \ref{fig:ood_iterations}. Finally, it can be seen from Figure \ref{fig:residual_and_error_tomography_ood}  that as more iterations (applications of the network in the case of baseline methods like UNet and ResNet, or, the value of ''maxIter'' in Algorithms \ref{alg:ipsolve} and \ref{alg:ipsolveDirect} for LA-Net and Hyper-ResNet) are added, our DRIP approach yields lower residual and error terms (that is, improved performance), while the neural proximal methods worsen with more iterations.
Specifically, Figure \ref{fig:residual_and_error_tomography_ood} demonstrates that using both standard and DRIP-based types of networks on in-distribution noise levels yields similar results among the different networks, in accordance with the quantitative results discussed earlier. However, upon reducing the noise levels, that is, feeding the network with out-of-distribution noise levels, it is observed that our DRIP approach offers favorable recovered images, and by applying more iterations of the DRIP networks it is possible to improve the quality of the reconstruction, unlike the case in baseline networks like UNet or ResNet, where more applications of the networks cause reconstruction degradation.

\begin{table}
    \centering
    \renewcommand{\arraystretch}{2}
    \begin{tabular}{|c|ccccc|}
    \hline 
    \multirow{2}{*}
    { Ground-truth \includegraphics[width=3cm, valign=c, trim={1cm 1cm 1cm 1cm},clip]{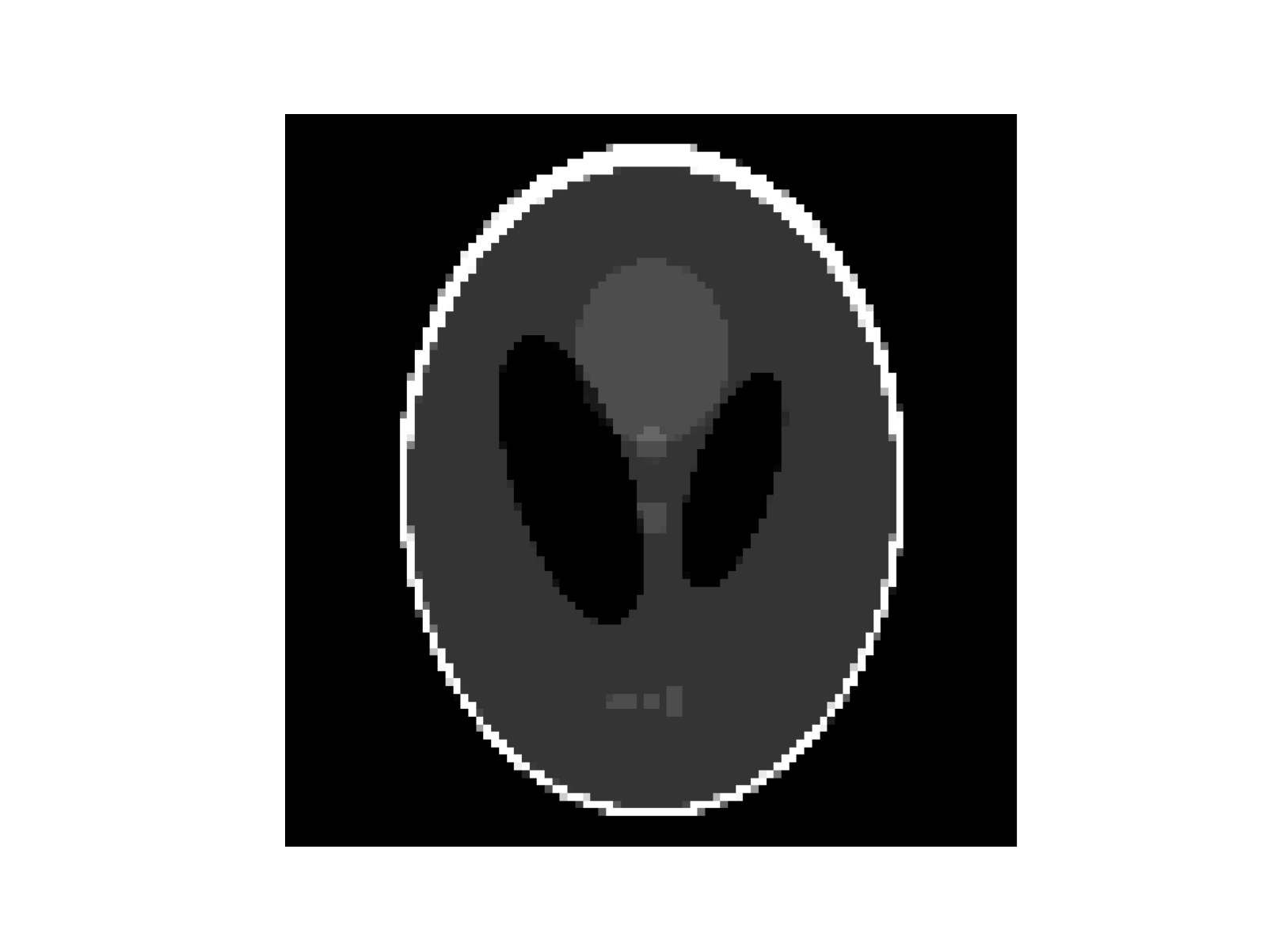}} & \tiny{LA-Net} & 
    \includegraphics[width=1.6cm, valign=c, trim={1cm 1cm 1cm 1cm},clip]{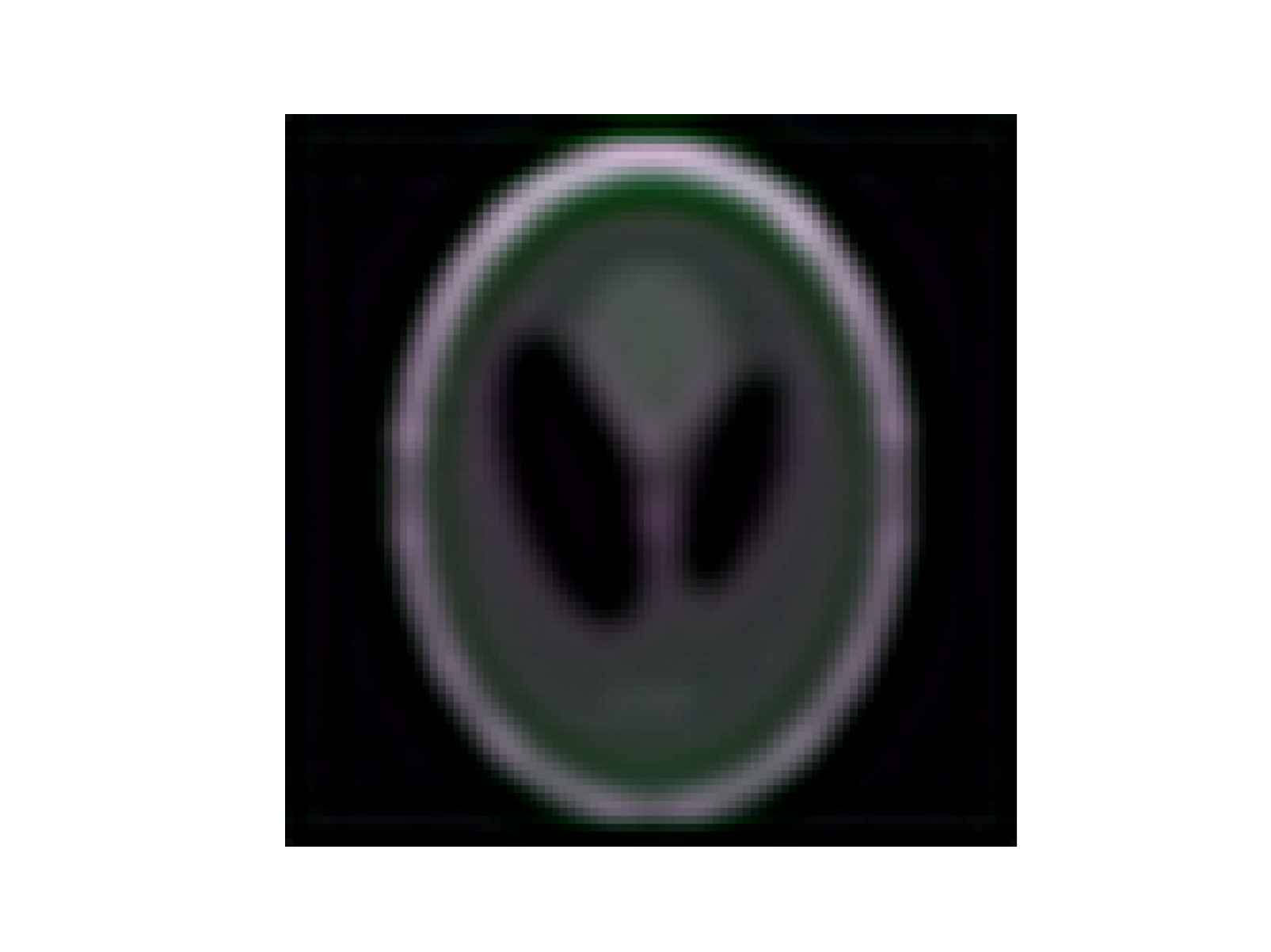}
     &
    \includegraphics[width=1.6cm, valign=c, trim={1cm 1cm 1cm 1cm},clip]{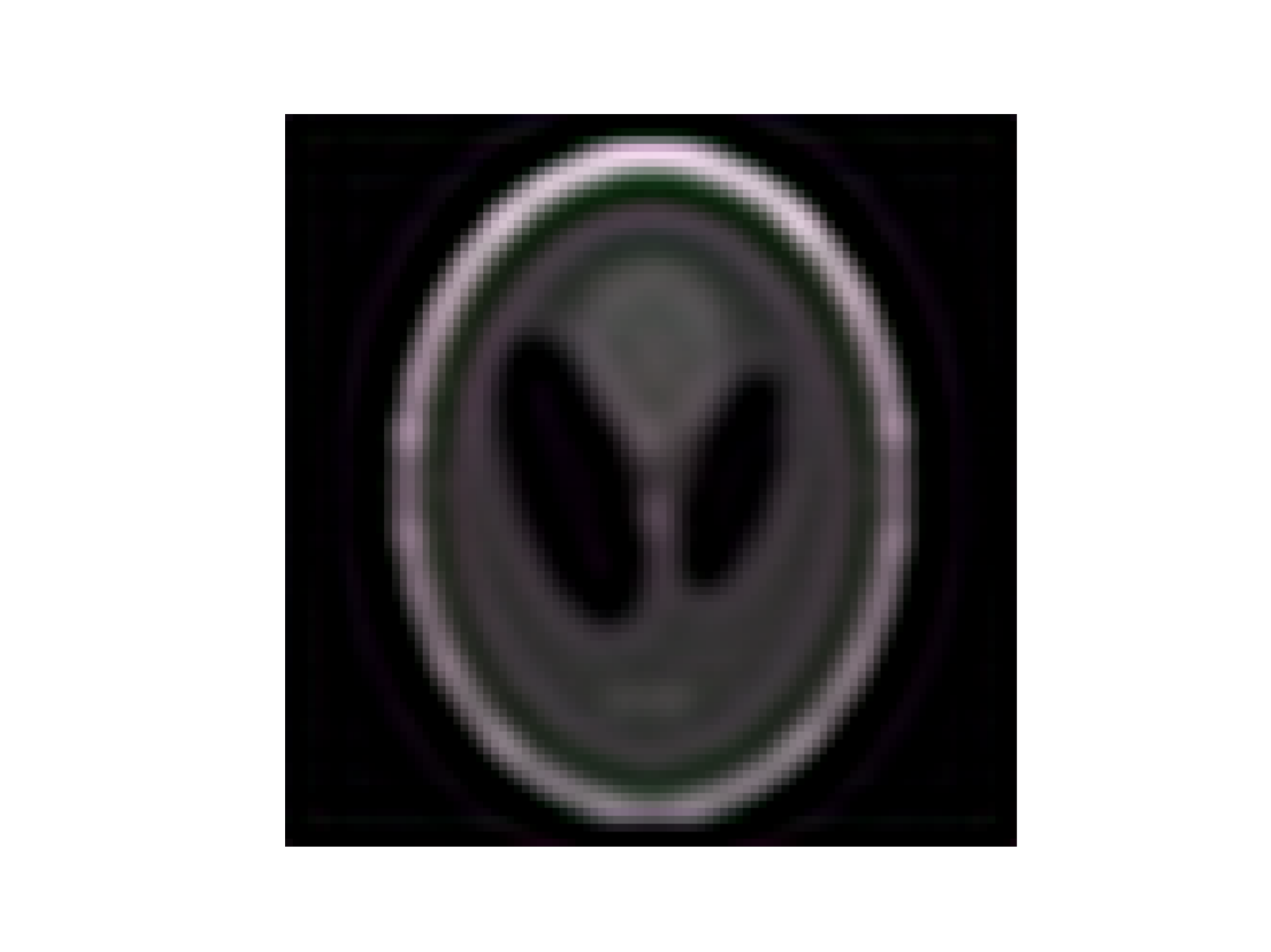} &
    \includegraphics[width=1.6cm, valign=c, trim={1cm 1cm 1cm 1cm},clip]{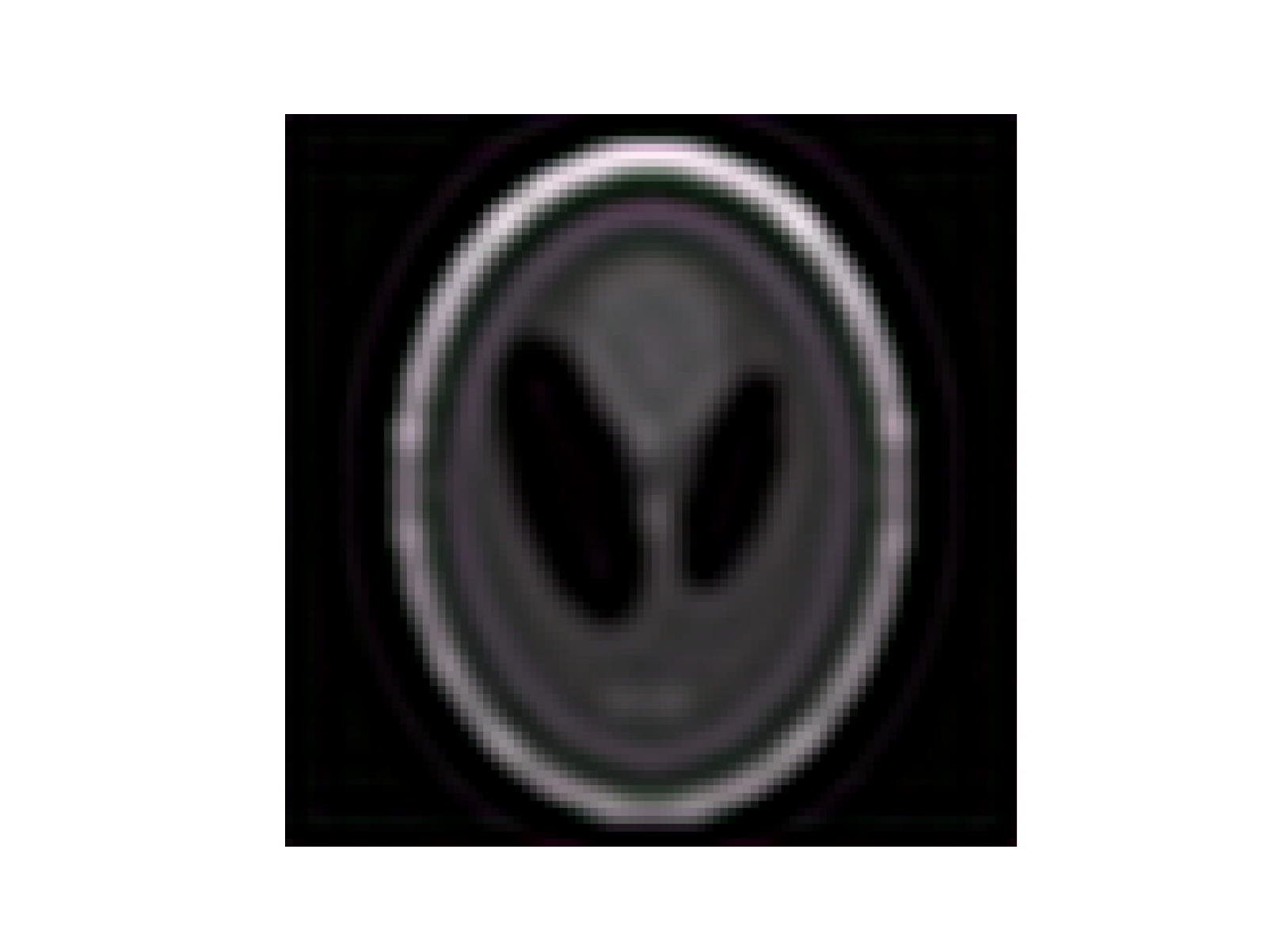} &
    \includegraphics[width=1.6cm, valign=c, trim={1cm 1cm 1cm 1cm},clip]{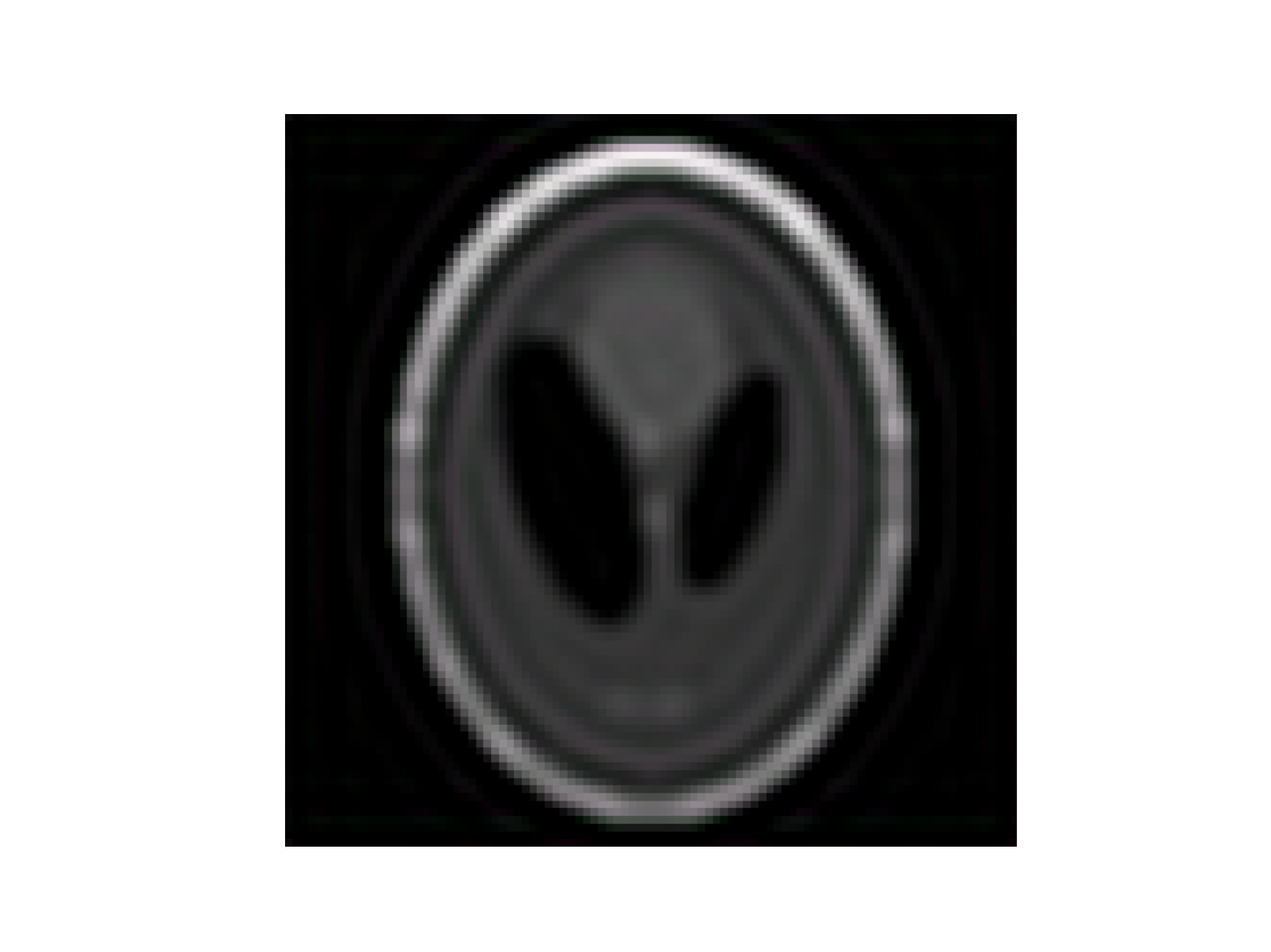}
    \\
        \cline{2-6}
     
     & \tiny{Hyper-ResNet} &  \includegraphics[width=1.6cm, valign=c, trim={1cm 1cm 1cm 1cm},clip]{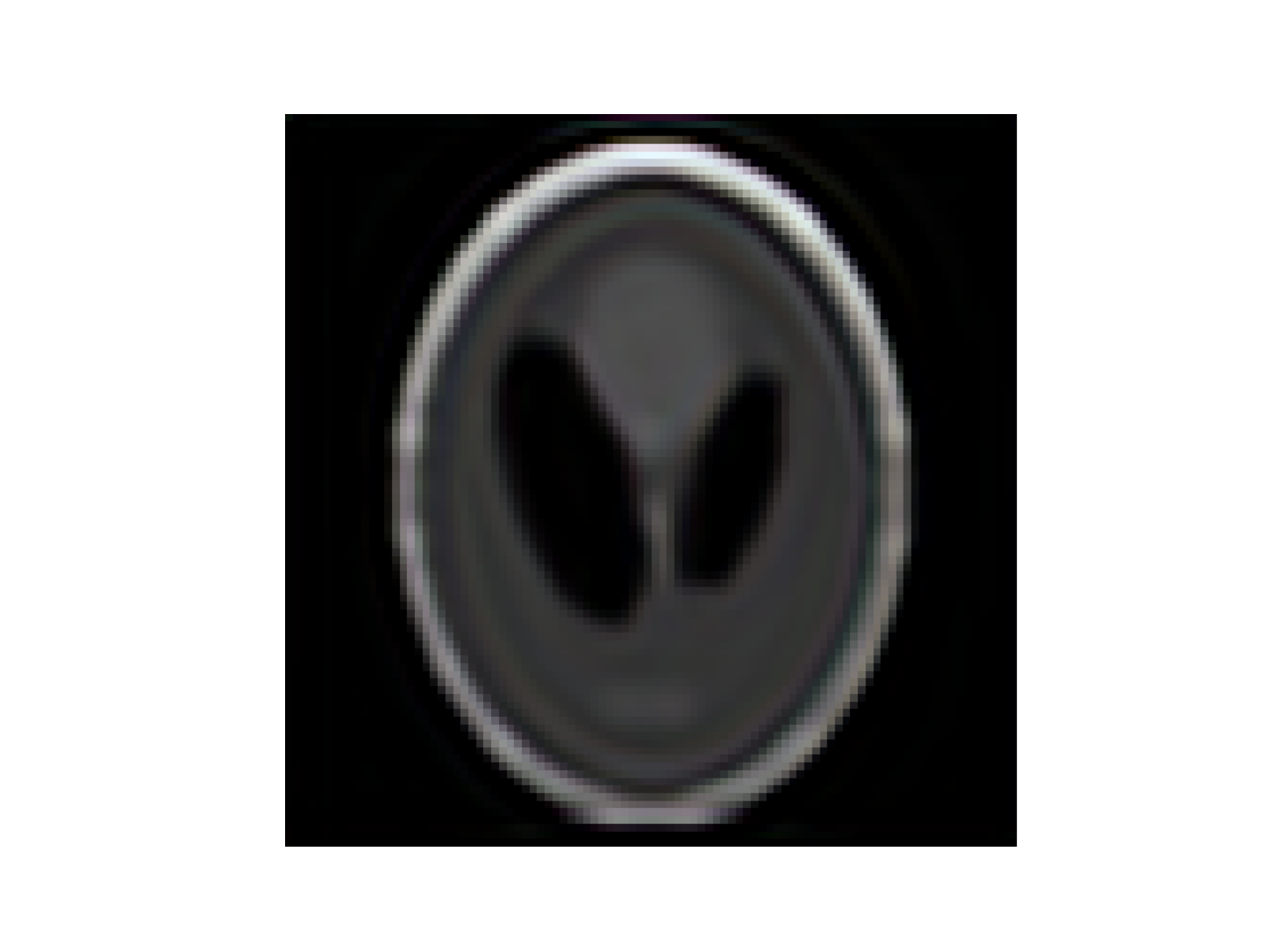}  &\includegraphics[width=1.6cm, valign=c, trim={1cm 1cm 1cm 1cm},clip]{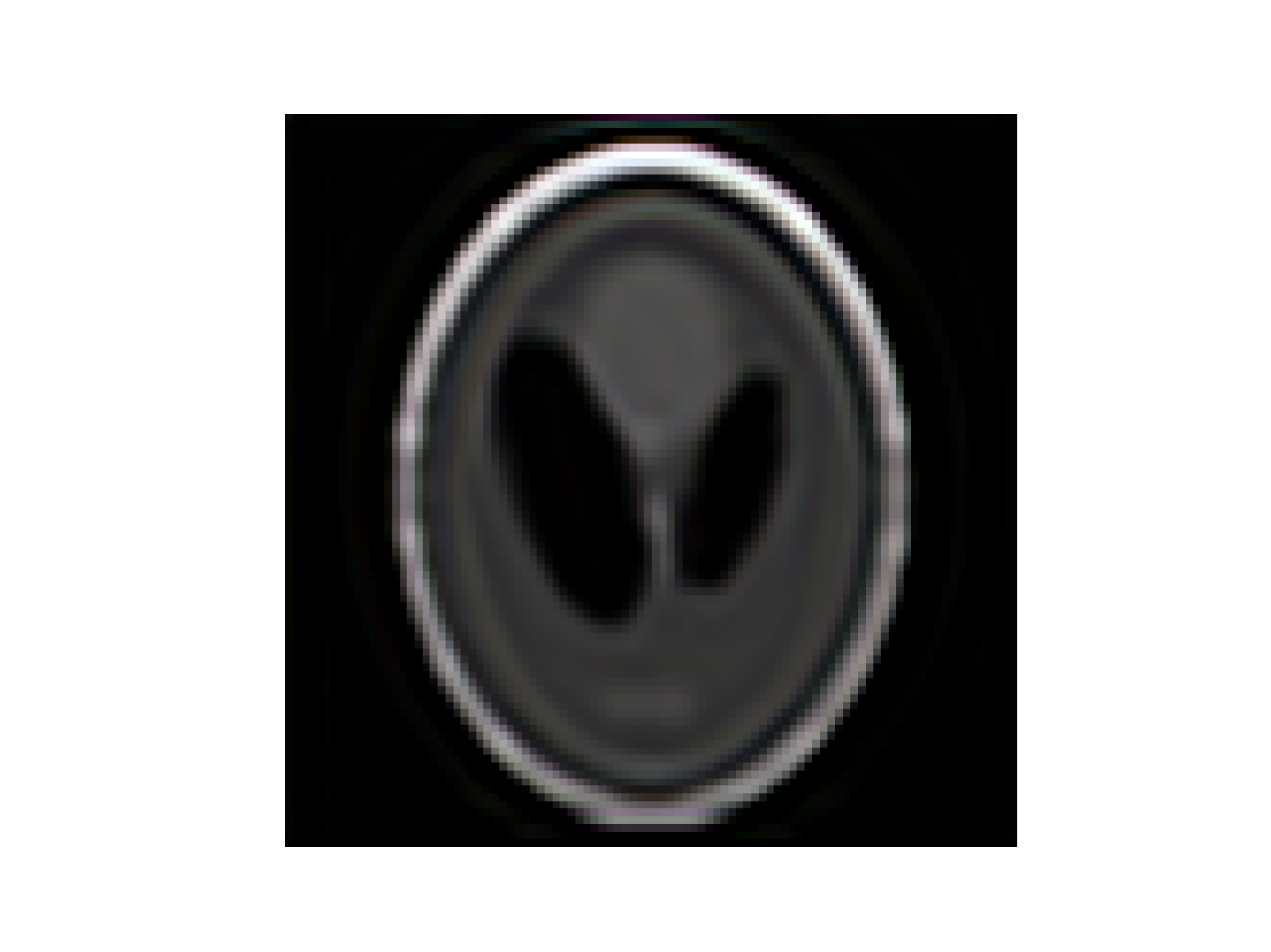} & \includegraphics[width=1.6cm, valign=c, trim={1cm 1cm 1cm 1cm},clip]{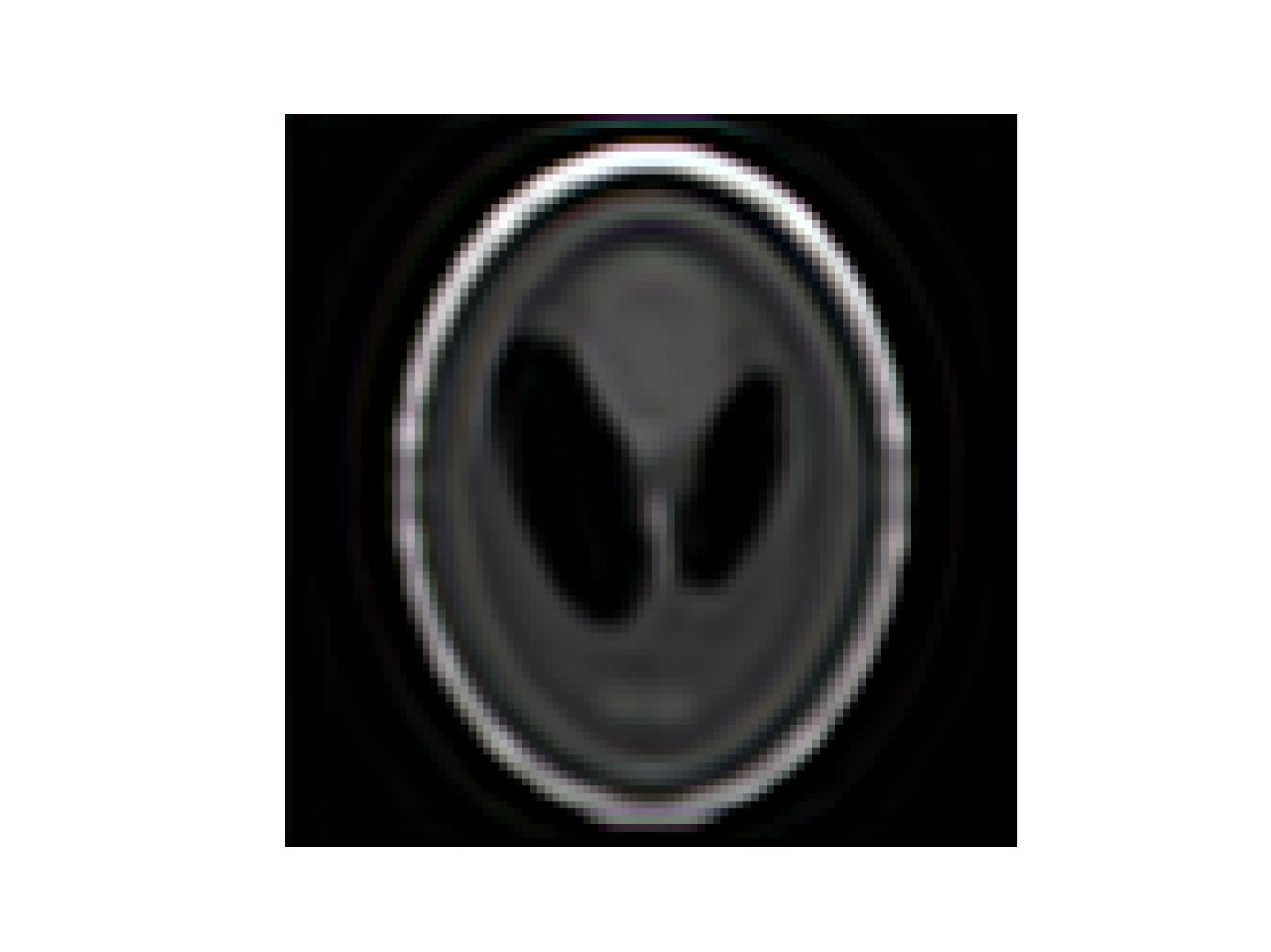} & \includegraphics[width=1.6cm, valign=c, trim={1cm 1cm 1cm 1cm},clip]{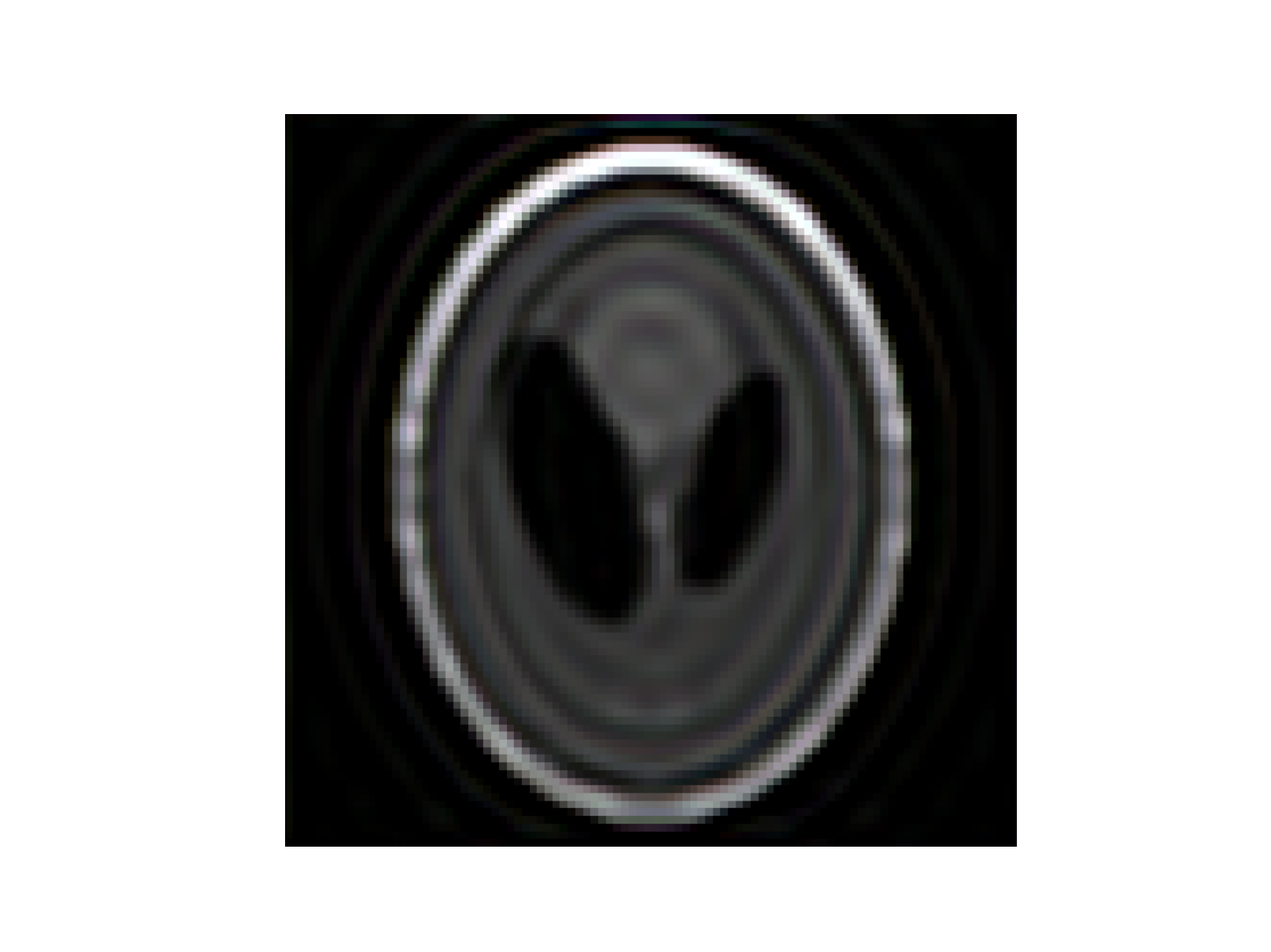}     \\
        \cline{2-6}
    
    \multirow{3}{*}{ Observed data \includegraphics[width=3cm, valign=c, trim={1cm 1cm 1cm 1cm},clip]{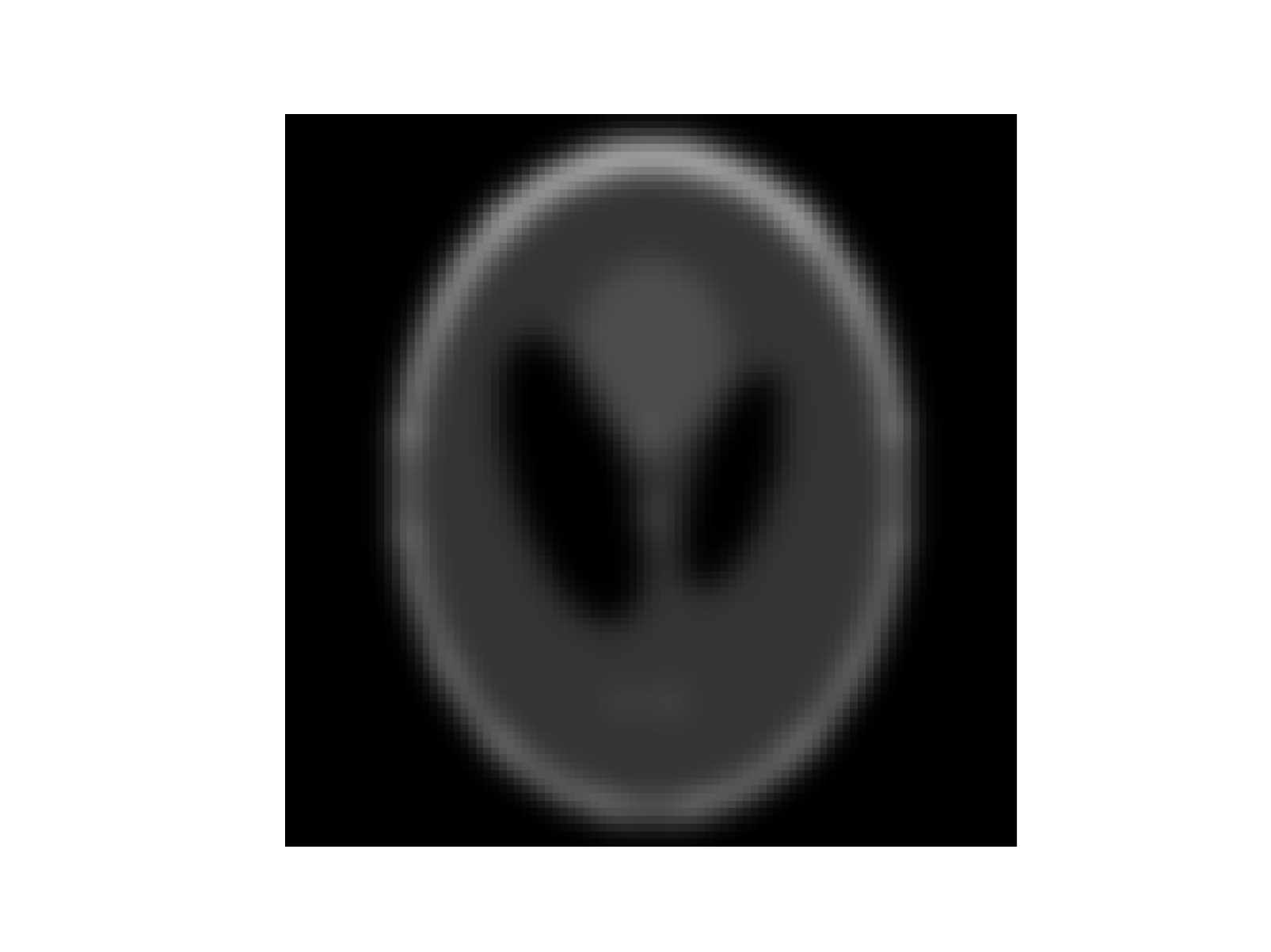}} & \tiny{{Neural-PGD}} & \includegraphics[width=1.6cm, valign=c, trim={1cm 1cm 1cm 1cm},clip]{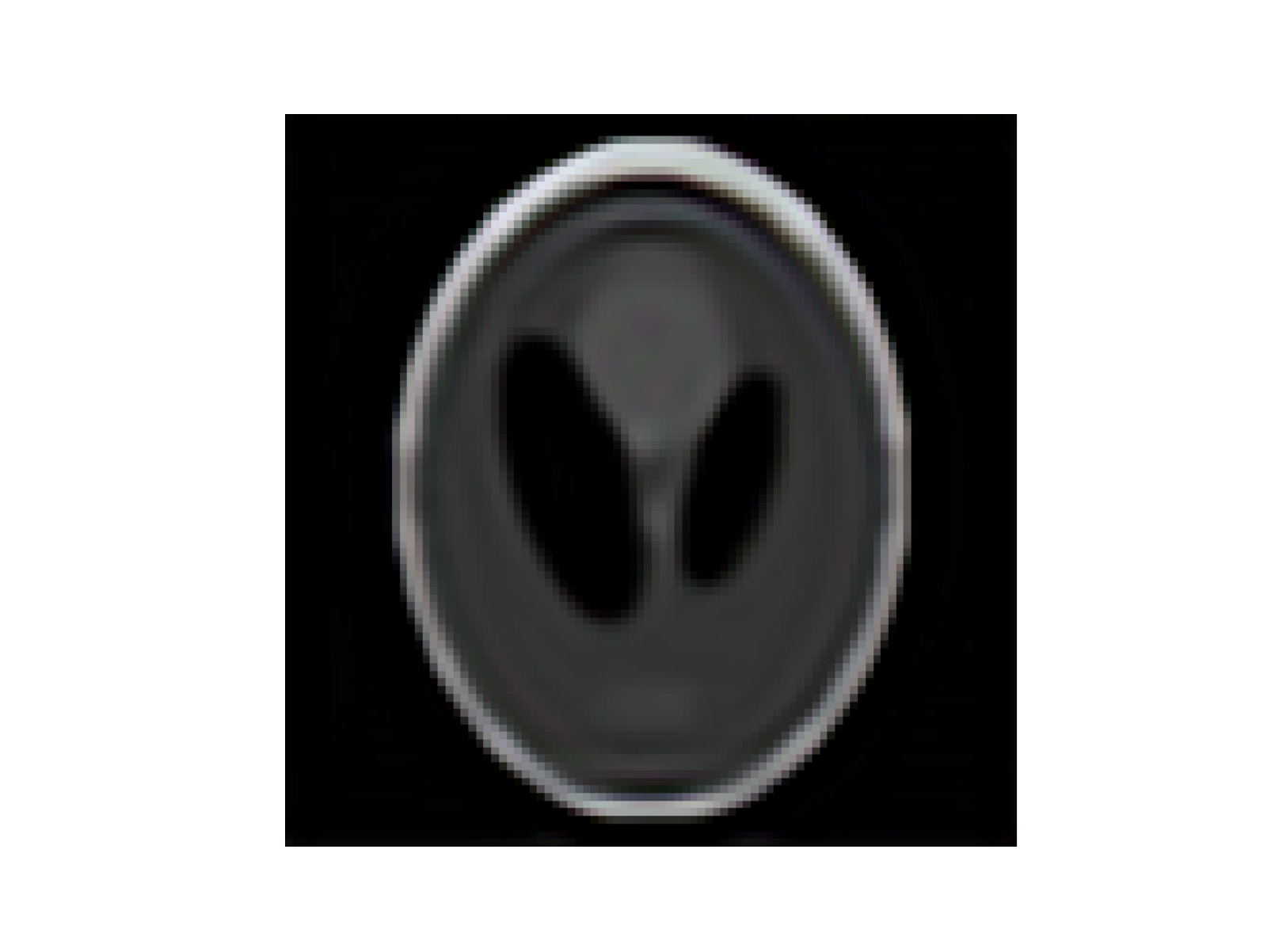}& \includegraphics[width=1.6cm, valign=c, trim={1cm 1cm 1cm 1cm},clip]{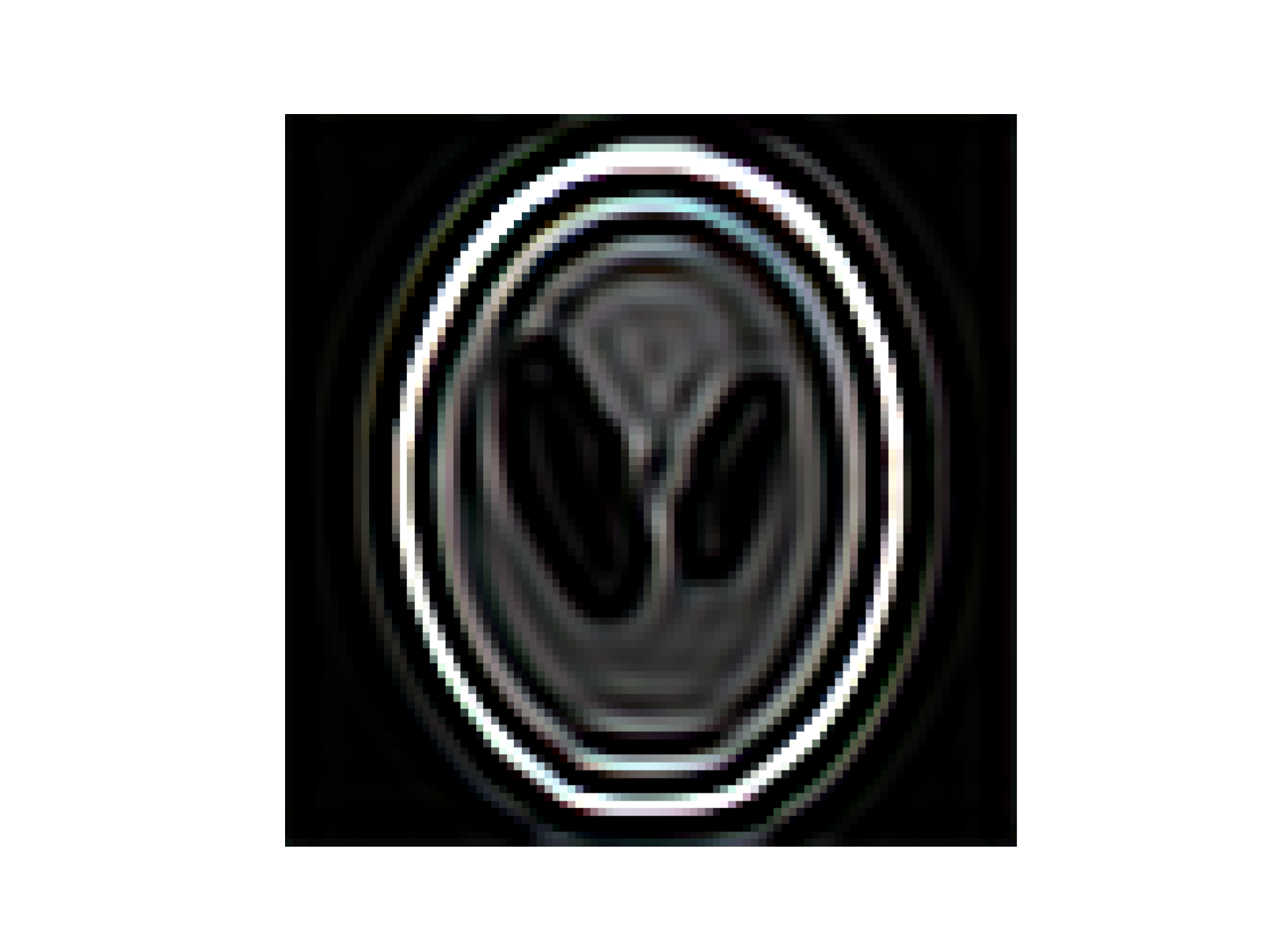}& \includegraphics[width=1.6cm, valign=c, trim={1cm 1cm 1cm 1cm},clip]{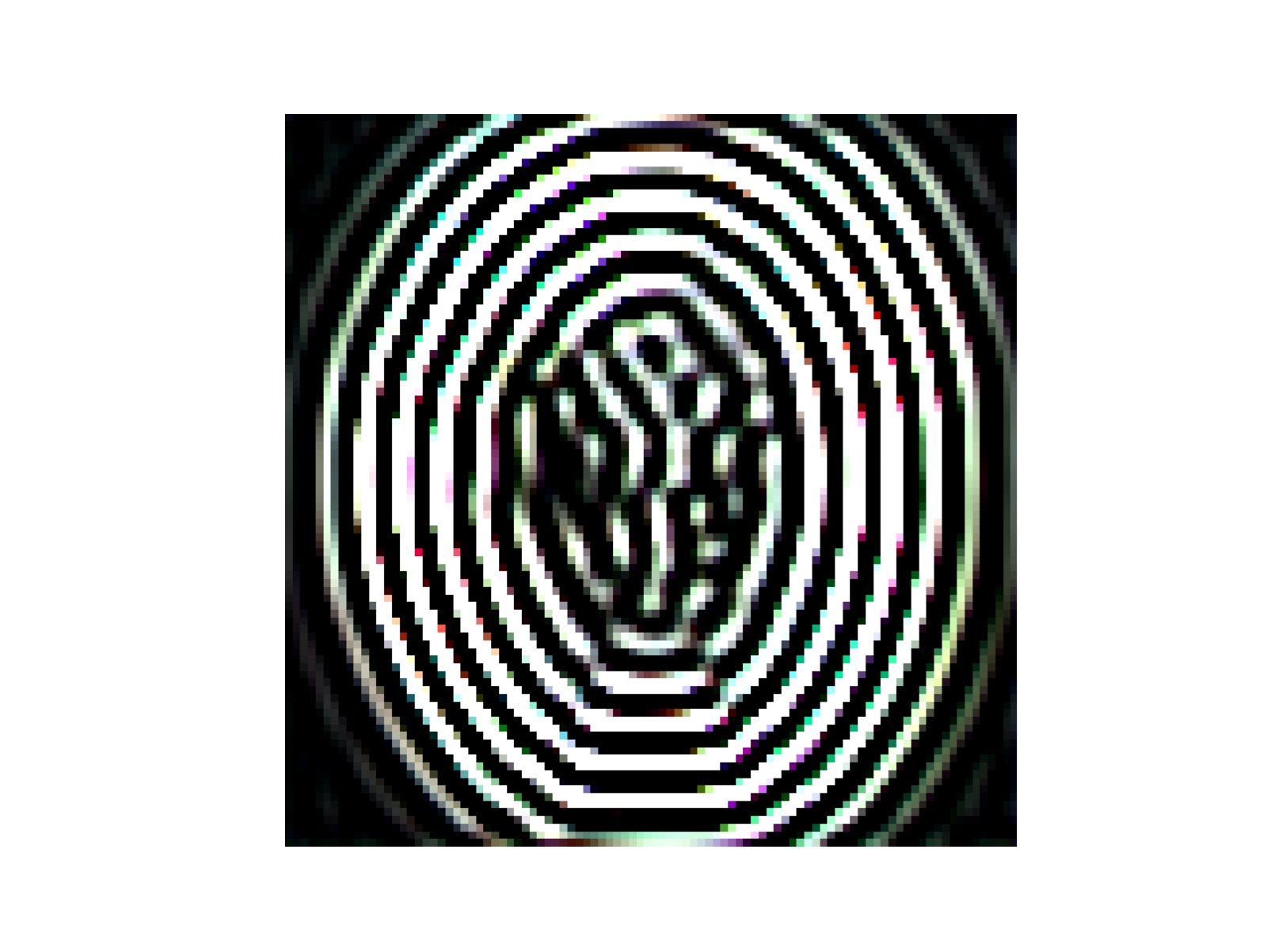}& \includegraphics[width=1.6cm, valign=c, trim={1cm 1cm 1cm 1cm},clip]{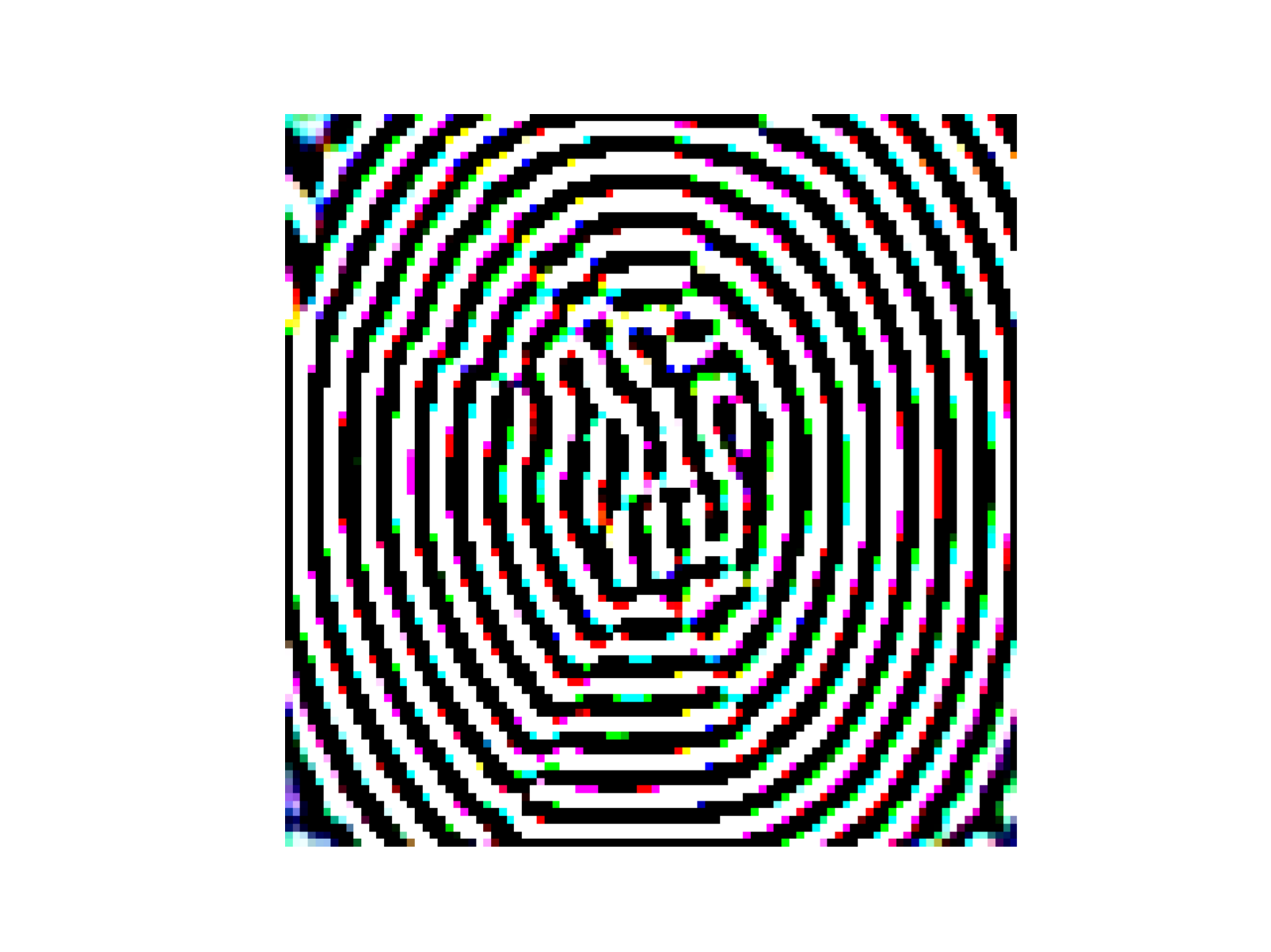}     \\
        \cline{2-6}

         & \tiny{UNet} & \includegraphics[width=1.6cm, valign=c, trim={1cm 1cm 1cm 1cm},clip]{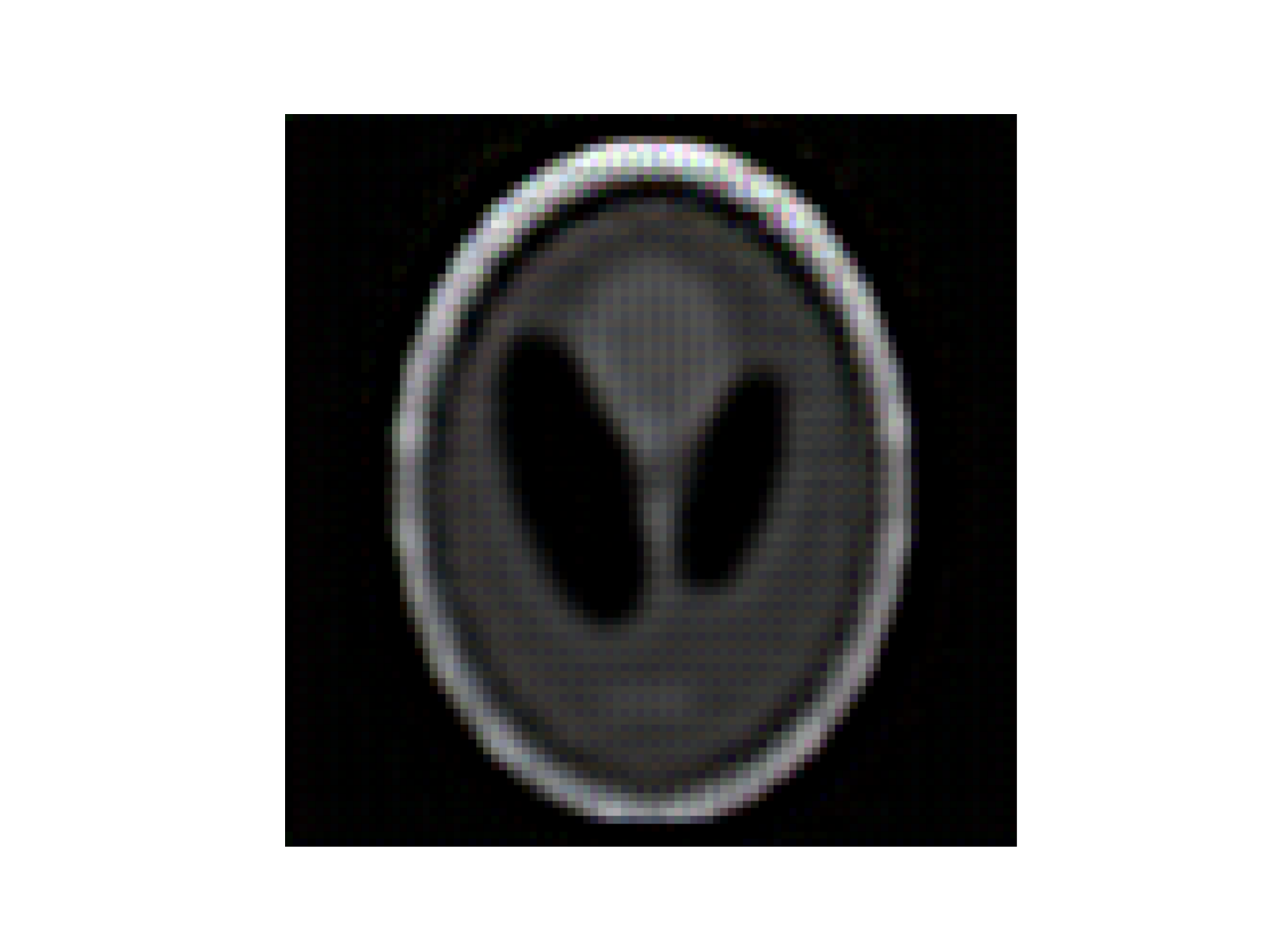}& \includegraphics[width=1.6cm, valign=c, trim={1cm 1cm 1cm 1cm},clip]{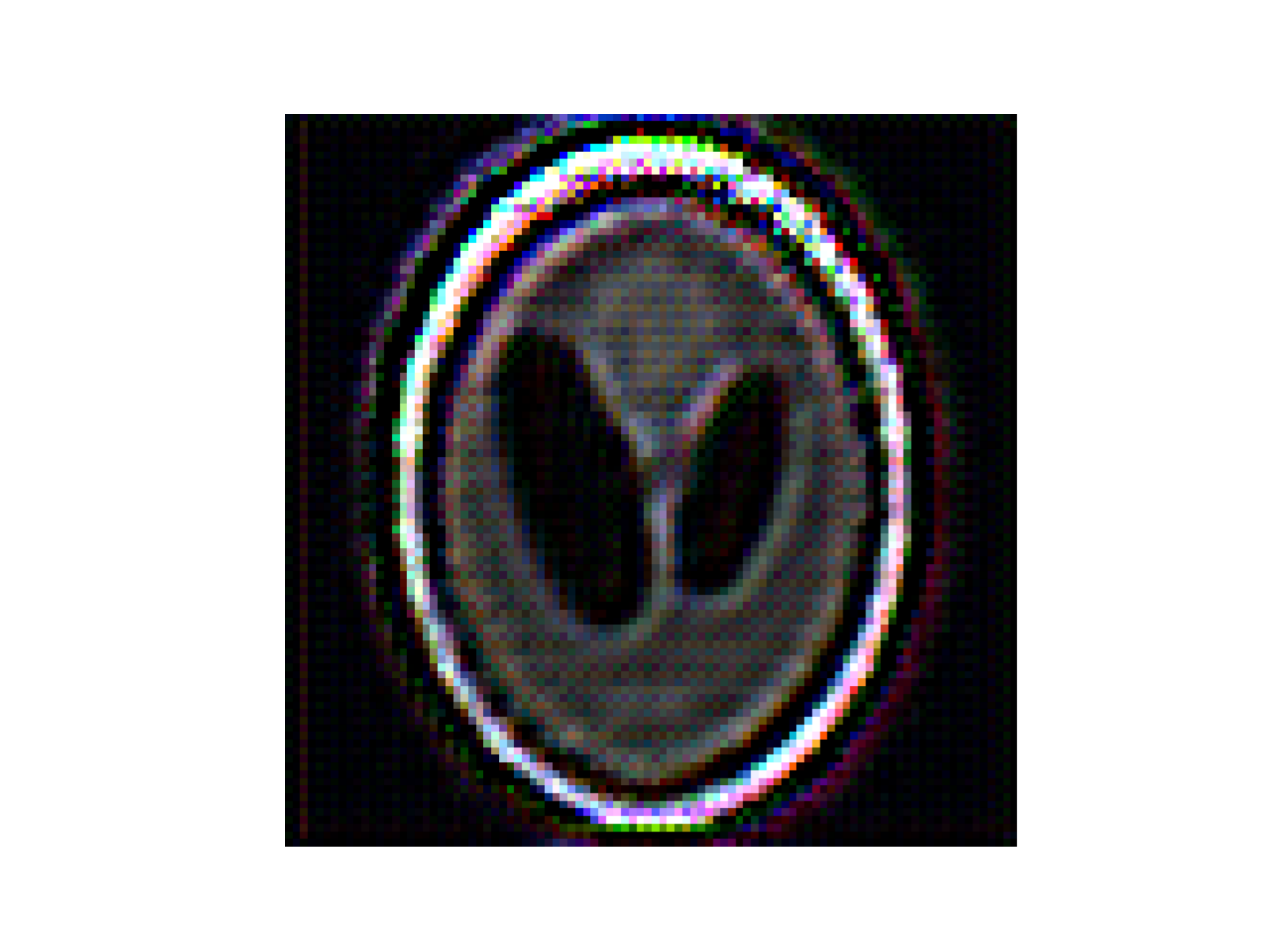}& \includegraphics[width=1.6cm, valign=c, trim={1cm 1cm 1cm 1cm},clip]{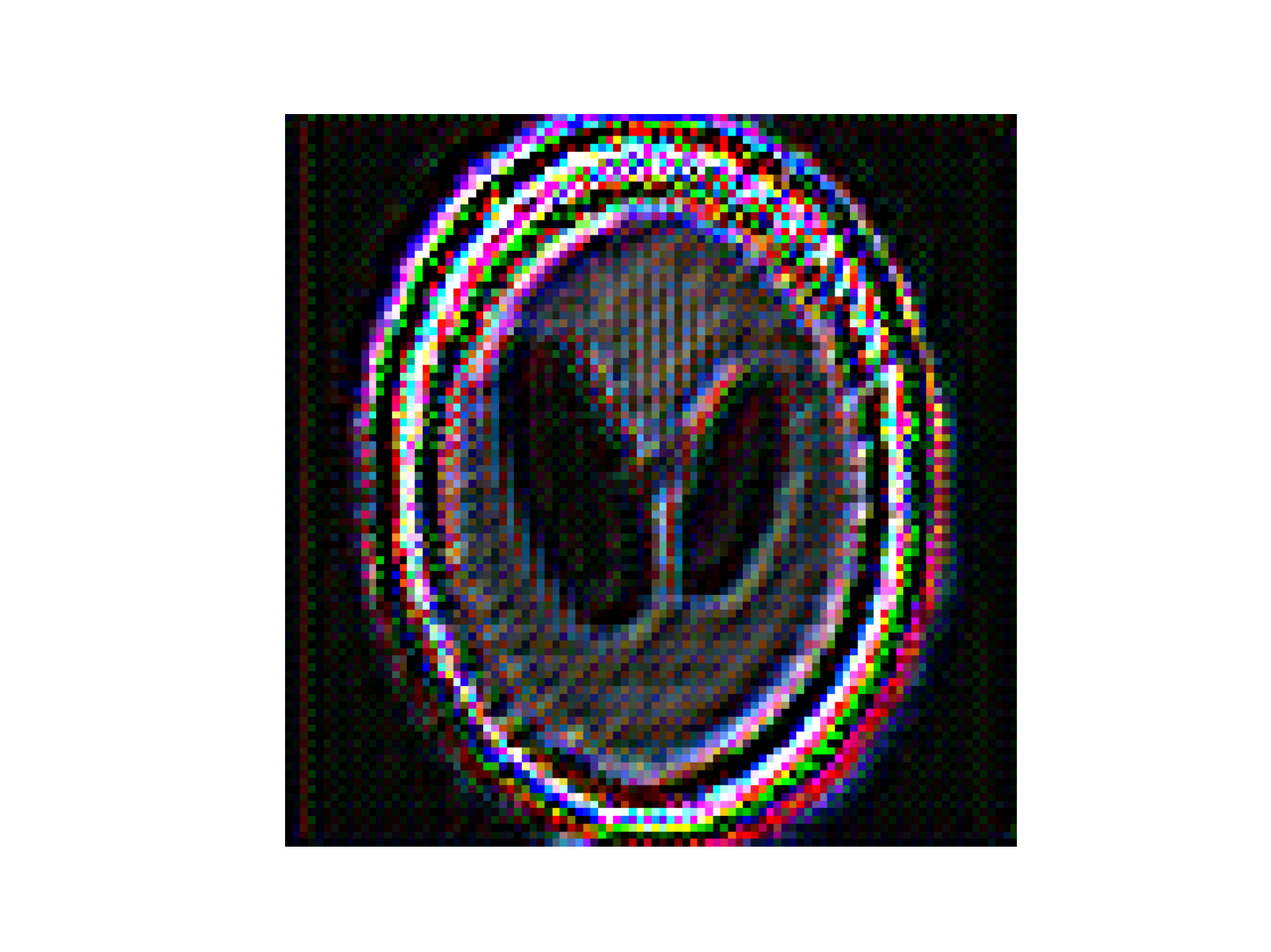}& \includegraphics[width=1.6cm, valign=c, trim={1cm 1cm 1cm 1cm},clip]{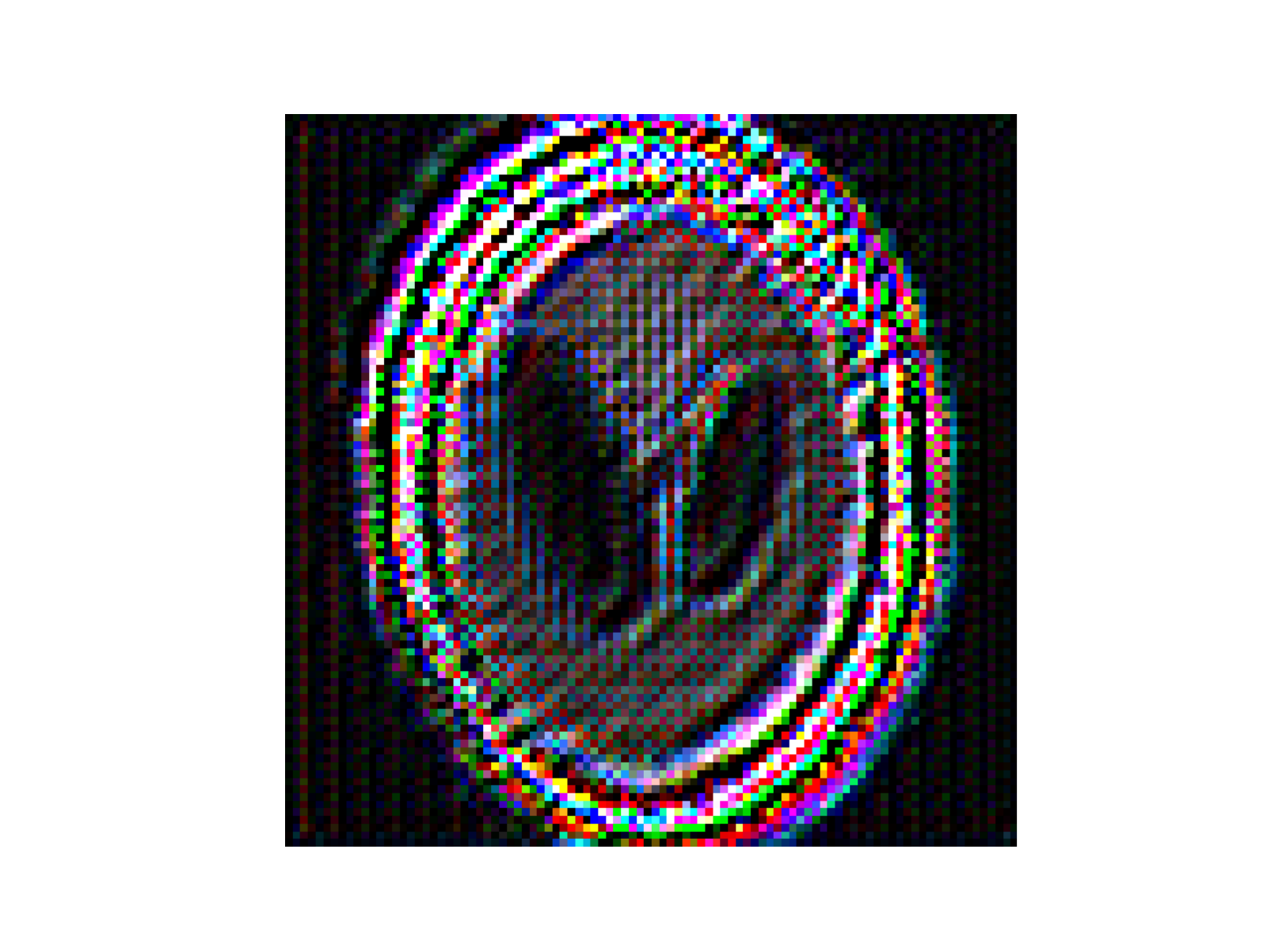}     \\
        \cline{2-6}
    
     & \tiny{ResNet}
     & \includegraphics[width=1.6cm, valign=c, trim={1cm 1cm 1cm 1cm},clip]{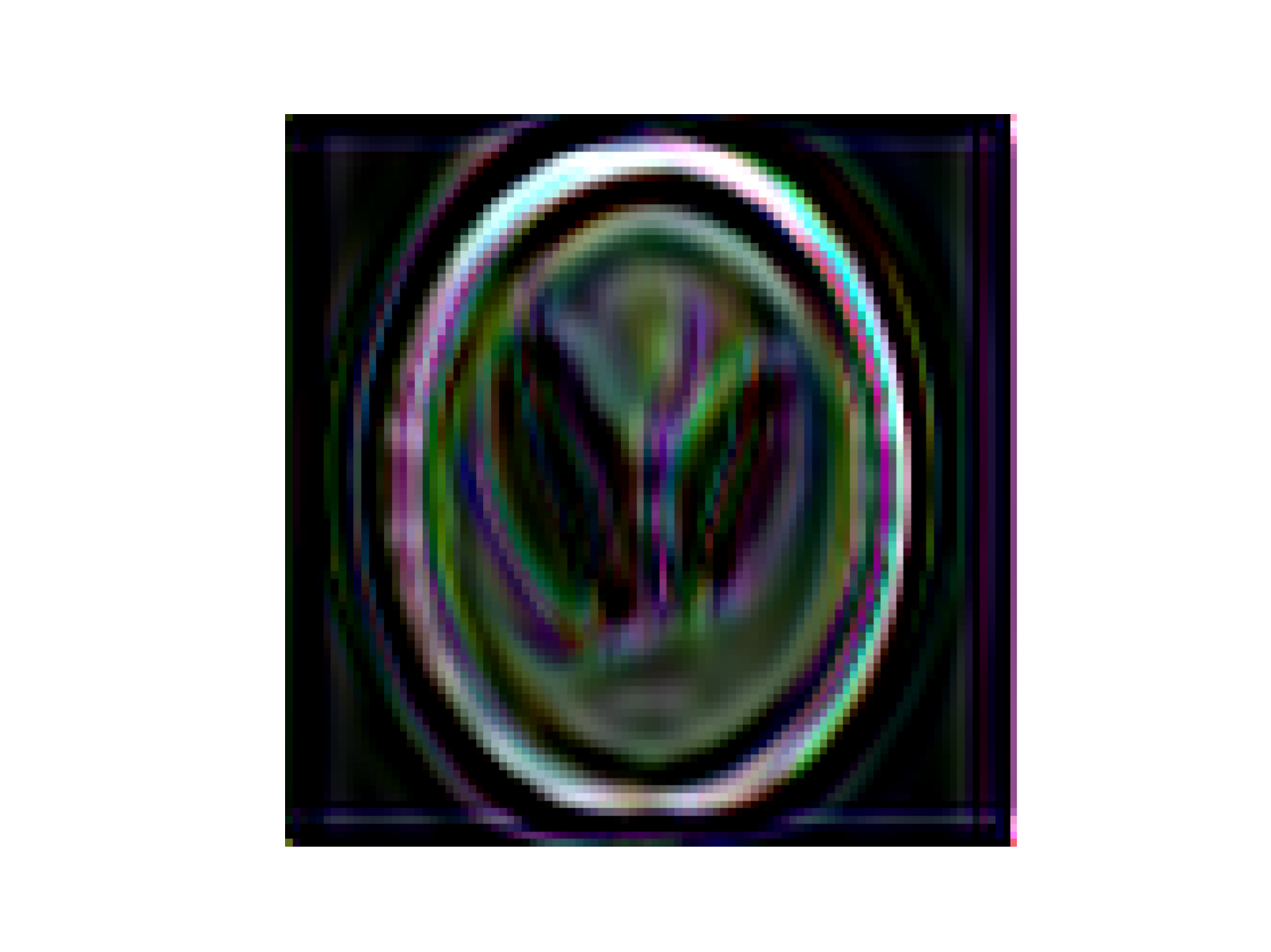}
     & \includegraphics[width=1.6cm, valign=c, trim={1cm 1cm 1cm 1cm},clip]{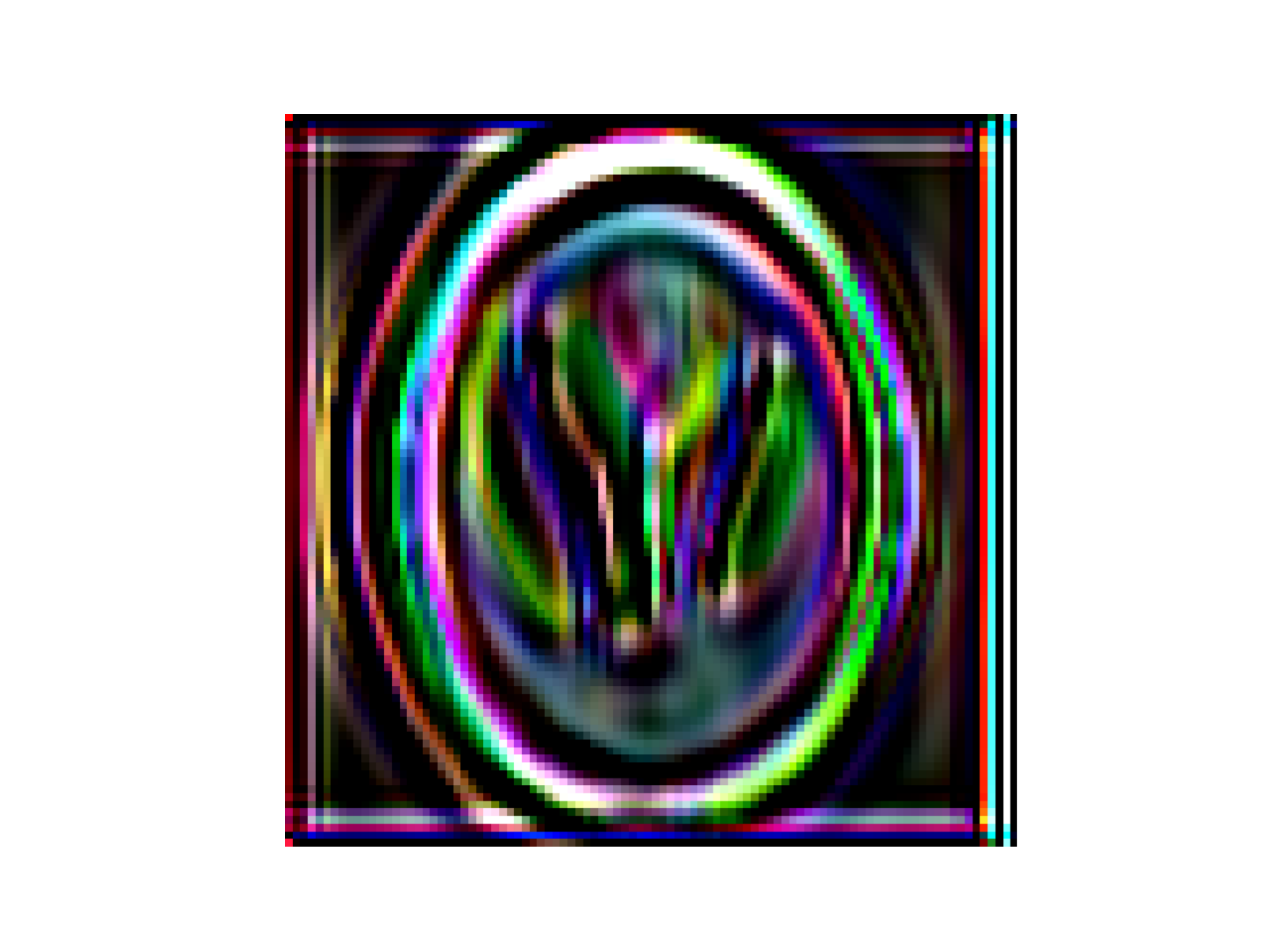}
     & \includegraphics[width=1.6cm, valign=c, trim={1cm 1cm 1cm 1cm},clip]{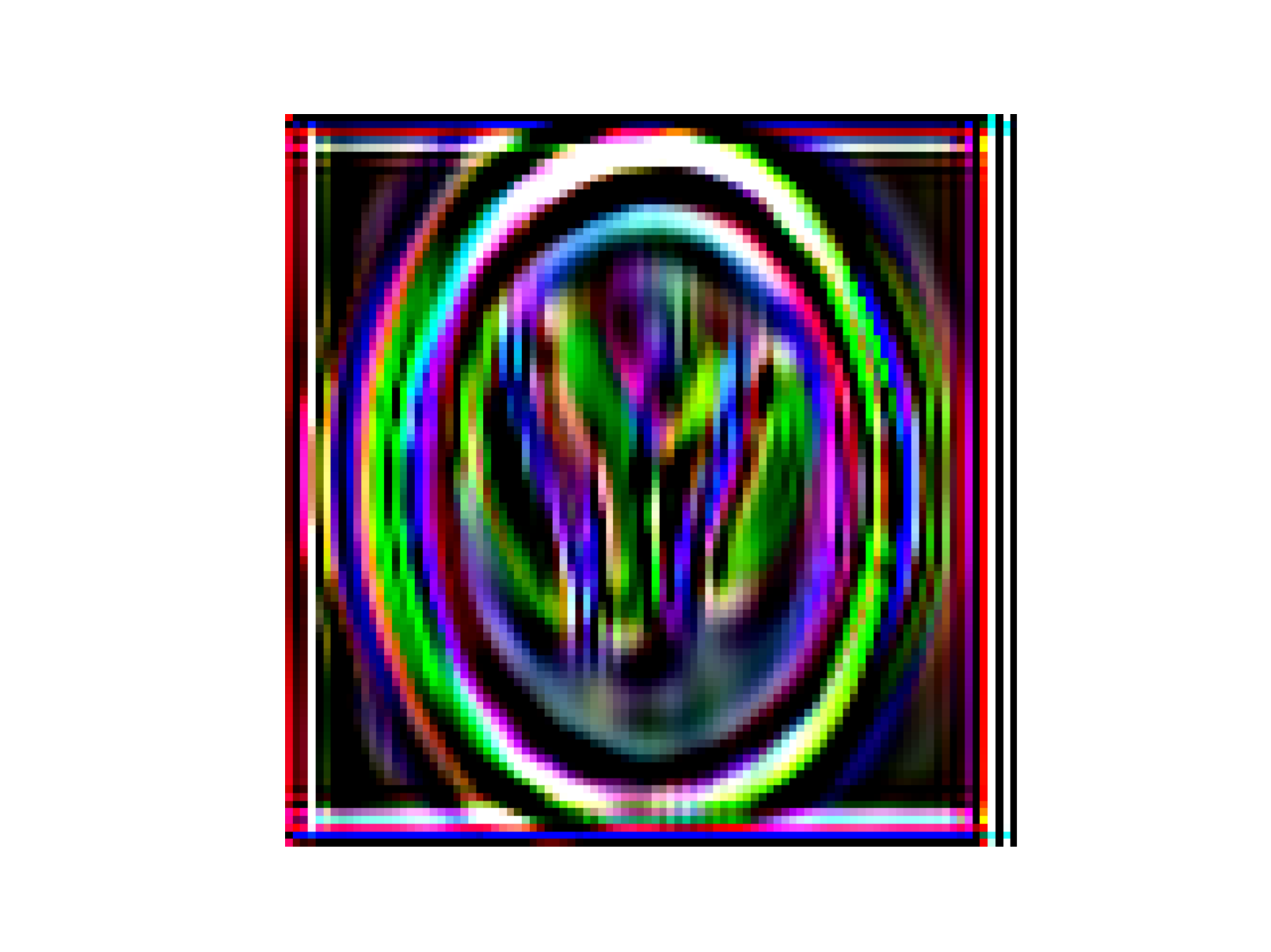}
     & \includegraphics[width=1.6cm, valign=c, trim={1cm 1cm 1cm 1cm},clip]{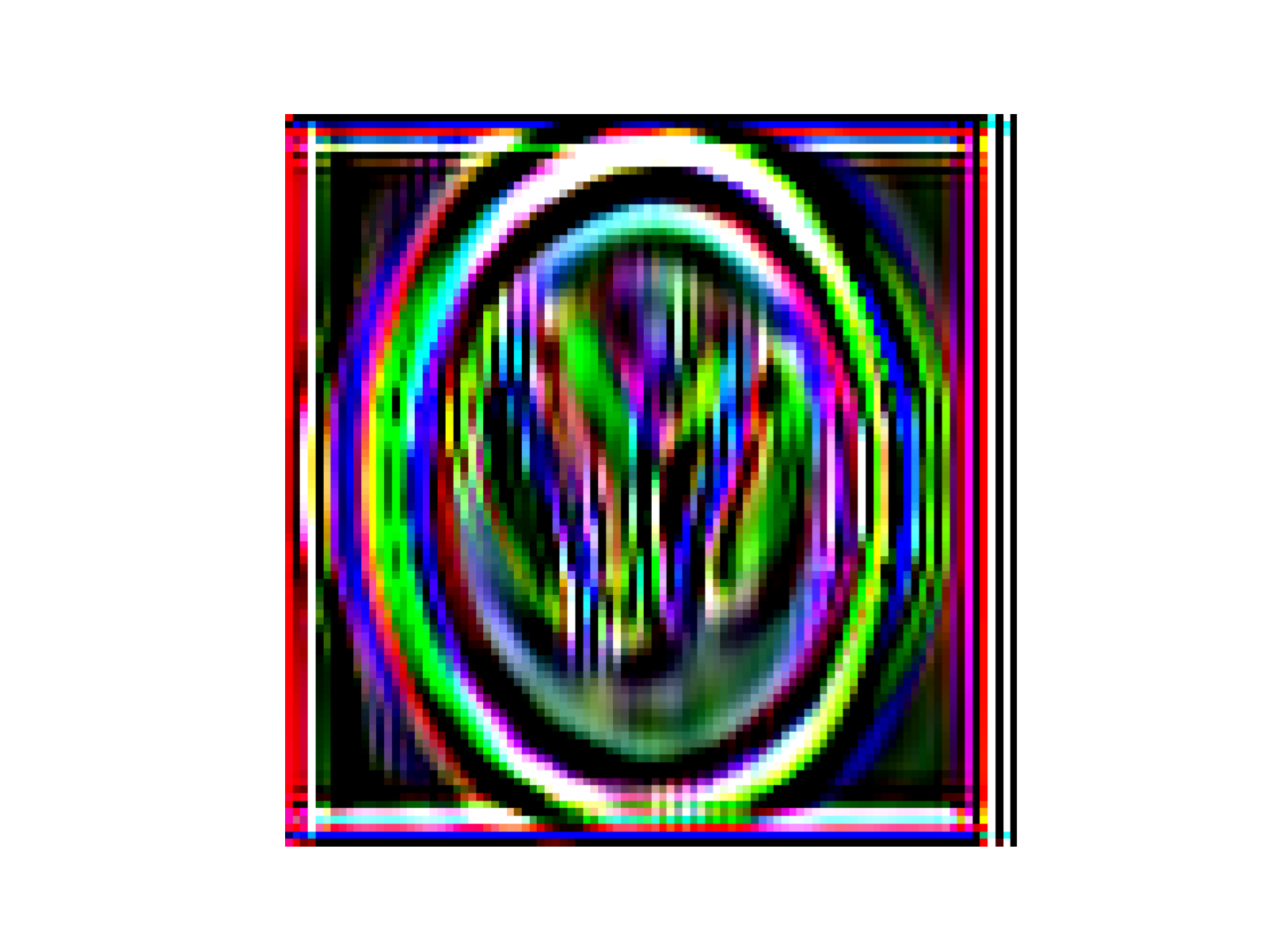}
     \\
     \hline
    \end{tabular}
    \captionof{figure}{The impact of iterative use the networks. Applying more iterations (going from left to right) of our LA-Net and Hyper-ResNet decreases the residual and error terms, while baseline UNet and ResNet networks diverge.}
    \label{fig:ood_iterations}
\end{table}

%%%%%%%%%%%%%%%%%%%%%%%%%%%%%%%%%%%%%%%%%%%%%

\section{Summary and Conclusions}
\label{sec6}

In this paper, we considered the approach of learning regularization functions for inverse problems using deep neural networks. Most recent ``process-based'' approaches are not based on variational techniques and therefore have the shortcoming that the resulting solution may not fit the data to a reasonable degree. Furthermore, while such approaches are motivated by the goal of minimizing a regularized objective function for the solution, in reality, there is no such objective to be minimized or evaluated using the deep network. 

We suggested a deep learning-based least-action regularization function. 
The particular choice of potentials (regularization) leads to a special form of residual network with a double skip connection. Its approximate minimization (i.e., setting its gradient to zero) leads to several forward-backward sweeps of a neural network {that we name LA-Net. To ease the computational burden of forward-backward iterations, we also propose and evaluate a feed-forward approximation of LA-Net, using a hyperbolic neural network, called Hyper-ResNet. While using LA-Net results in a regularized objective to be minimized, the Hyper-ResNet only approximates that process. Using a network that stems from a convex variational problem, such as LA-Net, offers theoretical advantages \cite{boydBook}. However, we found similar performance when using LA-Net and Hyper-ResNet, on the inverse problems we experimented with in this paper.}  

The results on two common inverse problems, namely, image deblurring and tomographic reconstruction, demonstrate that using the neural proximal approach we may face difficulties dealing with data fitting when the noise level changes or when the solution
is out of distribution. On the other hand, our DRIP method does not over or under fit the data, and is robust even when
the solution is out of distribution. That is partially thanks to the data-fitting projections that are applied during the reconstruction.

Unlike common applications in machine learning, the solution of inverse problems is motivated by directly fitting
the observed data to a given forward model. We therefore argue that while regularization techniques may change, the model should faithfully fit the data. Having a technique that combines the recent and powerful neural network architectures, and guarantees data fitting is thus important.
\bigskip
\section*{Acknowledgements and Funding Disclosure}
This research was supported by grant no. 2018209 from the United States - Israel
Binational Science Foundation (BSF), Jerusalem, Israel and by the Israeli Council for Higher Education (CHE) via the Data Science Research Center. ME
is supported by Kreitman High-tech scholarship.

\section*{Data Availability Statement}
The data supporting this study's findings are openly available and stated in the main text.

\bigskip
\section*{References}
\bibliographystyle{plain}
\bibliography{biblio} 

\end{document}